\documentclass[journal]{IEEEtran}

\ifCLASSINFOpdf
  \usepackage[pdftex]{graphicx}
\else
\fi

\usepackage{float}
\usepackage{caption}
\usepackage{subcaption} 
\usepackage{lipsum}
\usepackage{booktabs}
\usepackage{multirow}
\usepackage{hhline}
\usepackage{amsmath}
\usepackage{colortbl}
\usepackage{xcolor} 
\usepackage{amsfonts}

\DeclareMathOperator*{\argmax}{argmax} % thin space,
\DeclareMathOperator*{\argmin}{argmin} % thin space,
\usepackage{dblfloatfix}
\usepackage{url}
\usepackage{xcolor}
\usepackage{enumerate}

\usepackage[export]{adjustbox}
\usepackage{xspace}
\usepackage{todonotes}
\usepackage[ruled,vlined,linesnumbered]{algorithm2e}
\usepackage{paralist}

\usepackage[ruled,vlined]{algorithm2e}

\begin{document}

%\title{Real-World Adversarial Robustness Analysis of Real-time Semantic Segmentation Models for Autonomous Driving}
\title{On the Real-World Adversarial Robustness of Real-Time Semantic Segmentation Models for Autonomous Driving}

\author{
Giulio Rossolini, ~Federico Nesti, ~Gianluca D'Amico, ~Saasha Nair, \\ ~Alessandro Biondi, ~and Giorgio Buttazzo,~\IEEEmembership{Fellow,~IEEE}
% <-this % stops a space
%\thanks{M. Shell was with the Department
%of Electrical and Computer Engineering, Georgia Institute of Technology, Atlanta,
%GA, 30332 USA e-mail: (see http://www.michaelshell.org/contact.html).}% <-this % stops a space
\thanks{G. Rossolini and F. Nesti contributed equally to the work.}% <-this % stops a space
%\thanks{name.surname@santannapisa.it}
%\thanks{Manuscript received April 19, 2005; revised August 26, 2015.}
\\
\IEEEauthorblockA{Department of Excellence in Robotics \& AI, Scuola Superiore Sant'Anna, Pisa, Italy \\ name.surname@santannapisa.it}
%\\
}
\maketitle

\begin{abstract}
The existence of real-world adversarial examples (commonly in the form of patches) poses a serious threat for the use of deep learning models in safety-critical computer vision tasks such as visual perception in autonomous driving.
This paper presents an extensive evaluation of the robustness of semantic segmentation models when attacked with different types of adversarial patches, including digital, simulated, and physical ones.
A novel loss function is proposed to improve the capabilities of attackers in inducing a misclassification of pixels. %that geometrically reside far from the patch. 
Also, a novel attack strategy is presented to improve the Expectation Over Transformation method for placing a patch in the scene. Finally, a state-of-the-art method for detecting adversarial patch is first extended to cope with semantic segmentation models, then improved to obtain real-time performance, and eventually evaluated in real-world scenarios.
Experimental results reveal that, even though the adversarial effect is visible with both digital and real-world attacks, its impact is often spatially confined to areas of the image around the patch. This opens to further questions about the spatial robustness of real-time semantic segmentation models. 
\end{abstract}

\begin{IEEEkeywords}
Semantic Segmentation, real-world, adversarial attacks, adversarial examples, physical-world, autonomous driving.
\end{IEEEkeywords}

\IEEEpeerreviewmaketitle

\section{Introduction} \label{s:intro}

Deep neural networks reached super-human performance in many computer vision tasks. However, their results are typically threatened by a large set of inputs known as adversarial examples \cite{ae_srvey}. The existence of adversarial examples is an empirical proof of the poor robustness of deep learning models.
Most of the previous literature on adversarial examples focused on image classification and object detection tasks, whereas fewer works have been devoted to the assessment of the adversarial robustness of semantic segmentation models.

\emph{Semantic segmentation} (SS) plays a key role in the visual perception pipeline used in autonomous driving. Evaluating the robustness of SS models is therefore crucial, especially when they are used in safety-critical systems processing real-world images.
To this end, this paper proposes a general attack pipeline that can be used to craft adversarial patches against SS models. The proposed pipeline enables the use of \emph{multiple patches} to perform the attack in various locations of the image, which can be crafted by following both targeted and untargeted attack schemes.
Because of the timing constraints that must be typically satisfied in self-driving applications, the focus is put on \emph{real-time} SS models.

Multi-patch attacks are a threat that should seriously be considered in the design of robust computer vision models for autonomous driving: urban areas are typically filled with advertisement billboards, or road signs, each of which could become a potential attackable surface. Furthermore, we found that a double-patch attack is more effective than a single-patch attack, given that the effects of multiple adversarial patches can be combined.
%sum up to the same total area. 
%This is true as long as the total patch area is over a certain threshold, under which each of the two smaller patches retain no adversarial power anymore.

\begin{figure}
     \centering
     
     \begin{subfigure}{0.49\textwidth}
     \begin{subfigure}{0.495\textwidth}
         \centering
         \includegraphics[width=\textwidth, height=0.5\textwidth]{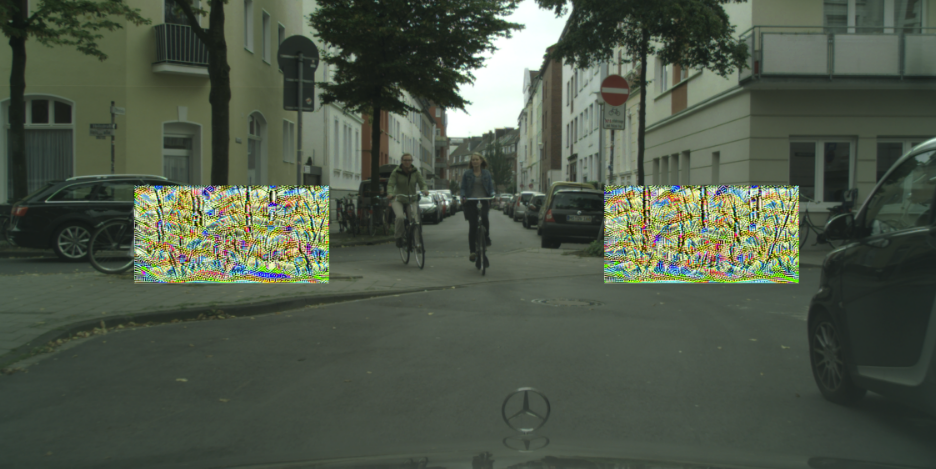}
         \vspace{-1.6em}
     \end{subfigure}
     \begin{subfigure}{0.495\textwidth}
         \centering
         \includegraphics[width=\textwidth, height=0.5\textwidth]{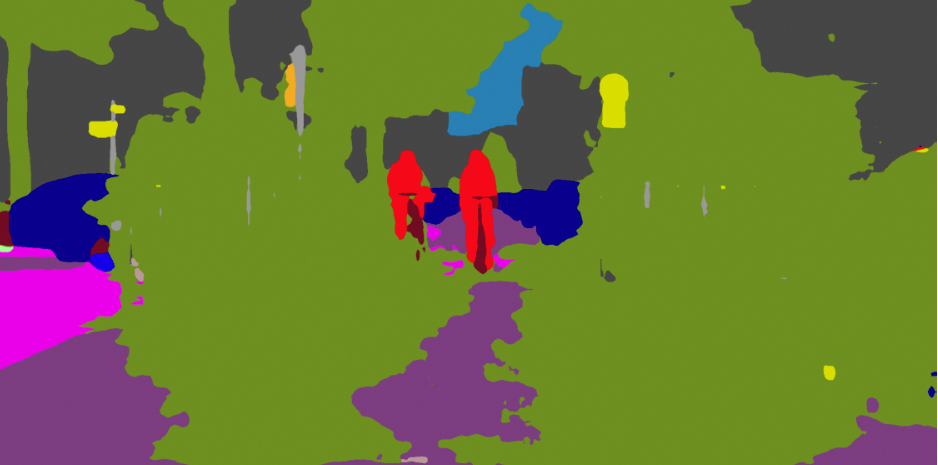}
         \vspace{-1.6em}
     \end{subfigure}
     \caption{}
     \end{subfigure}

     \begin{subfigure}{0.49\textwidth}
     \begin{subfigure}{0.495\textwidth}
         \centering
         \includegraphics[width=1.0\textwidth, height=0.5\textwidth]{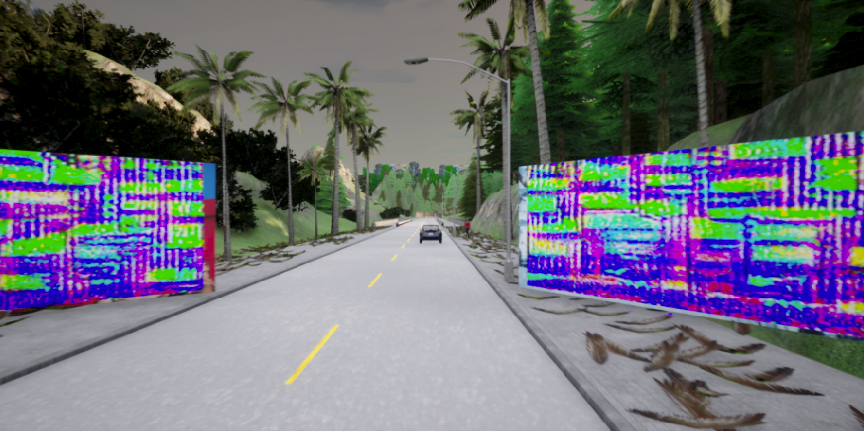}
          \vspace{-1.6em}
     \end{subfigure}
     \begin{subfigure}{0.495\textwidth}
         \centering
         \includegraphics[width=1.0\textwidth, height=0.5\textwidth]{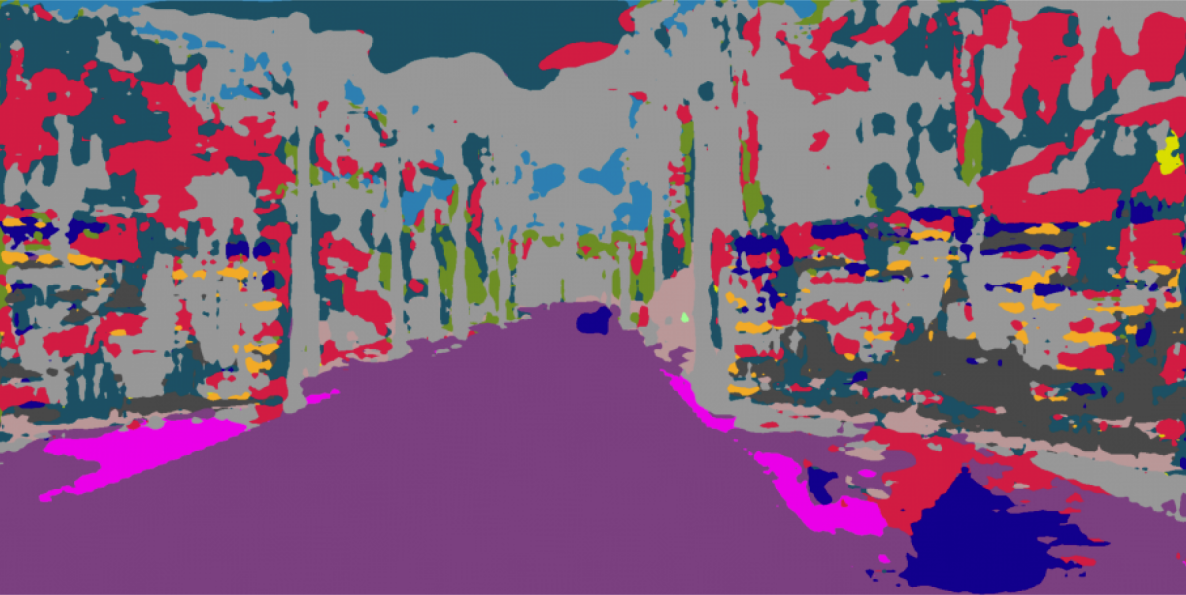}
        \vspace{-1.6em}
     \end{subfigure}
     \caption{}
     \end{subfigure}

     \begin{subfigure}{0.49\textwidth}
     \begin{subfigure}{0.495\textwidth}
         \centering
         \includegraphics[width=\textwidth, height=0.5\textwidth]{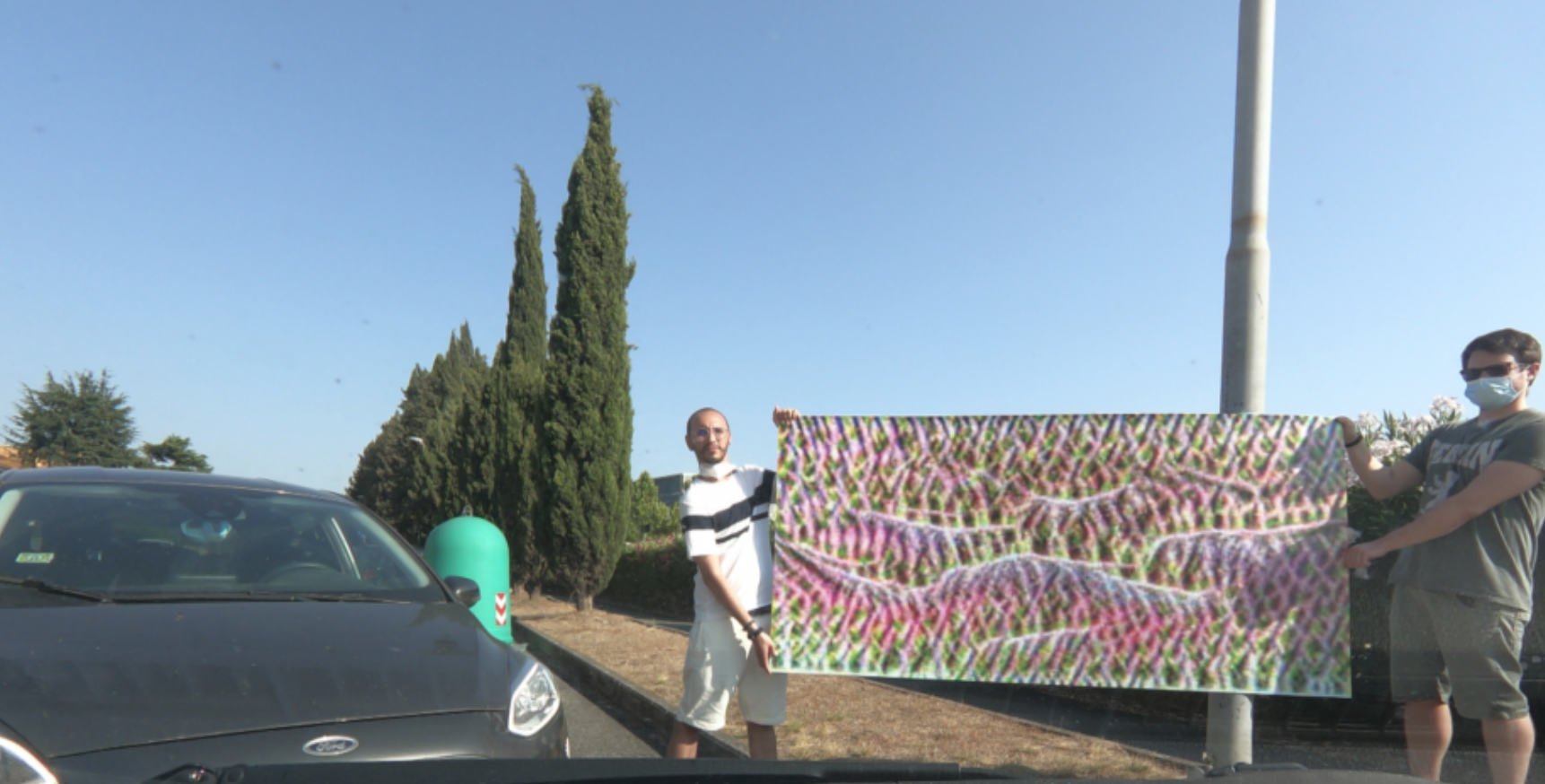}
         \vspace{-1.6em}
     \end{subfigure}
     \begin{subfigure}{0.495\textwidth}
         \centering
         \includegraphics[width=\textwidth, height=0.5\textwidth]{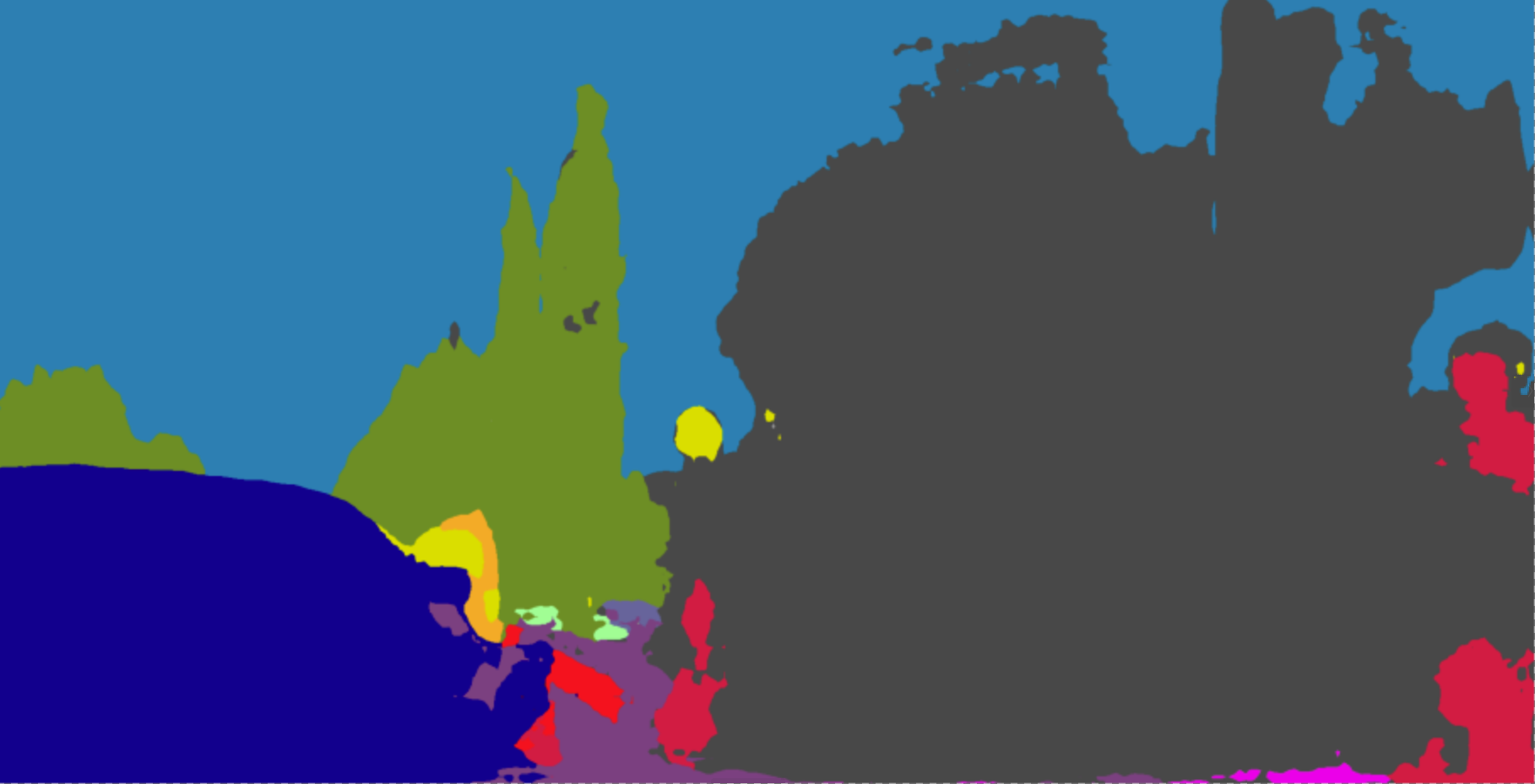}
        \vspace{-1.6em}
     \end{subfigure}
     \caption{}
     \end{subfigure}
    
    \caption{\small{Examples of effective patch-based attacks to scenes from (a) Cityscapes \cite{DBLP:conf/cvpr/CordtsORREBFRS16}, (b) the CARLA Simulator \cite{2017arXiv171103938D}, (c) a real-world environment. On the left-hand side, the figure shows three input images having adversarial patches, which are applied digitally, virtually, and physically. On the right-hand side, it reports the corresponding SS predictions obtained with DDRNet, BiSeNet, and ICNet, respectively. }}
    %the three under attack with the patch(s) applied, while on the right-hand side it reports the corresponding SS predictions obtained with DDRNet, BiSeNet, and ICNet, respectively.}
    \vspace{-1em}
    \label{fig:intro_exps}
\end{figure}

Particular attention is devoted to \emph{real-world attacks}.
Real-world datasets such as Cityscapes and custom real-world images are hence considered for evaluating the approaches studied in this paper. Furthermore, to enable a scalable testing based on scenes with patches placed as similar as possible as in the real world, we used the CARLA simulator~\cite{2017arXiv171103938D} to provide a realistic, fully-controllable environment, where billboards can be used as attackable surfaces. The geometric information from CARLA is used to design and perform a novel attack, named \emph{scene-specific} attack, which improves the Expection Over Transformation (EOT) formulation~\cite{pmlr-v80-athalye18b} with an accurate patch placement strategy.

Figure \ref{fig:intro_exps} shows a few examples of successful attacks on images from the Cityscapes dataset, on CARLA-generated images, and in the real world.

This work also points out the limitations of the standard cross-entropy loss function, as adopted by many other adversarial attacks, in attacking regions of the image that are geometrically far from the patch. Based on this observation, a novel loss function formulation is proposed and thoroughly tested against the standard cross-entropy. Experimental results show that our new formulation is much more effective than the cross-entropy loss in attacking the tested SS models.

Finally, a state-of-the-art adversarial patch detection method is extended for real-time SS models and tested in real-world scenarios.

The contributions of this paper can be summarized as follows:
\begin{itemize}
    \item it proposes a general attack pipeline that enables multi-patch attacks with both targeted and untargeted schemes against SS models;
    \item it proposes a novel loss function formulation that improves the capabilities of attackers with respect to the use of the standard cross-entropy loss;
    \item it presents an extensive evaluation of patch-based attacks against a set of real-time SS models using (i) real-world images from the Cityscapes dataset, (ii) synthetic images from the CARLA simulator, and (iii) other real-world images that where taken in our city;
    % \item \textcolor{blue}{TODO: rewrite last two} it extends a state-of-the-art detection method to be used with real-time SS models and evaluates its effectiveness against real-world adversarial patches;
    % \item it evaluates the effect of a printed adversarial patch in the real world as well as the benefits of the proposed detection method.
    \item it extends a state-of-the-art attack detection method to be used in real-time with SS models;
    \item it presents an extensive set of experiments aimed at validating the proposed attack and defense approaches in real-world scenarios.
    %as well as the concrete benefits of the proposed detection method.
\end{itemize}

The remainder of the paper is organized as follows: Section \ref{s:related} introduces the related work, Section \ref{s:proposed} presents the proposed methods and algorithms, Section \ref{s:exp} discusses the experimental results, and Section \ref{s:conclusions} concludes the paper.

\section{Related Work} \label{s:related}

The literature related to the topics addressed in this work is vast. Therefore, it has been categorized into (i) Real-time SS models, (ii) Adversarial studies on SS, (iii) Real-world adversarial examples, and (iv) Defenses against adversarial patches.

\paragraph{Real-Time SS Models} 
Early works on SS architectures, such as the Fully Convolutional Network (FCN) \cite{long_fully_2015} and the UNet \cite{ronneberger_u-net_2015}, were extended with methods that exploit spatial and context information of images: dilated convolutions, pyramid pooling modules, transformers and very deep layers, etc.~\cite{chen_deeplab_2016, pspnet_paper, yuan_object-contextual_2020, sun_high}.

Although such models show high accuracy, they require powerful hardware platforms and imply large execution times when processing high-resolution images \cite{icnet_paper}, making them not suited for real-time vision applications like autonomous driving.
To overcome this issue, several works \cite{icnet_paper, ddrnet_paper, BiSeNet_paper, gao2021rethink, li2020semantic} have recently proposed lightweight models that show acceptable accuracy on high-resolution images while taking into account real-time requirements. 
Among the vast number of available models, we restricted our attention to three popular real-time models: ICNet, DDRNet, and BiSeNet (see Section~\ref{s:exp} for the experimental settings and the supplementary materials for additional details).

The reason for selecting these models is to evaluate the robustness of several distinct approaches.
As it is well-known in the literature, the majority of real-time semantic segmentation models leverage two or more parallel paths for extracting context and spatial information \cite{app11198802, ddrnet_paper}. However, the internal operations they adopt are different and some of them can be more prone to adversarial attacks.
ICNet is one of the first real-time SS models that merges multiple paths with Cascade Features Fusion Units \cite{icnet_paper}. 
Although nowadays several models reach faster performance, ICNet is considered an important milestone that is still taken as a reference in recent studies (e.g.,~\cite{arnab_robustness_nodate, pfeuffer2019robust}). 
BiSeNet and DDRNet are instead state-of-the-art models, with smaller inference time with respect to ICNet, both based on splitting the extraction of context and spatial information between two parallel network branches. However, the former exploits self-attention modules, while the latter involves multiple residual points between the two branches.

\paragraph{Adversarial studies on SS}
%While it is evident that the short-term future of autonomous driving research tests interest on  fast SS models [CITE], the hard utilization of learning systems question their \textbf{real robustness} against possible adversarial attacks. 

Previous works \cite{bar_vulnerability_2021, 
metzen_universal_2017, 
DBLP:conf/iccv/XieWZZXY17,  
DBLP:journals/access/KangSDG20, 
2019arXiv190805005K,  
arnab_robustness_nodate, 
nakka_indirect_2019,
shen2019advspade} proved that both targeted and untargeted pixel-based perturbations easily fool SS models by extending well-known adversarial strategies (e.g., \cite{DBLP:journals/corr/SzegedyZSBEGF13,pgd_attack}) from image classification. 
Consequently, Nakka et al. \cite{nakka_indirect_2020} presented an interesting study on the robustness of SS models on autonomous driving datasets by showing that it is possible to perturb a precise area of pixels to change the SS prediction corresponding to specific objects placed in the whole image.
  
However, although the robustness of these convolutional architectures appears fragile at a first glance, the above works do not take two important aspects into consideration.
First, they evaluated the adversarial effect against non-real-time networks only. From our point of view, this leads to unrealistic assessments, since real-time architectures usually do not rely on the rich sequence of modules used by the evaluated networks, which could be practically prone to adversarial attacks \cite{nakka_indirect_2020}.
Second, and most importantly, they did not consider real-world adversarial objects, which might represent a real threat for driving scenarios. 

%{\color{red}
Differently from the strategies adopted by previous works, we believe that evaluating the robustness of a model for autonomous driving requires: (i) a precise selection of the tested models, and (ii) considering real-world adversarial examples.
In fact, digital adversarial perturbations are not reproducible in the real-world, since they can be used to attack an autonomous driving perception pipeline only when the attacker has control on the digital representation of the image. If this is assumed, more effective system-level attacks may be used, %likely also possible,
hence reducing the relevance of adversarial attacks to SS models.
Therefore, this work considers real-world adversarial examples as malicious objects that, once placed into a physical environment, are capable of externally injecting their adversarial effect. 
%}

\paragraph{Real-world Adversarial Examples}
Consistently with the latter observation, multiple works studied real-world adversarial examples for DNNs. 
Kurakin et al. \cite{kurakin_adversarial_2017} introduced a strategy for crafting physical-world adversarial pictures against image classification models, without the ability to manipulate the digital representation of inputs. That work opened to a new class of adversarial examples called real-world adversarial examples (RWAEs).
Athalye et al. \cite{pmlr-v80-athalye18b} improved the robustness of RWAEs by proposing the Expectation Over Transformation (EOT) algorithm, which accounts for transformations typical of real-world scenarios (acquisition noise, changes in the point of view, etc.) during the optimization process. This allows crafting objects that retain the adversarial property when capturing images from different points of views.  

Consequently, the EOT formulation opened to the development of adversarial patches \cite{brown_adversarial_2018}, which are robust, localized, image-agnostic perturbations that fool neural networks when placed in the input scene. 

Although extensive prior work presented physical attacks for image classification~\cite{brown_adversarial_2018, sharif_accessorize_2016, eykholt_robust_2018}, object detection \cite{DBLP:conf/eccv/WuLDG20, wu_physical_2020, lee_physical_2019, zhang_camou_2019}, optical flow \cite{2019arXiv191010053R}, LiDAR object detection \cite{Tu2020PhysicallyRA}, and depth estimation \cite{2020arXiv201003072Y}, only a few focused on autonomous driving tasks. One reason might be that testing the adversarial robustness in autonomous driving context is more challenging, as it requires a certain control on the driving environments. Other works \cite{zhang_camou_2019, wu_physical_2020} have shown autonomous driving simulators, and CARLA~\cite{2017arXiv171103938D} in particular, to be a viable solution in alleviating this issue by crafting and evaluating adversarial situations in virtual 3D environments. 

Please note that all the works mentioned above investigated the effect of adversarial objects on tasks different from SS. The first study addressing the evaluation of the robustness of real-time SS models against real-world adversarial attacks was proposed in \cite{prev_work}, where both the Cityscapes dataset \cite{DBLP:conf/cvpr/CordtsORREBFRS16} and CARLA were used to investigate the effect of adversarial patches in the real world. However, no experiments were reported to evaluate the robustness of SS models against multi-patch scenarios and targeted attacks. Also, no defense strategies were proposed to mitigate adversarial vulnerability at runtime.
The present paper fills these gaps by presenting novel attack objectives,  additional experiments in multi-patch scenarios, and a real-time adversarial detection mechanism.

\paragraph{Defense mechanisms against adversarial patches}
The literature presents a wide set of defense mechanisms against adversarial attacks. While the methods used to detect generic adversarial examples are the vast majority \cite{ae_srvey}, \cite{nesti_detecting2021}, other works addressed the problem of detecting patches.

Some works \cite{saha_role_2020, chiang_adversarial_2021, metzen2021meta, 2020arXiv200502313R} exploit gradient-masking or adversarial training strategies to increase the robustness of the model against adversarial patches. 
%These methods do not provide detection of adversarial patches. As such, they are not considered for comparison.
Although such mechanisms help reduce the adversarial effect induced by patches, they do not provide any strategy to detect attacked images that are still dangerous for the tested models.
As such, this class of defense methods is not considered for comparison.

Methods that detect images attacked with adversarial patches are based on masking and occlusion strategies \cite{xiang_patchguard_2021, xiang_patchcleanser_2021,chou_sentinet_2020, xu_lance_2019, li_detecting_2021} and run-time statistical analysis \cite{co_real, coverage_defense}. 
Although both these two approaches achieve high detection performance, most of them introduce significant latency during the model inference and hence are not suitable for real-time applications. 
Furthermore, please note that most of the works mentioned in this section are designed to operate on image classification and object detection models, and only some of them can be adapted to SS architectures. 

To the best of our knowledge, the only method that can be directly extended to cope with SS and is capable of providing real-time performance in detecting images affected by adversarial patches is HyperNeuron (HN) \cite{co_real}. HN is based on detecting the over-activation of internal neurons, which is a symptom of the presence of an adversarial patch. 

In the present paper, we propose an extension of the HN algorithm that further improves the detection performance in the SS domain.
First, a feature compression step is introduced to reduce the number of features to be processed at run time.
This is of crucial importance when dealing with SS models, where the features space may have high dimensions. Second, the selection of over-activated features is simplified by using a threshold computed off-line rather than expensive operations performed at run-time.
Based on these ideas, Section~\ref{s:proposed} presents a novel method that achieves similar performance to HN, but with significantly lower latency, making it more suitable for real-time applications.

\paragraph{This work}
This work provides a comprehensive study of the robustness of real-time SS models applied to autonomous driving scenarios. 
This is accomplished by defining and evaluating several real-world adversarial patch attacks, a novel loss function formulation, and a real-time patch detection mechanism. Such an extensive evaluation is missing in the literature, and could represent a promising milestone for studying and evaluating the robustness of semantic segmentation models for autonomous driving.

\section{Patch-based attack pipeline} \label{s:proposed}
This section presents the adopted notation, the attack pipeline used to generate real-time adversarial attack to SS models, the proposed loss function formulation, and the Fast Patch Detection Algorithm.

\subsection{Preliminaries}

We consider input images with height $H$, width $W$, and $C$ channels, denoted as $x \in [0,1]^{H \times W \times C}$.
An SS model discriminating between $N_c$ classes is thus represented by a function $f: x \mapsto [0,1]^{(H \cdot W) \times N_c}$, which gives the predicted class-probability scores associated with each image pixel $i$.
More specifically, the predicted probability score for pixel $i$ corresponding to class $j$ is denoted by $f_{i}^{j}(x) \in [0,1]$.

The predicted SS of each pixel $i$, denoted by $SS_i(x)$, is then computed by extracting the class with the highest probability score of the pixel, i.e., $SS_i(x) = { \argmax_{j \in \{1,...,N_c\} }}{f_i^j(x) }$. The complete semantic segmentation prediction $SS(x)$ is then the collection $\{SS_i(x), ~ i = 1,...,H \times W \}$.
%$SS_i(x)~, \forall i \in \{1,...,H \times W\}$.

The ground truth for the SS of $x$ is defined as $y \in \mathbb{N}^{H \times W}$, and assigns the correct class (in $\{1,...,N_c\}$) to each pixel.

%The performance of the SS models is evaluated by computing the cross-entropy loss $\mathcal{L}_{CE}(f_i(x), y_i) = -log({f_i^{y_i}(x)})$. Thus, for each pixel $i$, the model's prediction $f_i$ is compared against the ground truth class $y_i$.

An adversarial patch of height $\tilde{H}$ and width $\tilde{W}$ is denoted by $\delta \in [0,1]^{\tilde{H} \times \tilde{W} \times C}$, where 
$\tilde{H}<H$ and $\tilde{W}<W$. 

In a multi-patch setting, we consider a set of patches $\Delta = \{\delta_k: k=1, ..., N_p\}$, where $N_p$ is the number of patches used for the attack.

%This set of patches is then added to the original image $x$ to obtain a patched image $\tilde{x}$.
%Thus, the output of the SS model on this patched image would now be $f(\tilde{x})$.

\subsection{General Attack Pipeline}

\begin{figure*}[!t]
\centering
\makebox[\columnwidth]{\includegraphics[scale=0.21]{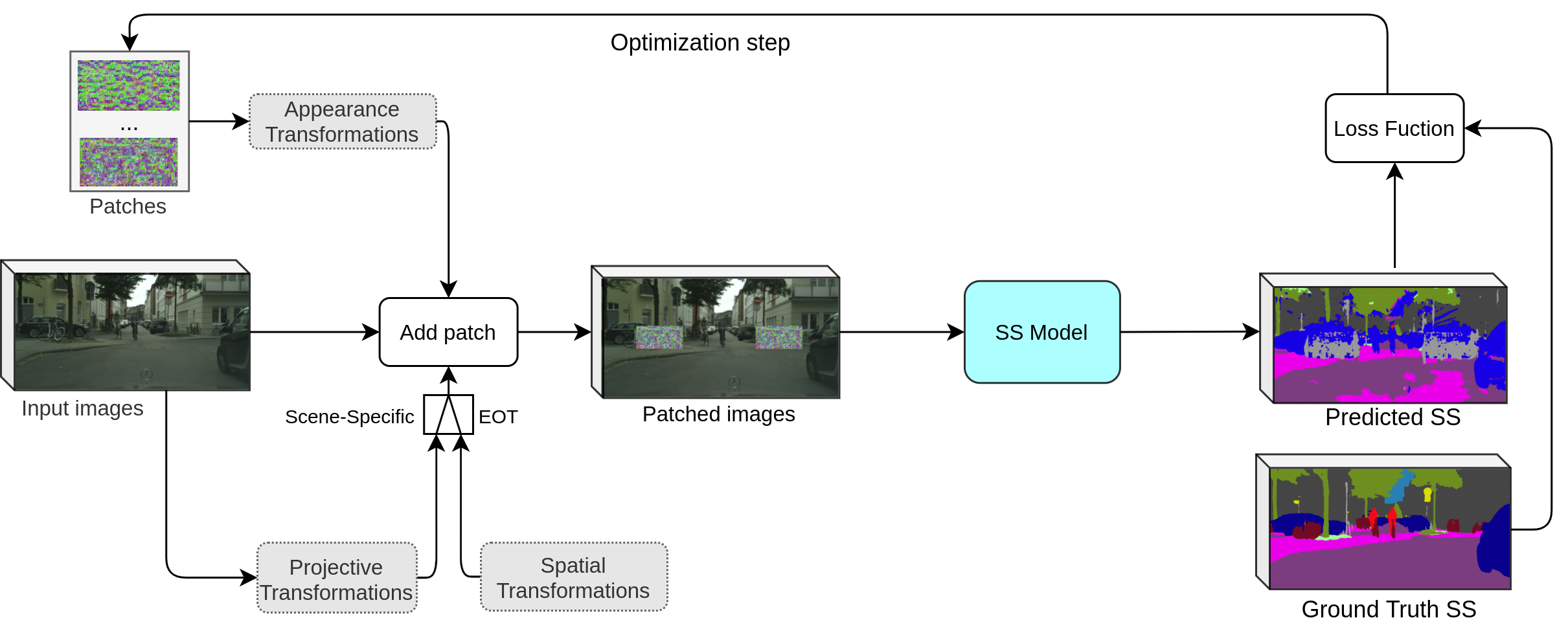}}
\caption{\small{Scheme of the proposed approach for crafting both the EOT-based and the scene-specific patches.}}
\vspace{-1em}
\label{fig:attack_scheme}
\end{figure*}

All the patch-based attacks considered in this paper are generated with the pipeline illustrated in Figure \ref{fig:attack_scheme}.

% Figure~\ref{fig:attack_scheme}.

Inspired by the universal (i.e., image-agnostic) attacks \cite{2016arXiv161008401M} and the EOT-based attacks \cite{pmlr-v80-athalye18b}, the objective is to find an optimal patch set $\Delta^*$ by optimizing a certain loss function $\mathcal{L}$ for all the patched images in expectation, according to the distribution of transformations used to apply the patch set $\Delta$ on the image set $\mathbf{X}$.

In particular, we define:
\begin{compactitem}
    \item A \textbf{set of appearance-changing transformations $\Gamma_a$}: each element of $\Gamma_a$ is a composition of illumination changes (brightness and contrast) and noise addition (uniform or Gaussian). This set of transformations is randomly sampled and the selected transformation is directly applied to the patches $\Delta$. The typical parameters of these transformations are randomized during the optimization to make the patches robust to illumination changes and acquisition noise. 
    \item A \textbf{patch placement function $\eta$} that defines which portion of the original image $x$ is occupied by the patches. The function $\eta$ can have different definitions depending on the chosen attack. Section \ref{ss:specific} provides details of the patch placement function.
    \item A \textbf{patch application function $g(x,  \Delta, \Gamma_a, \eta)$} that replaces the area(s) of the image $x$ specified by $\eta$ with transformed versions of the patches in $\Delta$ obtained by a transformation randomly selected from $\Gamma_a$, hence returning the patched image $\tilde{x}$.
    \item A \textbf{loss function} $\mathcal{L}$, which is the objective function to be optimized. $\mathcal{L}$ consists of a weighted sum of multiple loss sub-functions. The adversarial effect is obtained through the optimization of the adversarial loss $\mathcal{L}_{adv}$ (detailed in Section \ref{ss:loss}). To ensure that the patch transfers well to the real world, two additional losses are considered to account for the \emph{physical realizability} of the patch: smoothness loss $\mathcal{L}_S$ and non-printability score $\mathcal{L}_N$. Please, refer to the supplementary material for details.
\end{compactitem}

The optimization problem can therefore be written as 
\begin{equation}
    \Delta^* = \argmin_{\Delta} ~ \mathbb{E}_{x \in \mathbf{X}, \zeta_a \in \mathbf{\Gamma_a}, \eta} ~ 
    \mathcal{L}(f(\tilde{x}), y) ~
\end{equation}

\noindent whereas its practical iterative implementation at step $t$ is:
\smallskip
%\begin{equation} \label{e:opt}
%\begin{aligned}
%    \delta_{k, t+1} = \mathrm{clip}_{[0,1]} \left(\delta_{k, t} + \epsilon \cdot\sum_{x \in \mathbf{X}} \sum_{\substack{\zeta_a\in\mathbf{\Gamma_a}\\ \eta}}\nabla_{\delta_{k, t}} \mathcal{L}(f(\tilde{x}), y) \right ), 
%\end{aligned}
%\end{equation} 
%
\begin{equation} \label{e:opt}
\begin{aligned}
    \delta_{k, t+1} = \mathrm{clip}_{[0,1]} \left(\delta_{k, t} + \epsilon \cdot\sum_{x \in \mathbf{X}} \nabla_{\delta_{k, t}} \mathcal{L}(f(\tilde{x}), y) \right ), 
\end{aligned}
\end{equation} 
%\label{eq:adv_method}

\noindent where $\tilde{x} = g(x, \Delta, \Gamma_a, \eta)$, $k=\{1, ..., N_p\}$, and $\epsilon$ represents the step size.

The following subsections detail the optimization objectives, the patch placement functions considered for the EOT-based and scene-specific attacks, and the loss functions used.

\subsection{Untargeted and Targeted attacks}
The attacker's objective is encoded in the performed optimization to craft adversarial patches. %{\color{red} The attacker's objective is encoded into the loss function used during the optimization.} 
The attacker might want to maximize the prediction error of the network, regardless of the output classes (\textit{untargeted} attack), or force the network prediction towards a specific output (\textit{targeted} attack). 

The objective of an untargeted attack is then to maximize the loss function $\mathcal{L}_{adv}(f(\tilde{x}), y)$, defined later in Section \ref{ss:loss}, where $y$ is the ground-truth label.
Hence, in Equation \eqref{e:opt}, $\mathcal{L}(f(\tilde{x}), y) = -\mathcal{L}_{adv}(f(\tilde{x}), y)$ (the additional losses for physical realizability of patches are neglected for simplicity).

\begin{figure}
\centering
    \begin{subfigure}{0.155\textwidth}
         \centering
         \adjincludegraphics[width=\textwidth, trim={{.3\width} {.35\height} {.35\width} {.25\height}},clip]{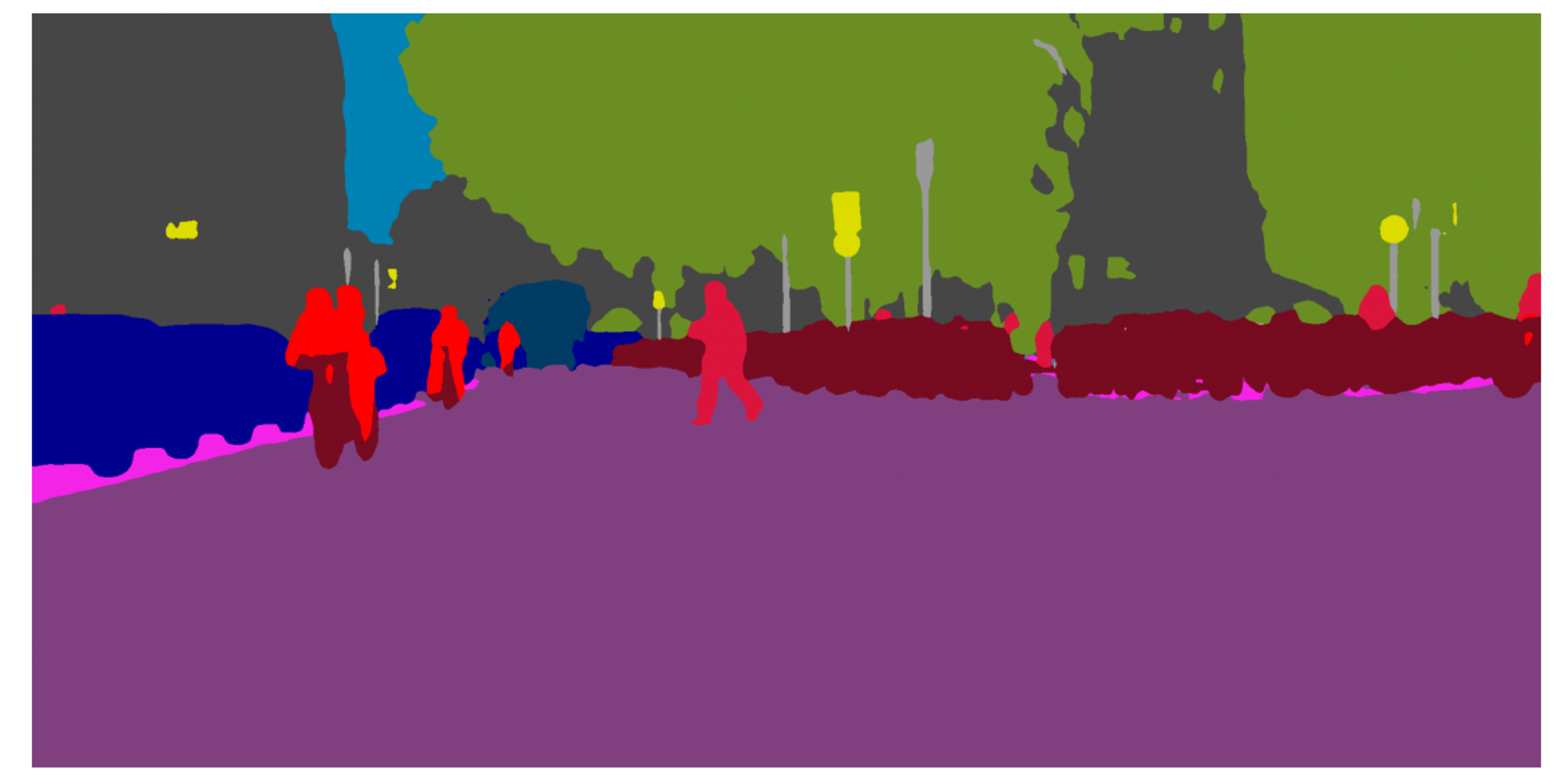}
         %caption{}
         \label{fig:label_targeted}
    \end{subfigure}
    \begin{subfigure}{0.155\textwidth}
         \centering
         \adjincludegraphics[width=\textwidth, trim={{.3\width} {.35\height} {.35\width} {.25\height}},clip]{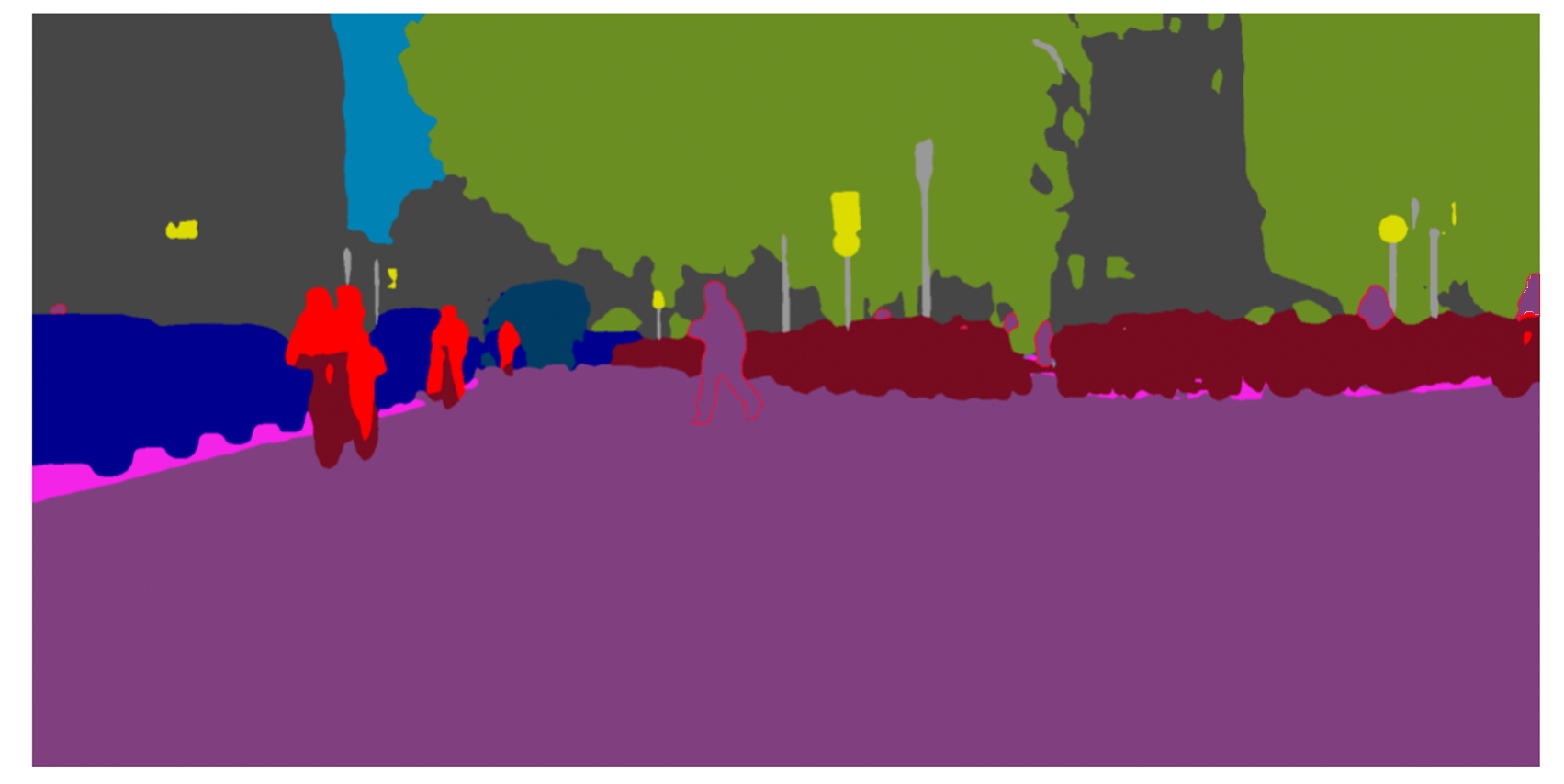}
         %caption{}
         \label{fig:label_road}
    \end{subfigure}
    \begin{subfigure}{0.155\textwidth}
         \centering
         \adjincludegraphics[width=\textwidth, trim={{.3\width} {.35\height} {.35\width} {.25\height}},clip]{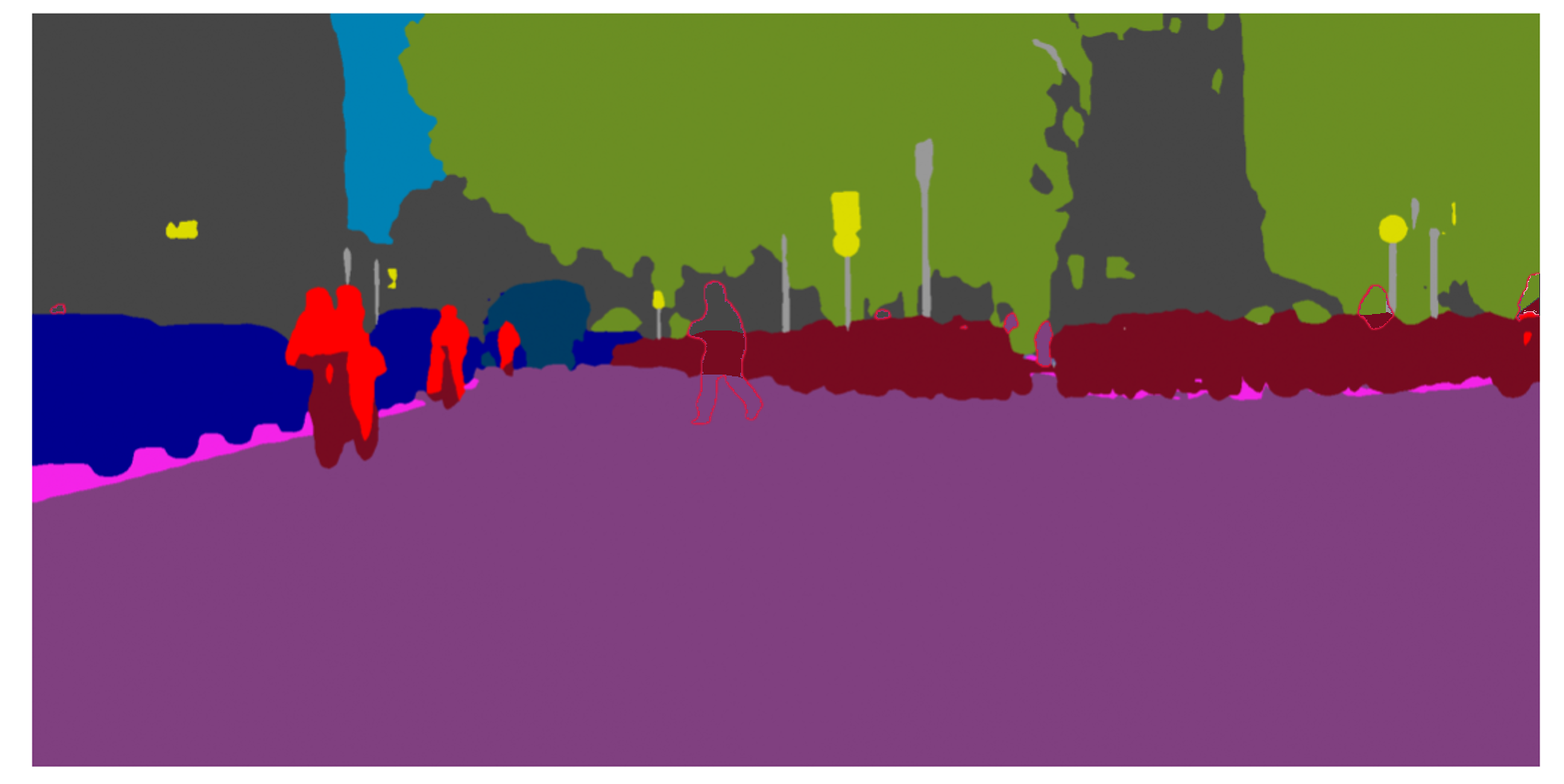}
         %caption{}
         \label{fig:label_nn}
    \end{subfigure}
        \vspace{-1.5em}
        \caption{\small{Illustration of how the original label (left) can be modified to attack the class \emph{pedestrian}: by changing \emph{pedestrian} with \emph{road} class (middle), or by using the nearest neighbor approach (right). The boundary of the pedestrian are kept in the original color for illustrative purposes.}}
        \vspace{-1.5em}
        \label{fig:targeted_labels}
\end{figure}

Conversely, to perform a targeted attack, the attacker has to first specify the desired prediction of the network. There are many ways to define a target for this problem (for instance, providing the label of a completely different scene), but we focus to a case that is more interesting for real-world applications: forcing the network to make a specific class ``disappear" from its prediction. This can be done by uniformly changing the pixels belonging to class $c_{\text{attacked}}$ into the ones of another class $c_{\text{target}}$, or by applying the nearest neighbor algorithm \cite{nakka_indirect_2020} to provide a more realistic target label, as illustrated in Figure \ref{fig:targeted_labels}. The nearest neighbor algorithm associates to each pixel belonging to $c_{\text{attacked}}$ the class of the closest pixel of a different class.

We define our target label as $y_t = \tau(y, c_{\text{attacked}}, c_{\text{target}})$, where $c_{\text{attacked}}$ might indicate either a specific class, or the nearest neighbor approach (specified as a pseudo-class \textbf{NN}).

The targeted attack is then performed by considering, in Equation \eqref{e:opt}, $\mathcal{L}(f(\tilde{x}), y) = \mathcal{L}_{adv}(f(\tilde{x}), y_t)$.

Both targeted and untargeted attacks are evaluated in Section \ref{s:exp}.

\subsection{EOT-based and Scene-specific attack} \label{ss:specific}
The patch placement within the image might follow two different approaches: (i) randomizing the patch position, scale and rotation at each iteration (i.e., using the EOT method), or (ii) using accurate projective transformations.

While the first is a general-purpose method for real-world adversarial patch generation (hereby, refered to as ``EOT-based attack"), the latter, named \emph{scene-specific attack}, exploits the geometrical information provided by the CARLA simulator to compute camera extrinsic and intrinsic matrices, and the pose of the attackable surface (a billboard). This enables the computation of precise camera-to-billboard 3D rototranslation (details in the supplementary materials) that is used to warp the patch according to the actual point of view of the camera in each image of the dataset. 

Formally, the function $\eta$ in Equation \eqref{e:opt} is a composition of randomized translation, rotation, and scaling for the EOT-based attack, or a precise projective transformation for the scene-specific attack.

This novel method presents two main advantages: it generates stronger attacks (since the placement is more accurate), and it does not require randomization of the patch placement (saving time during optimization).

To apply this method, a digital representation of the target scene is required to extract geometrical data. Although CARLA can import cities via OpenStreetMaps\footnote{\url{https://www.openstreetmap.org/}}, some amount of manual effort is required to model 3D meshes and include objects in the virtual world. Such objects must be carefully designed to ensure that patches will transfer well to the real world.
The transfer issues between CARLA and the real world will be investigated in a future work.
Section \ref{s:exp} provides a comparison of this method against the EOT-based attack.

\subsection{Proposed loss function} \label{ss:loss}

The pixel-wise cross-entropy (CE) loss, denoted by $\mathcal{L}_{CE}$, has been shown to work well for untargeted digital attacks by adding a perturbation $r$ to pixels values~\cite{nakka_indirect_2019}~\cite{bar_vulnerability_2021}. In this case, the loss is $\mathcal{L}_{adv}(f(x+r), y) = \frac{1}{|\mathcal{N}|}\cdot \sum_{i \in \mathcal{N}} \mathcal{L}_{CE}(f_i(x+r), y_i)$, where $f_i(x+r)\in[0, 1]^{N_c}$ is the output of the model for each pixel (i.e., a probability distribution over the $N_c$ classes), and $\mathcal{N} = \{1,...,H \times W\}$ is the entire set of pixels in $\tilde{x}$.
This formulation can be changed to generate stronger attacks against SS models.

If $\mathcal{\tilde{N}}_k =\{1,..., \tilde{H}_k \times \tilde{W}_k \}  \subseteq \mathcal{N}$ denotes the pixels belonging to a patch $\delta_k$ only, the totality of the pixels belonging to the patches can be defined as $\mathcal{\tilde{N}} = \bigcup\limits_{k=1}^{N_p} \mathcal{\tilde{N}}_k$.

Then, the subset of pixels that do not belong to any patch and are still predicted correctly by the model is denoted by:
%\useshortskip
\begin{equation}
\mathbf{\Upsilon} = \{ i \in  \mathcal{N} \setminus \mathcal{\tilde{N}} \quad | \quad SS_i(\tilde{x}) = y_i \}.
\end{equation}

Using $\mathbf{\Upsilon}$, the previous pixel-wise CE loss computed on $\mathcal{N} \setminus \mathcal{\tilde{N}}$ can be split into two terms:
%\useshortskip
\begin{equation}
    \mathcal{L}_M^{\tilde{x}}  =  \sum_{i \in \mathbf{\Upsilon}} \mathcal{L}_{CE}(f_i(\tilde{x}), y_i), \quad
    \mathcal{L}_{\overline{M}}^{\tilde{x}}  =  \sum_{i \notin \mathbf{\Upsilon}} \mathcal{L}_{CE}(f_i(\tilde{x}), y_i) ~.
\end{equation}

\noindent
where $\mathcal{L}^{\tilde{x}}_M$ describes the cumulative CE for the misclassified pixels, while $\mathcal{L}^{\tilde{x}}_{\overline{M}}$ refers to the remaining ones.

Note that both $\mathcal{L}^{\tilde{x}}_M$ and $\mathcal{L}^{\tilde{x}}_{\overline{M}}$ do not include pixels of the patch to focus the attack on image areas away from the patch.

Using these separate loss terms, we prevent that the contributions of the correctly classified pixels gets obscured by the other term. Therefore, the gradient of the overall adversarial loss is redefined as follows:
%\useshortskip
\begin{equation}
\nabla_{\delta_k} \mathcal{L}_{adv}(f(\tilde{x}), y) = \gamma \cdot \frac{ \nabla_{\delta_k} \mathcal{L}_M^{\tilde{x}}}{||\nabla_{\delta_k} \mathcal{L}_M^{\tilde{x}}||_{2}} + (1-\gamma) \cdot \frac{ \nabla_{\delta_k} \mathcal{L}_{\overline{M}}^{\tilde{x}}}{||\nabla_{\delta_k} \mathcal{L}_{\overline{M}}^{\tilde{x}})||_{2}}  ~,
\end{equation}

\noindent where $\gamma \in [0,1]$ is a balancing factor that determines whether the optimization should focus on decreasing the number of correctly classified pixels or improving the adversarial strength for the currently misclassified pixels, and $k=\{1, ..., N_p\}$. In other words, $\gamma$ balances between the importance of $\mathcal{L}_M$ and $\mathcal{L}_{\overline{M}}$ at each iteration $t$ of the optimization problem in Equation~\eqref{e:opt} depending on the number of pixels that are not yet misclassified.

To provide an automatic tuning of $\gamma$ at each iteration, an adaptive value of $\gamma = \frac{|\Upsilon|}{|\mathcal{N} \setminus \mathcal{\tilde{N}}|}$ is proposed. 
The idea is to initially focus on boosting the \emph{number} of misclassified pixels. As this number increases, the focus of the loss function gradually shifts toward improving the adversarial strength of the patch on the misclassified pixels.

Section \ref{s:exp} provides an extensive analysis of the proposed loss function by comparing multiple values of $\gamma$ with the standard pixel-wise CE measured both on $\mathcal{N} \setminus \mathcal{\tilde{N}}$ and $\mathcal{N}$ (which is used by \cite{nakka_indirect_2019}), suggesting that our formulation is indeed more general and effective for this kind of attacks.

\subsection{Fast Patch Detection Algorithm} \label{s:proposed_det}

This section describes the Fast Patch Detection Algorithm (FPDA) designed to recognize SS predictions affected by adversarial patches. 
Inspired by the HN method \cite{co_real}, the basic idea of the proposed approach relies on identifying and counting the presence of over-activated internal features, commonly caused by adversarial attacks to induce erroneous predictions in deep learning models.

Before introducing the FPDA algorithm, some additional notation is required. As other deep learning models, SS models are composed of a sequence of layers $(\ell_0, \ell_1, ..., \ell_{L})$, from the input layer to the output layer, respectively.

Given an SS model $f$ evaluated on an input image $x$, $v^{\ell}_{c_\ell,i_\ell,j_\ell}(x)$ represents the activation value of a neuron at layer $\ell$, having position $(c_\ell,i_\ell,j_\ell)\in \{1,...,{C_{\ell}}\}\times \{1,...,{H_{\ell}}\}\times \{1,...,{W_{\ell}\}}$, where $C_{\ell}$, $H_{\ell}$ and $W_{\ell}$ denote the number of channels, height, and width of the features at layer $\ell$.

The pseudo-code of FPDA is reported in Algorithm \ref{alg:agg-SRC}. The function \texttt{getActivation}($f,x,\ell$) (\textit{line 1}) returns all the neuron activations at the layer $\ell$, denoted by $A_\ell = \{ v^{\ell}_{c_\ell,i_\ell,j_\ell}(x), \quad \forall c_\ell, i_\ell, j_\ell \}$.
Since the number of neurons at layer $\ell$ might be large for SS models, making the next algorithmic steps computationally expensive and therefore not suited for real-time applications, a \textit{features-compression} is performed at \textit{line 2}, returning $\mathcal{C}_\ell = \{\max_{c_\ell \in C_\ell} |v^{\ell}_{c_\ell,i_\ell,j_\ell}(x)|, \quad \forall i_\ell, j_\ell \}$. This operation preserves the properties of the over-activated neurons by using the $\max$ operator.

\newcommand{\algrule}[1][.2pt]{\par\vskip.5\baselineskip\hrule height #1\par\vskip.5\baselineskip}
\begin{algorithm}
\SetAlgoLined
 \KwIn{$f$, $x$, $\ell$, $\mu_\ell$, $\sigma_\ell$, $\theta_\nu$, $\varrho$}
 \KwOut{Detection flag}
 \algrule
 $A_\ell = \texttt{getActivation}(f, x, \ell)$ \\
 $\mathcal{C}_\ell= \{~\max_{c_\ell \in C_\ell}~ |v^{\ell}_{c_\ell,i_\ell,j_\ell}(x)|~), \quad v^{\ell}_{c_\ell,i_\ell,j_\ell}(x) \in A_\ell \}$\\
 $\hat{\mathcal{C}_\ell} = (\mathcal{C}_\ell - \mu_\ell)/\sigma_\ell$ \\
 $\mathcal{S}_\ell = \hat{\mathcal{C}_\ell}.\texttt{where}(\hat{\mathcal{C}_\ell}  > \theta_\nu)$\\
 $score = \sum{\mathcal{S}_\ell}$ \\
 \If{$score > \varrho$}{
    Return \textbf{True}
 }
 Return \textbf{False}
\caption{Fast Patch Detection Algorithm.}
\label{alg:agg-SRC}
\end{algorithm}

Then, $\mathcal{C}_\ell$ is normalized (\textit{line 3)} using the mean $\mu_\ell$ and standard deviation $\sigma_\ell$, which are computed offline from a dataset of clean images (i.e., no patched inputs). Here, $\hat{\mathcal{C}}_\ell$ denotes the normalized features.

To filter out all the common activations resulting from clean inputs, a subset of $\hat{\mathcal{C}}_\ell$ is selected (\textit{line 4}), including only those neurons with an activation larger than a \textit{selection-threshold} $\theta_\nu$ and thus deemed as unsafe. The threshold $\theta_\nu$ is determined offline as the \textit{$\nu$-percentile} value computed on the normalized features $\hat{\mathcal{C}}_\ell$ of the previously defined clean dataset.

Finally, a score value is obtained (\textit{line 5}) as the sum of these over-activated features. Such a score is then compared (\textit{line 6}) with a \textit{decision-threshold} $\varrho$ in order to decide whether the prediction $f(x)$ is safe or not (i.e., whether $x$ is genuine or includes an adversarial patch). Threshold $\varrho$ is tuned offline on a second dataset, composed of both clean and patched images.

The performance of the proposed algorithm depends on $\nu, \varrho$, and the two datasets used to extract $\mu_\ell$, $\sigma_\ell$, $\theta_\nu$, and $\varrho$. These parameters and the related tuning strategies are discussed in details in Section \ref{s:exp}. 

It is useful to highlight the main differences between the proposed method and HN. Although both achieve similar detection performance, our approach has a significantly smaller computation time thanks to the features-compression step described above. Furthermore, while the features-selection step of HN is performed at run-time on the layer activations, our approach exploits an off-line computation of a threshold $\theta_{\nu}$, computed on a given clean dataset, to reduce the run time detection latency.

Section \ref{s:exp} presents exhaustive experiments aimed at assessing the detectability of patched images and the real-time performance of the algorithm. A comparison with the original HN formulation is also provided.
%For a fair comparison, we also extended the original HN for GPU execution.
Finally, a comprehensive analysis against defense-aware attacks is shown to illustrate the strengths and weaknesses of FPDA.

\section{Experimental Evaluation} \label{s:exp}

% This section presents the set of experiments carried out to evaluate the performance of the proposed approach in comparison with other state-of-the-art SS network models as well as attack and defense approaches.
This section presents the set of experiments carried out to evaluate the performance of the proposed attack and defense approaches.
%in comparison with other state-of-the-art SS network models as well as attack and defense approaches.

First, the experimental setup is described, including the networks, the datasets, the hyperparameters, the performance metrics, and the hardware involved.
Then, the results of the untargeted (single- and double-patch) and targeted attacks are presented on Cityscapes; the effect of untargeted single- and double-patch attacks are reported for the CARLA-generated images; finally, the detection results of FPDA are showed.

% \begin{itemize}
%     \item Experimental setup
%     \item Study of the "hyperparameters"/ablation (gamma value, distance of the patch, ...)
%     \item Single-patch untargeted (specific on Carla + EOT on Cityscapes)
%     \item Multi-patch untargeted (specific on Carla + EOT on Cityscapes)
%     \item Multi-patch targeted (local attacks)
%     \item Patch defense (?)   
%     \item Real-world patch
% \end{itemize}

\subsection{Experimental Setup} \label{ss:setup}

All the experiments were performed using PyTorch~\cite{pytorch} and a set of 8 NVIDIA-A100 GPUs, while the CARLA simulator was run on a system powered by an Intel Core i7 with 12GB RAM and a GeForce GTX 1080 Ti GPU.

The optimizer of choice was Adam \cite{adam}, with learning rate empirically set to 0.5. The effect of the adversarial patches on the SS models was evaluated using the mean Intersection-over-Union (mIoU) and mean Accuracy (mAcc) \cite{minaee_image_2020} on the subset of the image pixels not belonging to the patch.

The code is available at this link: \url{https://github.com/retis-ai/SemSegAdvPatch}.

\paragraph{Datasets}
Several datasets were used for the experiments.

The Cityscapes dataset \cite{DBLP:conf/cvpr/CordtsORREBFRS16} is one of the most common dataset of driving images for semantic segmentation. It is composed of 2975 and 500 high resolution images ($1024\times2048$) for training and validation, respectively.
This dataset was used to perform single- and multi-patch attacks, both with untargeted and targeted formulation. These patches were optimized on 250 images randomly sampled from the training set, while the entire validation set was used to evaluate the effectiveness of the resulting universal patches.

Other datasets were created using the CARLA simulator by modifying the built-in Town01 map to insert some billboards that served as attackable surfaces. In particular, two different versions of three scenes (denoted by `\emph{scene1}', `\emph{scene2}', and `\emph{scene3}') were considered: one for single-patch attacks (i.e., one billboard close to the road), and one for double-patch attacks (i.e., two billboards). To mimic the setting used in Cityscapes, RGB images of size $1024\times2048$, along with their corresponding SS tags, were collected by placing a camera on-board the ego vehicle.

For each CARLA scene (three single-patch and three multi-patch), a dataset of 150 images was collected (with no patch attached) for patch optimization. These datasets contain information about the position and orientation of both the camera and the billboard, to allow computing the roto-translations and projection matrices for different points of view in the scene. Details can be found in the supplementary material.

Once optimized, the resulting patch is imported in CARLA and applied on the target billboard. Additional 100 images per scene were collected and used for the performance evaluation of the attack. Please note that, since these datasets already include a patch, the metrics are evaluated on the entire image, and therefore produce lower mIoU and mAcc values in the random case with respect to the Cityscapes dataset.

To obtain an acceptable performance of the selected networks on such CARLA datasets, it was necessary to fine-tune them. To this purpose, a training and a validation dataset (denoted by `\emph{val}') were also collected. Additional details on the fine-tuning process are reported in the supplementary material.

Finally, an additional custom dataset of real-world images was collected to optimize the real-world patch that was eventually printed. This dataset is detailed in Section \ref{ss:rw_patch}.

\paragraph{Models}
Three real-time SS models suited for autonomous driving applications were used to evaluate the adversarial attacks investigated in this paper, namely DDRNet~\cite{ddrnet_paper}, BiSeNet~\cite{BiSeNet_paper}, and  ICNet~\cite{icnet_paper}.

The three models were tested in two settings: one with Cityscapes, using the pre-trained weights provided by the authors of the models, and one on CARLA, where the models were refined with our fine-tuning procedure (further details are reported in the supplementary material).
Table~\ref{table:models_perf} summarizes the performance of these models. 

\begin{table}%[!t]
\centering
\resizebox{8.45cm}{!}{%
\begin{tabular}{|c|c|c|}
\hline
model        & \multicolumn{2}{c|}{mIoU / mAcc} \\ \hline
   & cityscapes      & CARLA (val - scene1 - scene2 -scene3)
\\ \hline
ICNet   & 0.78 ~/~0.85       & 0.70~/~0.84 - 0.53~/~0.70 - 0.64~/~0.74 - 0.62~/~0.74 \\ \hline
BiSeNet & 0.69~/~0.78       & 0.47~/~0.69 - 0.47~/~0.69 - 0.61~/~0.74 - 0.47~/~0.73      \\ \hline
DDRNet  & 0.78~/~0.85       & 0.72~/~0.88 - 0.54~/~0.74 - 0.62~/~0.76 - 0.64~/~0.78      \\ \hline
\end{tabular}
}

\caption{\small{mIoU and mAcc of the tested models on Cityscapes (pre-trained) and our CARLA dataset (fine-tuned).}}
\label{table:models_perf}
\vspace{-1em}
\end{table}

%%%%%%%%%%%%%%%%%%%%%%%%%%%%%%%%%%%%%%%%%%%%%%%%%%%

\subsection{Effects of Untargeted Attacks on Cityscapes}
\label{s:unt_cityscapes}

%{\color{red}
In this subsection, the untargeted attack is evaluated on the Cityscapes dataset using both single- and double-patch attacks.
Three patch sizes were used to test the effect of small, medium, and large patches. In particular, the sizes considered for the single-patch attack are $150\times300$, $200\times400$ and $300\times600$ pixels, while, for the double-patch attacks, we used 
$106\times212$, $141\times282$, and $212\times424$ pixels. This setting allowed to make a fair comparison between the double and single-patch formulations, since the overall area covered in each type of attack is roughly the same.
%Furthermore, in order to provide a fair comparison between the single and double-patch attacks, the patch-sizes corresponding to each patch of the latter formulation are $106\times212$, $141\times282$ and $212\times424$. In this way, the area covered by patches crafted with a double-patch attack and the single-patch attack are the same.

% The non-robust patches (without EOT) were optimized by placing them at the center of the image at each training iteration (i.e., $\eta(\cdot) =$ fixed position) and applying no appearance transformations (i.e., $\Gamma_a =\emptyset$).

%The robust optimizations with EOT apply multiple digital transformations. 
Transformation $\Gamma_a$ includes only Gaussian noise with standard deviation 5\% of the image range, whereas the patch placement function $\eta$ includes random scaling ($80\% - 120\%$ of the initial patch size) and random translation defined as follows: if $(c_x, c_y)$ is the center of the image, the position of the patch is randomized within the range $(c_x\pm \tilde{r} \cdot \tilde{W}/2~,~c_y\pm \tilde{r} \cdot \tilde{H}/2)$, where $\tilde{r} \in [0,1]$ is random variable with a uniform distribution.
The translation range was kept limited, rather than considering the full image space, to ensure better optimization stability and faster convergence.
The same transformation settings were used for the double-patch attack, with the only difference that the center $(c_x, c_y)$ corresponding to each of the two patches are the center of the left and right halves of the image.
The patches were optimized over 200 epochs.
%among 250 images of the Cityscapes training set and then evaluated on the validation set.

As shown in Table \ref{table:results_digital}, the double-patch formulation achieves, in general, higher attack performance with respect to the single patch version. In particular, for DDRNet and ICNet, the double-patch attack gets a lower mIoU in all the tested sizes. Different considerations arise for BiSeNet, where the small and medium sizes achieve better results on the single-patch attack, suggesting that the individual patch dimension are more relevant than the number of patches involved.

\begin{figure*}
     \centering
     \begin{subfigure}{\textwidth}
     \begin{subfigure}{0.19\textwidth}
         \centering
         \includegraphics[width=\textwidth]{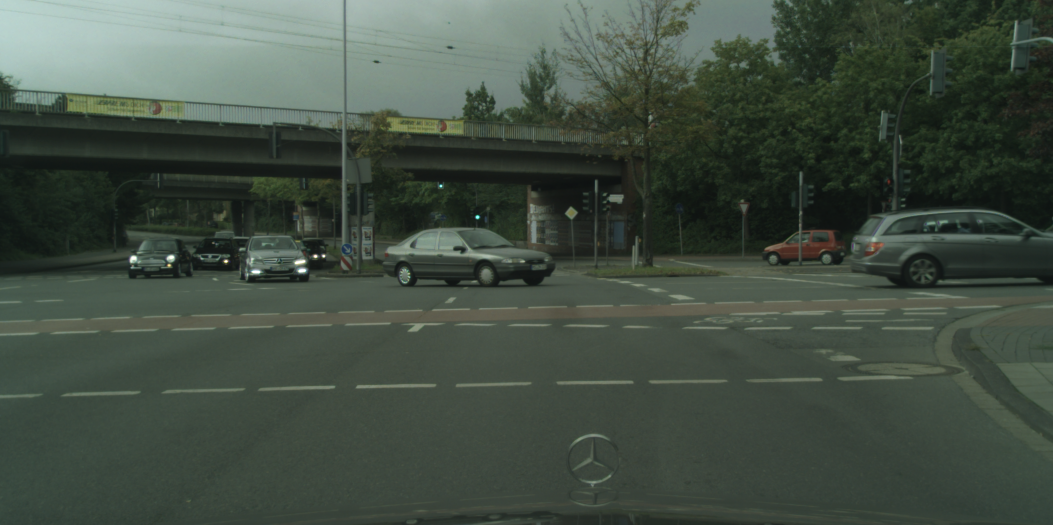}
         \vspace{-1em}
     \end{subfigure}
     \begin{subfigure}{0.19\textwidth}
         \centering
         \includegraphics[width=\textwidth, height=0.5\textwidth]{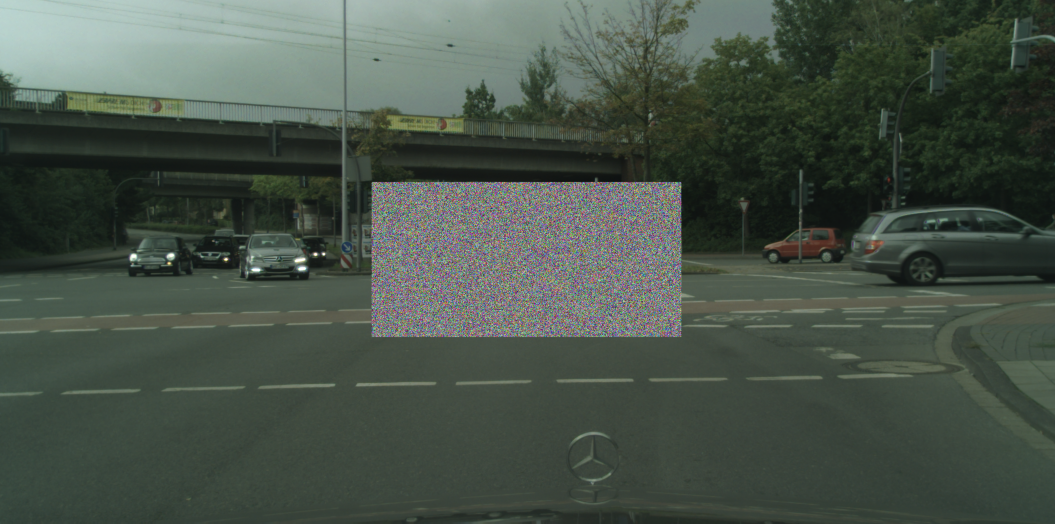}
         \vspace{-1em}
     \end{subfigure}
     \begin{subfigure}{0.19\textwidth}
         \centering
         \includegraphics[width=\textwidth, height=0.5\textwidth]{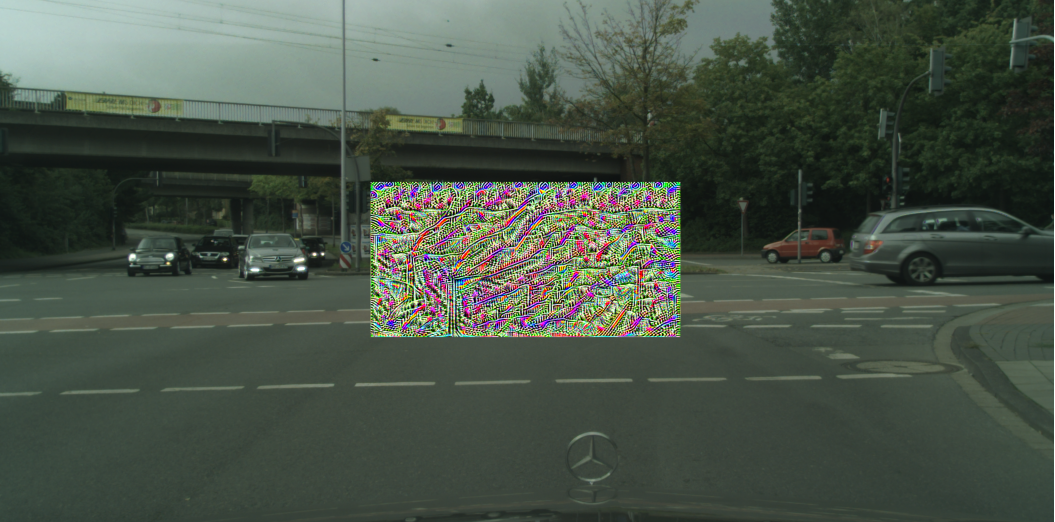}
         \vspace{-1em}
     \end{subfigure}
     \begin{subfigure}{0.19\textwidth}
         \centering
         \includegraphics[width=\textwidth, height=0.5\textwidth]{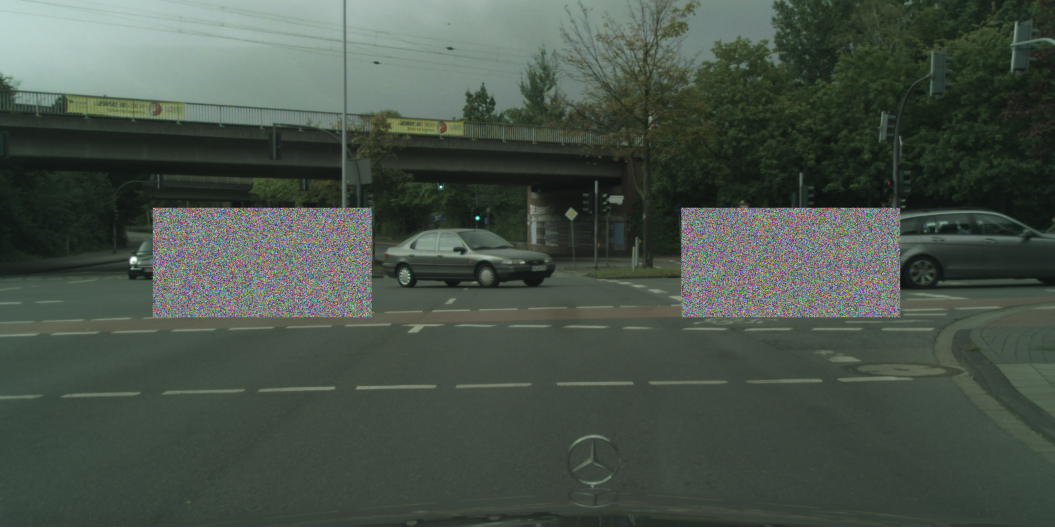}
         \vspace{-1em}
     \end{subfigure}
     \begin{subfigure}{0.19\textwidth}
         \centering
         \includegraphics[width=\textwidth, height=0.5\textwidth]{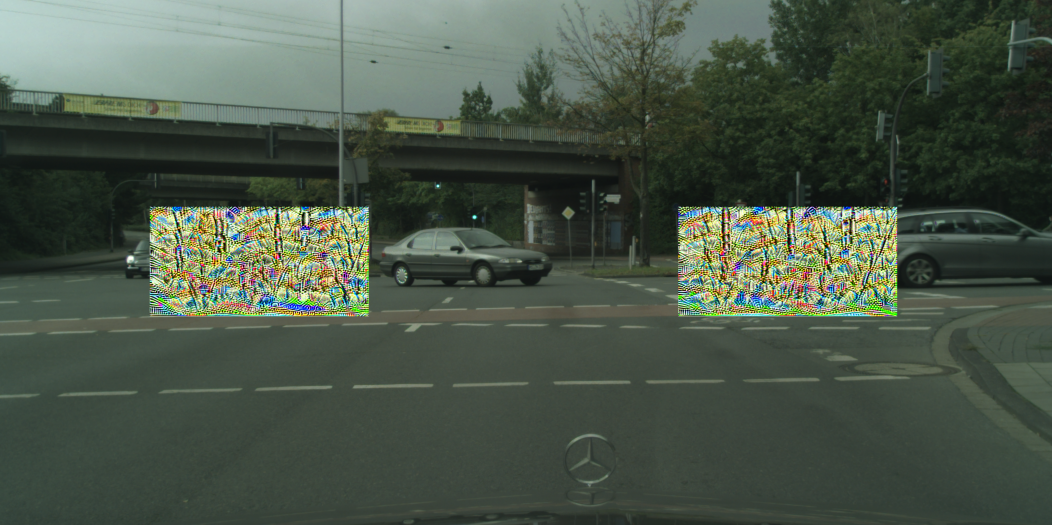}
         \vspace{-1em}
     \end{subfigure}
     \end{subfigure}

     \begin{subfigure}{\textwidth}
     \begin{subfigure}{0.19\textwidth}
         \centering
         \includegraphics[width=\textwidth]{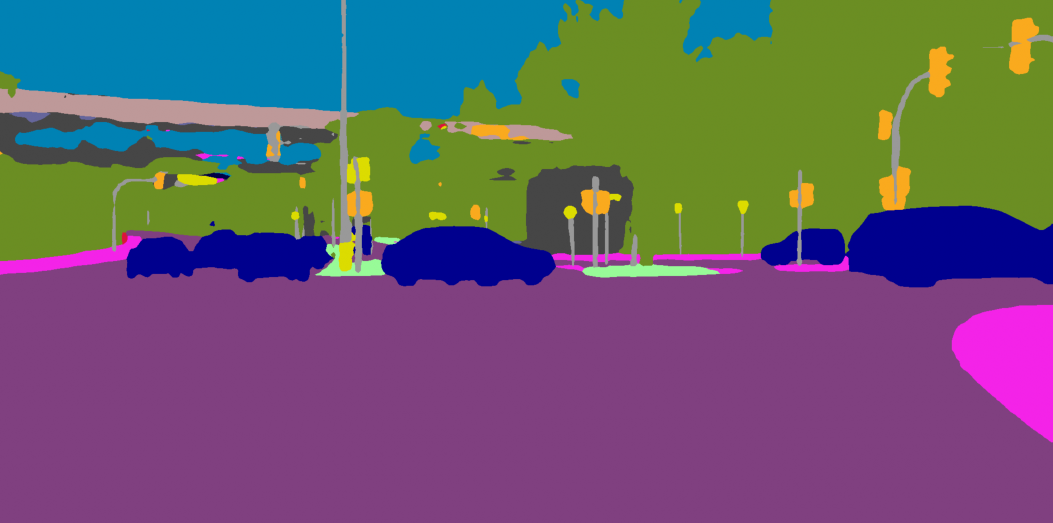}
         \vspace{-1.5em}
         \caption{}
     \end{subfigure}
     \begin{subfigure}{0.19\textwidth}
         \centering
         \includegraphics[width=\textwidth]{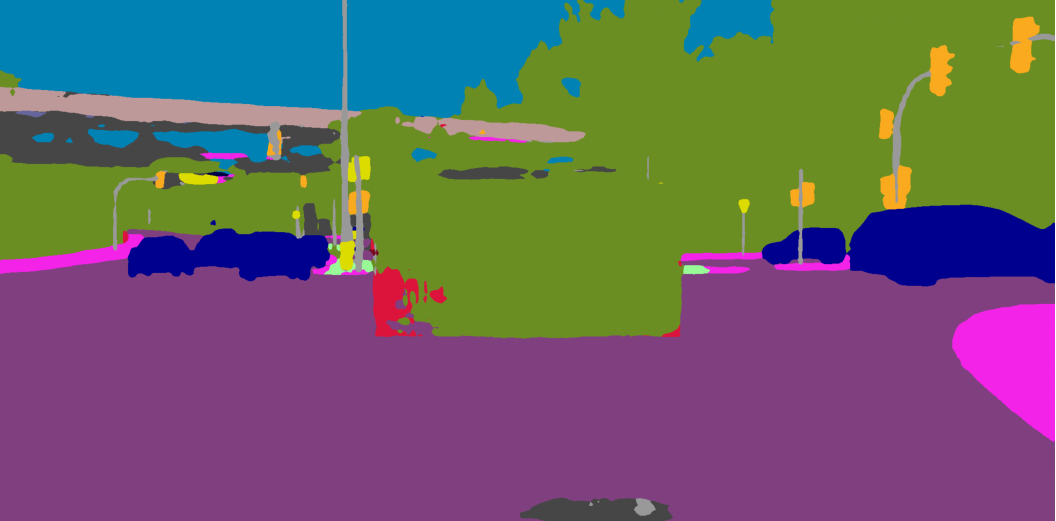}
         \vspace{-1.5em}
         \caption{}
     \end{subfigure}
     \begin{subfigure}{0.19\textwidth}
         \centering
         \includegraphics[width=\textwidth]{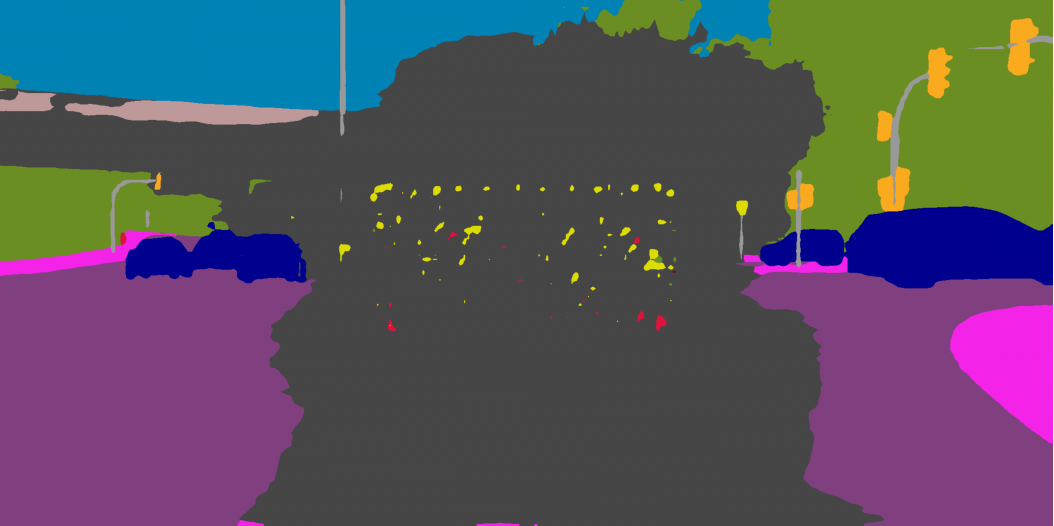}
         \vspace{-1.5em}
         \caption{}
     \end{subfigure}
     \begin{subfigure}{0.19\textwidth}
         \centering
         \includegraphics[width=\textwidth]{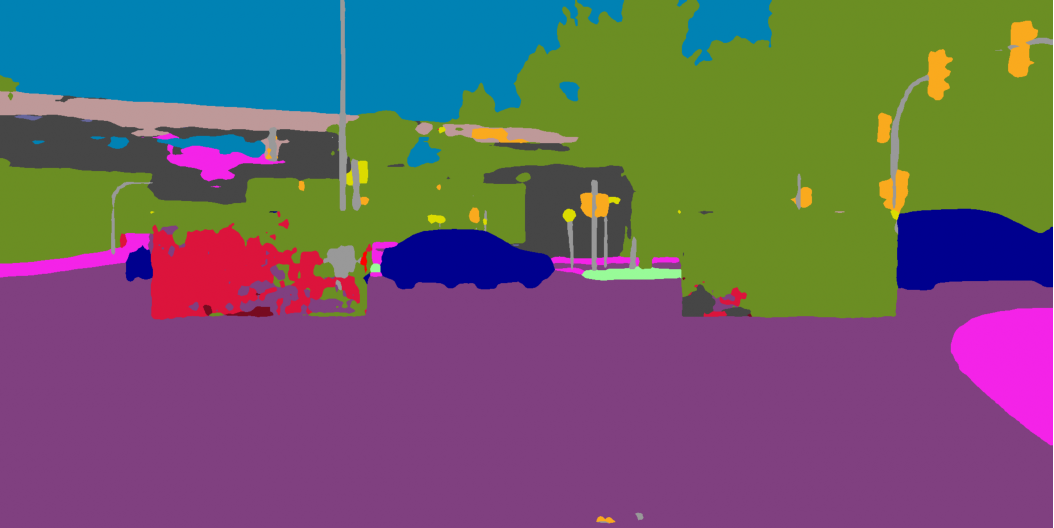}
         \vspace{-1.5em}
         \caption{}
     \end{subfigure}
     \begin{subfigure}{0.19\textwidth}
         \centering
         \includegraphics[width=\textwidth]{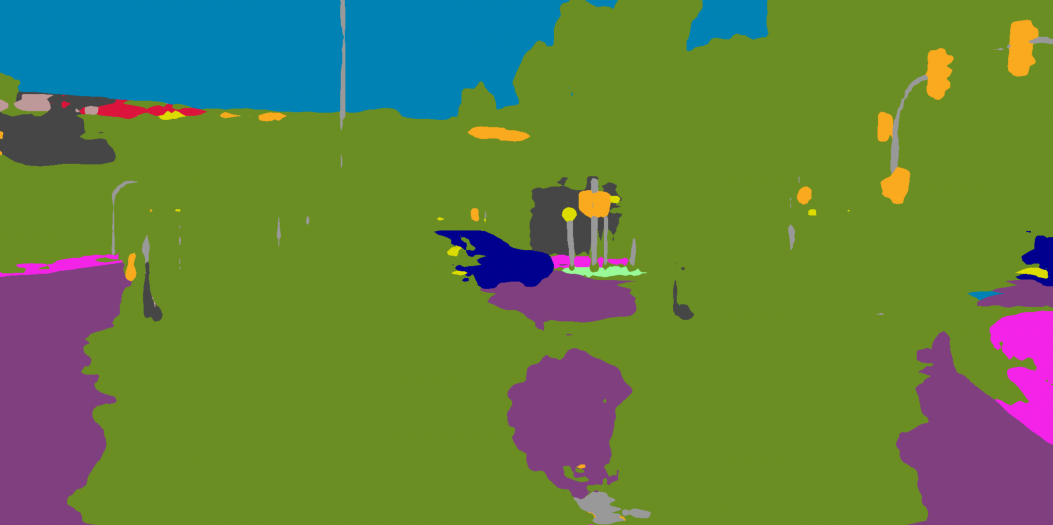}
         \vspace{-1.5em}
         \caption{}
     \end{subfigure}
     \end{subfigure}
        \vspace{-0.5em}
        \caption{\small{Images of the Cityscapes validation set and their semantic segmentations obtained from DDRNet with no patch (a) random patch ($300 \times 600$) (b), adversarial patch (c), double  random patches ($212 \times 424$) (d) and double adversarial patches  (e).}}
        \vspace{-1em}
        \label{fig:digital_exps}
\end{figure*}

Figure \ref{fig:digital_exps} shows the adversarial effect produced by the proposed attacks, illustrated for the single- ($300 \times 600$) and double-patch formulation ($212 \times 424$),  with respect to random patches. 
%}
%and a comparison between the single ($300 \times 600$) and double-patch ($212 \times 424$) attacks. The random patch, both applied one or two times, does not affect the prediction even in pixels close. The adversarial patch instead attacks portion of the image that are 

\begin{table*}[!t]
\centering
\resizebox{0.75\textwidth}{!}{
\begin{tabular}{|c|c|c|c|c|c|c|} 
\hline
\multirow{3}{*}{Model}   & \multicolumn{6}{c|}{mIoU - mAcc (rand / EOT)}                                                                                                                                                                                                                                                               \\ 
\cline{2-7}
                         & \multicolumn{2}{c|}{$150 \times 300$}                                                              & \multicolumn{2}{c|}{$200 \times 400$}                                                             & \multicolumn{2}{c|}{$300 \times 600$}                                                              \\ 
\hhline{|~------|}
                         & \multicolumn{2}{c|}{{\cellcolor[rgb]{0.937,0.937,0.937}}double - $106 \times 212$}                 & \multicolumn{2}{c|}{{\cellcolor[rgb]{0.937,0.937,0.937}}double - $141 \times 282$}                & \multicolumn{2}{c|}{{\cellcolor[rgb]{0.937,0.937,0.937}}double - $212 \times 424$}                 \\ 
\hline
\multirow{2}{*}{ICNet}   & 0.76  / 0.62                                    & 0.84 / 0.72                                      & 0.75  / 0.55                                    & 0.83  / 0.66                                    & 0.75  / 0.43                                    & 0.82  / 0.48                                     \\ 
\hhline{|~------|}
                         & {\cellcolor[rgb]{0.937,0.937,0.937}}0.74 / 0.49 & {\cellcolor[rgb]{0.937,0.937,0.937}}0.83 / 0.57  & {\cellcolor[rgb]{0.937,0.937,0.937}}0.72 / 0.38 & {\cellcolor[rgb]{0.937,0.937,0.937}}0.81 / 0.47 & {\cellcolor[rgb]{0.937,0.937,0.937}}0.68 / 0.21 & {\cellcolor[rgb]{0.937,0.937,0.937}}0.77 / 0.30  \\ 
\hline
\multirow{2}{*}{BiSeNet} & 0.67 / 0.48                                     & 0.76 / 0.63                                      & 0.67 / 0.32                                     & 0.75 / 0.46                                     & 0.65 / 0.22                                     & 0.74 / 0.34                                      \\ 
\hhline{|~------|}
                         & {\cellcolor[rgb]{0.937,0.937,0.937}}0.64 / 0.50 & {\cellcolor[rgb]{0.937,0.937,0.937}}0.72 / 0.64  & {\cellcolor[rgb]{0.937,0.937,0.937}}0.63 / 0.45 & {\cellcolor[rgb]{0.937,0.937,0.937}}0.71 / 0.57 & {\cellcolor[rgb]{0.937,0.937,0.937}}0.60 / 0.20 & {\cellcolor[rgb]{0.937,0.937,0.937}}0.68 / 0.30  \\ 
\hline
\multirow{2}{*}{DDRNet}  & 0.77 / 0.70                                     & 0.84 / 0.79                                      & 0.77 / 0.63                                     & 0.84 / 0.74                                     & 0.76 / 0.53                                     & 0.83 / 0.60                                      \\ 
\hhline{|~------|}
                         & {\cellcolor[rgb]{0.937,0.937,0.937}}0.75 / 0.60 & {\cellcolor[rgb]{0.937,0.937,0.937}}0.82 /  0.71 & {\cellcolor[rgb]{0.937,0.937,0.937}}0.73 / 0.55 & {\cellcolor[rgb]{0.937,0.937,0.937}}0.81/ 0.62  & {\cellcolor[rgb]{0.937,0.937,0.937}}0.72 / 0.35 & {\cellcolor[rgb]{0.937,0.937,0.937}}0.79 / 0.43  \\
\hline
\end{tabular}
}
\caption{\small{Adversarial patch results in mIoU and mAcc (calculated out of patches areas) extracted from the validation set of Cityscapes. Each cell reports the mIoU obtained from a random patch (no optimization) and the EOT optimization. The white rows refer to single patch attacks, while the grey rows refer to double patches having the same areas of the single configuration. }}
\label{table:results_digital}
\end{table*}

% \paragraph{Double patch case}
% TODO INSERT TABLE

%%%%%%%%%%%%%%%%%%%%%%%%%%%%%%%%%%%%%%%%%%%%%%%%%%%

\subsection{Effects of Targeted Attacks on Cityscapes}

While the objective of an untargeted attack is to induce the maximum error in the network's prediction regardless of the classes that are predicted, a targeted attack must follow a much more constrained optimization process. 

In particular, by pushing the network's prediction towards the target label $y_t$, we are asking the patch not only to change the prediction on the pixels originally belonging to the class $c_{\text{attacked}}$ to the class $c_{\text{target}}$ (which is already more complex than an untargeted attack), but also to keep all the other pixels untouched. 

This resulted to be a very difficult task, especially when considering image-agnostic attacks: the patch should generalize the targeted attack for an entire set of images that might differ largely on the distribution of the attacked class (e.g., pedestrians have different location and appearances in different images).

Previous work~\cite{nakka_indirect_2020} on targeted localized adversarial perturbations for SS models only deal with \emph{image-specific} (i.e., non image-agnostic) attacks: this means that the perturbations are tested on the same single image they are optimized on.
%(i.e., $|\mathbf{X}|=1$). 

We provide a similar analysis to understand whether there are classes that can be easily attacked with a real-world attack, further validating our loss function formulation when attacking particular classes of interest for the real-world case.

Table \ref{t:local_targeted} reports the effects of some image-specific patch attacks of interest for the real-world case. We decided to perform a double-patch attack with a total area of 600$\times$300, since it is the strongest attack we tested. In fact, we empirically found that it is very difficult to carry out these targeted attacks with a single patch. The table shows the IoU of the attacked class at the beginning and at the end of the optimization process. IoU values are averaged on 100 different attacks. Different networks show different strengths and weaknesses: in particular, DDRNet and BiSeNet are easily attackable with a \textbf{road}$\rightarrow$\textbf{sidewalk} attack, while ICNet is not. Conversely, ICNet suffers the \textbf{sidewalk}$\rightarrow$\textbf{NN} attack, while BiSeNet does not. This table also gives us an indication of whether it might be possible to perform universal attacks. In fact, if an average image-specific attack is not completely successful (i.e., IoU $\approx$ 0 on the attacked class), there is no hope that the attack will extend to an entire set of images. 

\begin{table}[] 
\centering
\resizebox{0.5\textwidth}{!}{
\begin{tabular}{|l|l|l|l|l|}  \hline
Attack &  ICNet & BiSeNet & DDRNet  \\ \hline
\textbf{road} $\rightarrow$ \textbf{sidewalk} & 0.96 / 0.81 & 0.96 / 0.07 & 0.97 / 0.00 \\
\textbf{sidewalk} $\rightarrow$ \textbf{NN}   & 0.77 / 0.12 & 0.78 / 0.47 & 0.84 / 0.08 \\
\textbf{pedestrian} $\rightarrow$ \textbf{NN} & 0.48 / 0.19 & 0.56 / 0.30 & 0.65 / 0.26 \\
\textbf{car} $\rightarrow$ \textbf{NN}        & 0.86 / 0.45 & 0.86 / 0.44 & 0.90 / 0.30 \\ \hline
\end{tabular} 
}
\caption{\small{Summary of the effectiveness of some image-specific attacks for different networks. Each cell reports the IoU of the attacked class at the start and at the end of the optimization process. IoU values are averaged on 100 different attacks.}\label{t:local_targeted}}
\end{table}

We empirically assessed this argument by performing universal targeted attacks for the attacks defined above. Table \ref{t:universal_targeted} presents the IoU of the attacked class for each attack and each network. Please note that these values are only indicative of the performance of the attack: in fact, since we are evaluating image-agnostic attacks, we are applying the same patch to each image of the validation set, which surely differ for the distribution of the pixels belonging to each class. 

\begin{figure}
     \centering
     
     \begin{subfigure}{0.49\textwidth}
     \begin{subfigure}{0.32\textwidth}
         \centering
         \includegraphics[width=\textwidth]{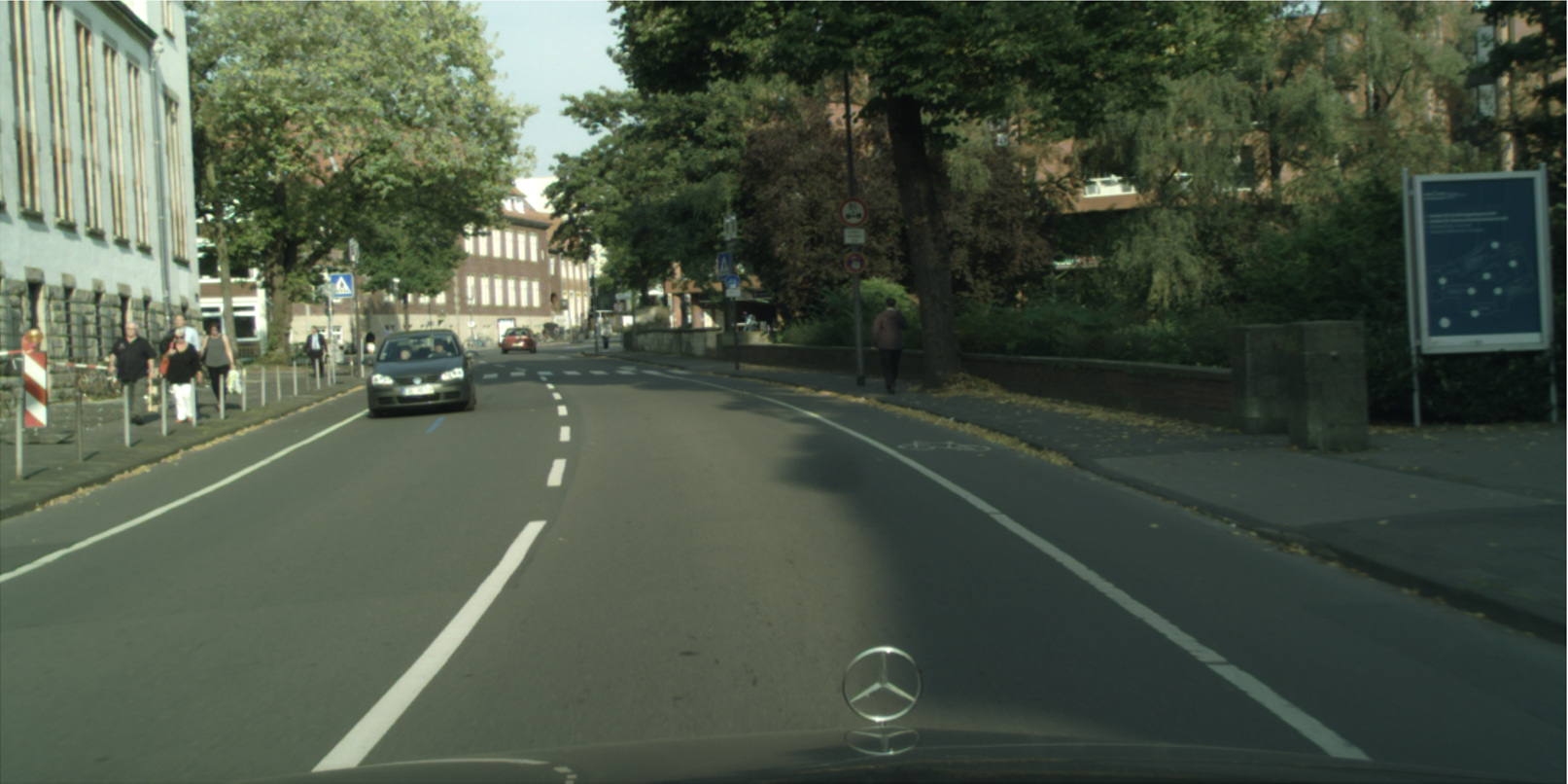}
         \vspace{-1.6em}
     \end{subfigure}
     \begin{subfigure}{0.32\textwidth}
         \centering
         \includegraphics[width=\textwidth]{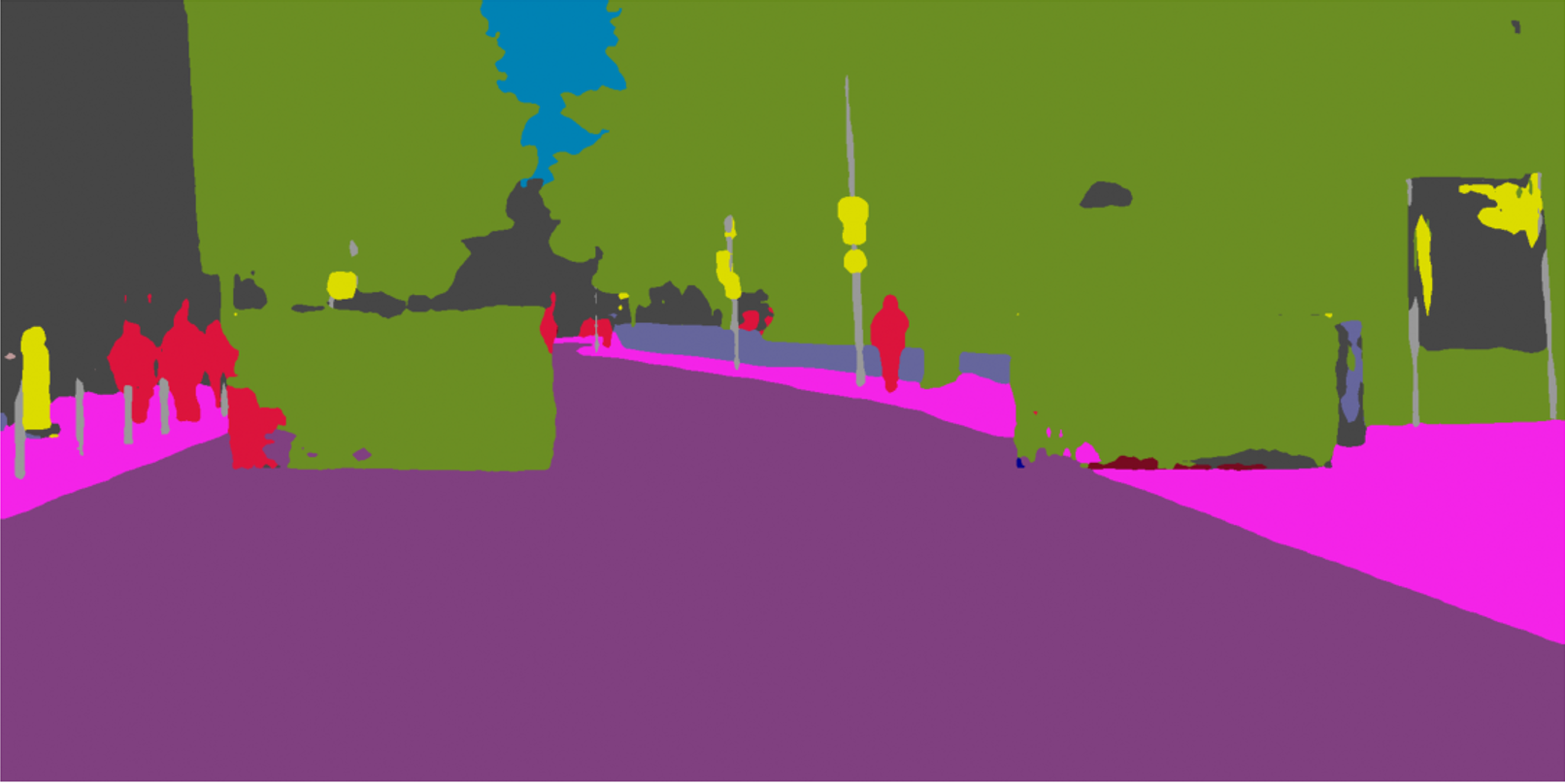}
        \vspace{-1.6em}
     \end{subfigure}
     \begin{subfigure}{0.32\textwidth}
         \centering
         \includegraphics[width=\textwidth]{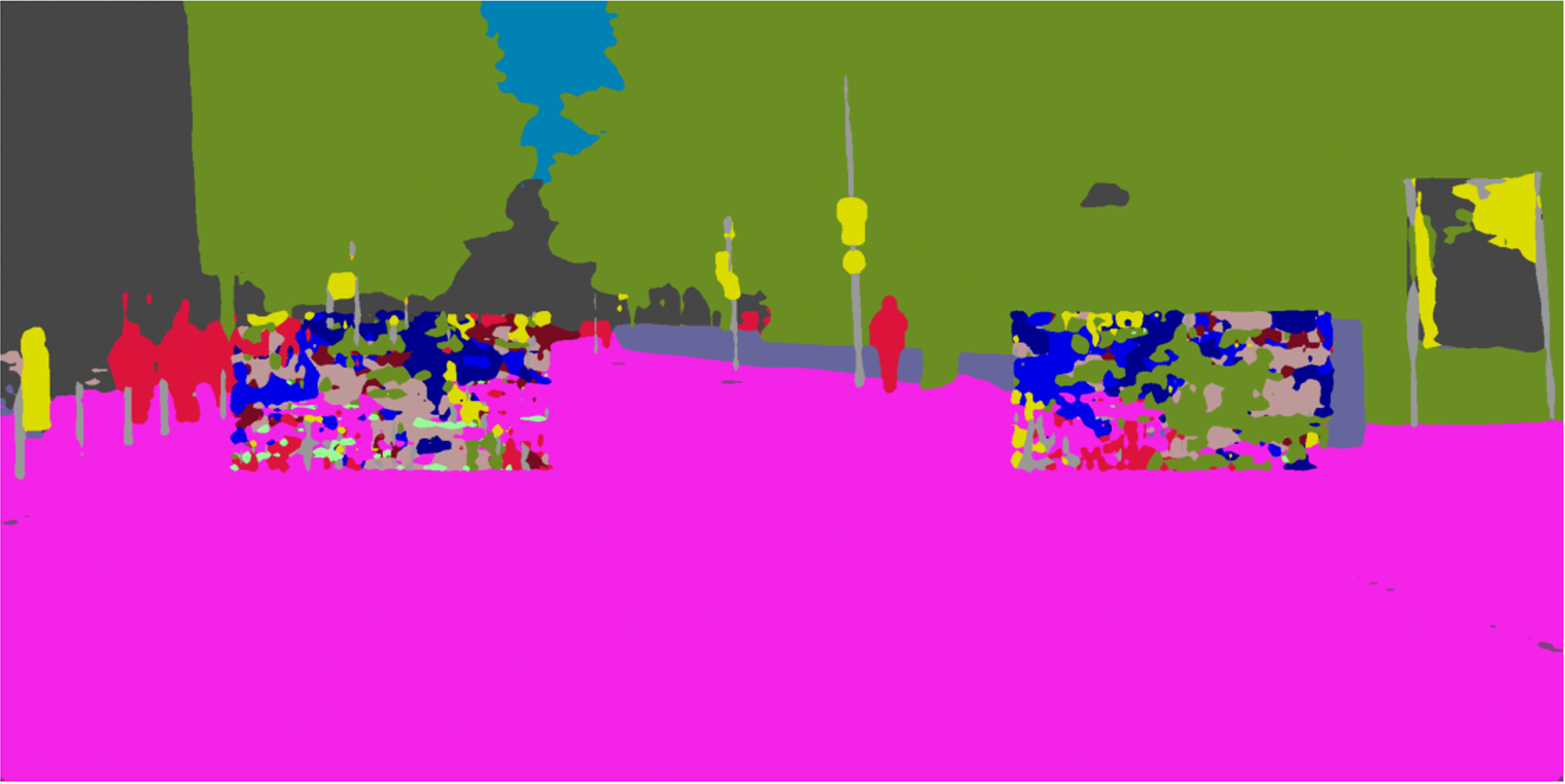}
        \vspace{-1.6em}
     \end{subfigure}
     \caption{}
     \end{subfigure}
     
     \begin{subfigure}{0.49\textwidth}
     \begin{subfigure}{0.32\textwidth}
         \centering
         \includegraphics[width=\textwidth]{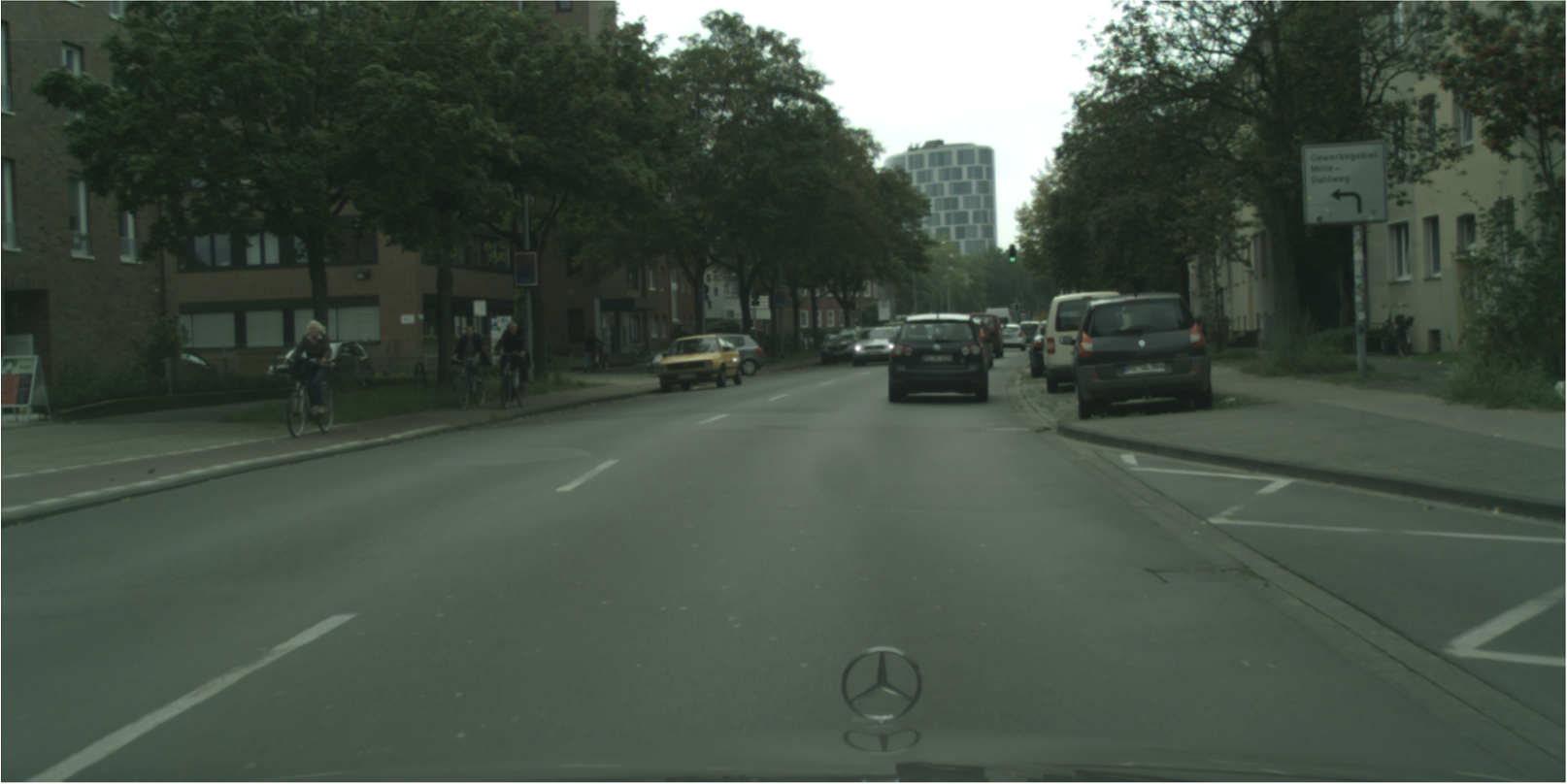}
         \vspace{-1.6em}
     \end{subfigure}
     \begin{subfigure}{0.32\textwidth}
         \centering
         \includegraphics[width=\textwidth]{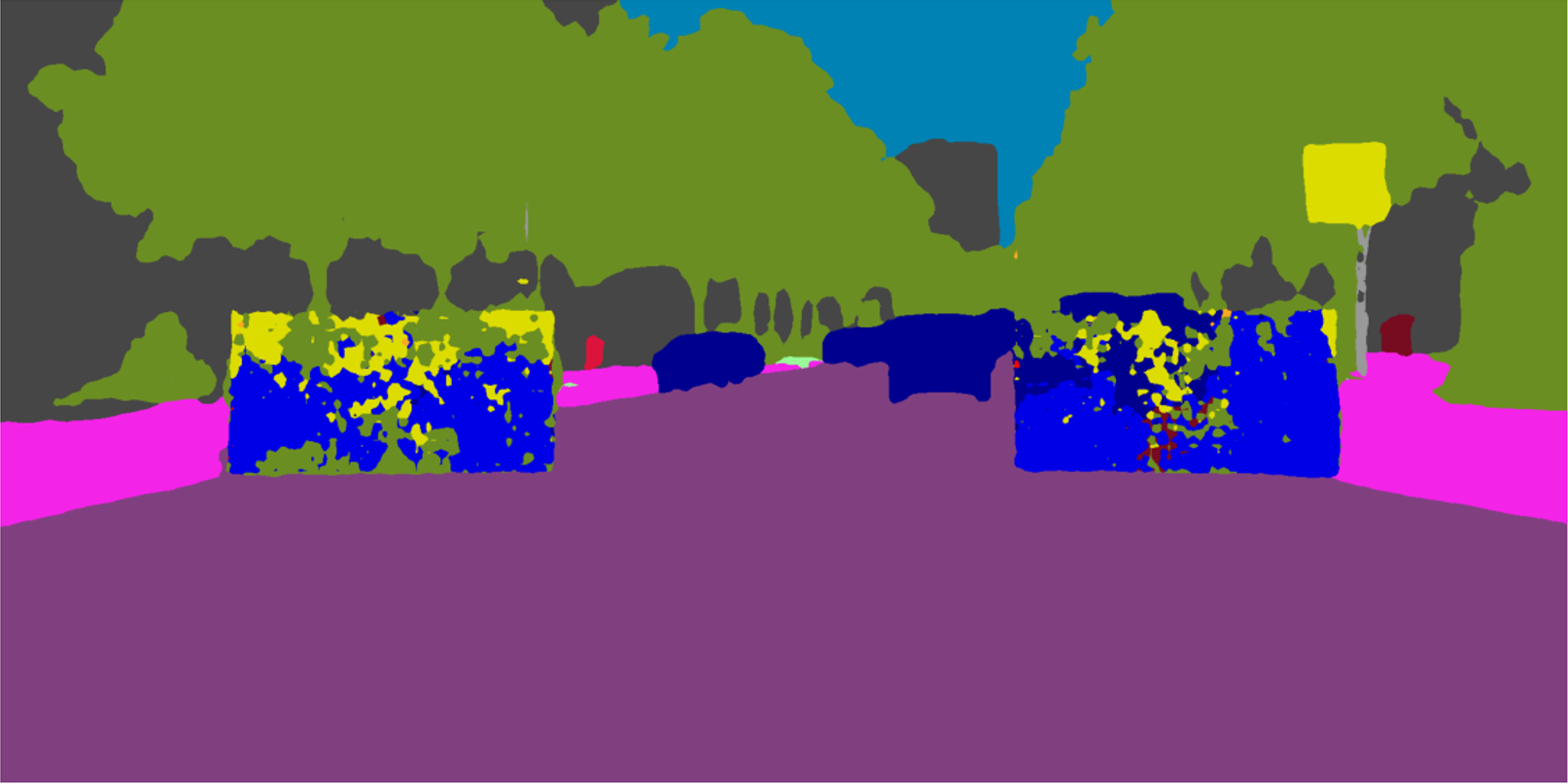}
        \vspace{-1.6em}
     \end{subfigure}
     \begin{subfigure}{0.32\textwidth}
         \centering
         \includegraphics[width=\textwidth]{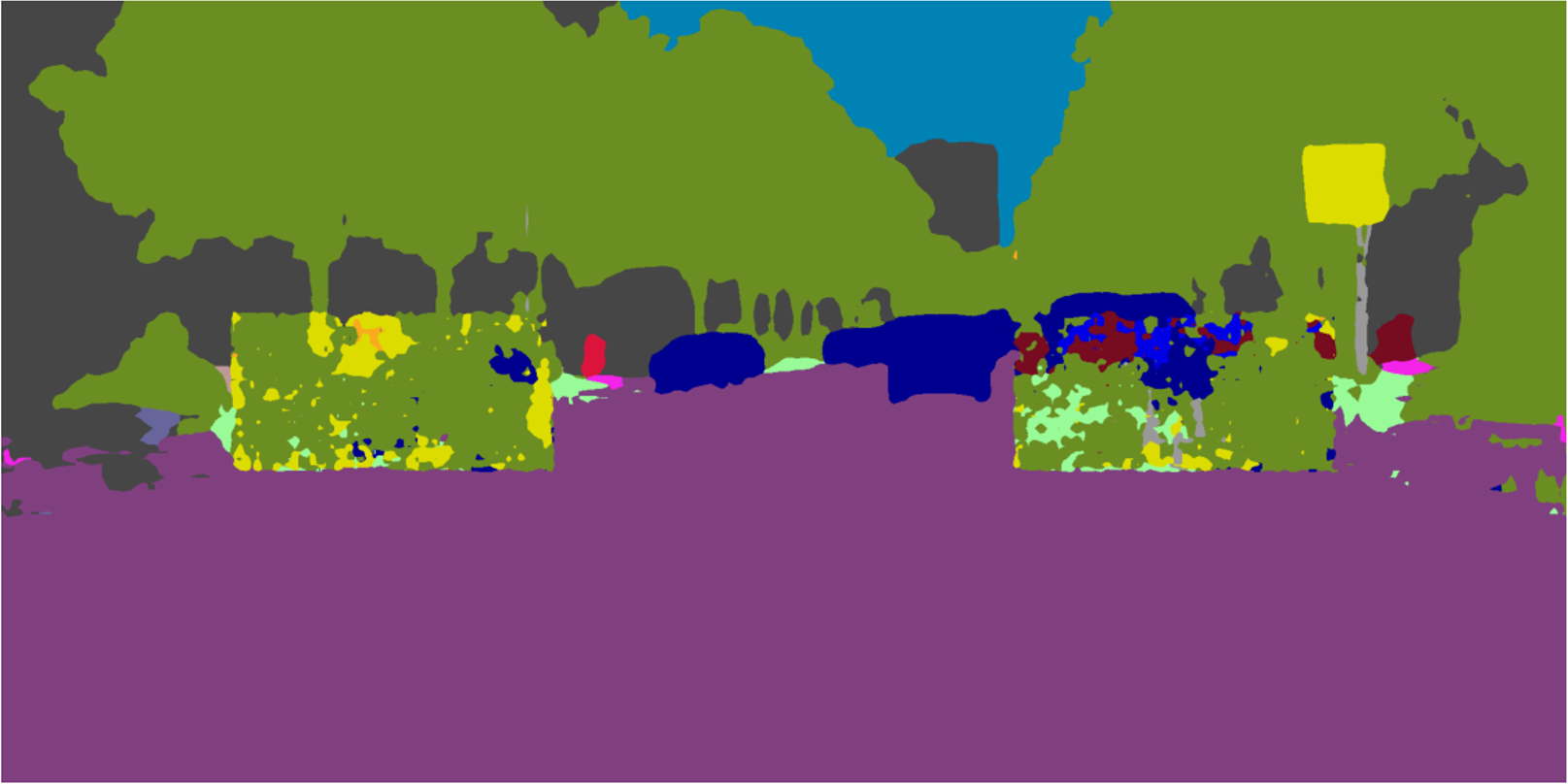}
        \vspace{-1.6em}
     \end{subfigure}
     \caption{}
     \end{subfigure}
     
     \begin{subfigure}{0.49\textwidth}
     \begin{subfigure}{0.32\textwidth}
         \centering
         \includegraphics[width=\textwidth]{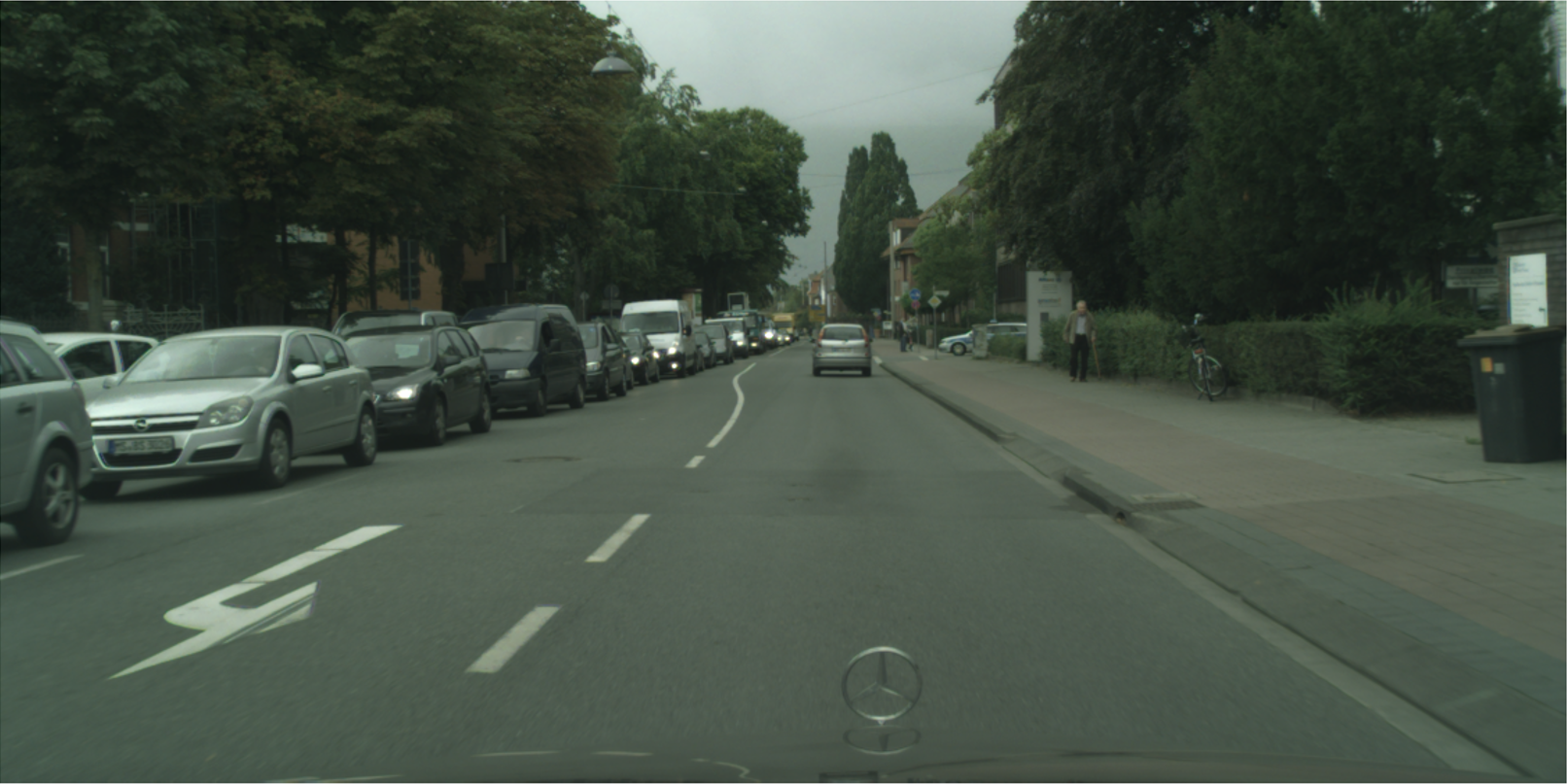}
         \vspace{-1.6em}
     \end{subfigure}
     \begin{subfigure}{0.32\textwidth}
         \centering
         \includegraphics[width=\textwidth]{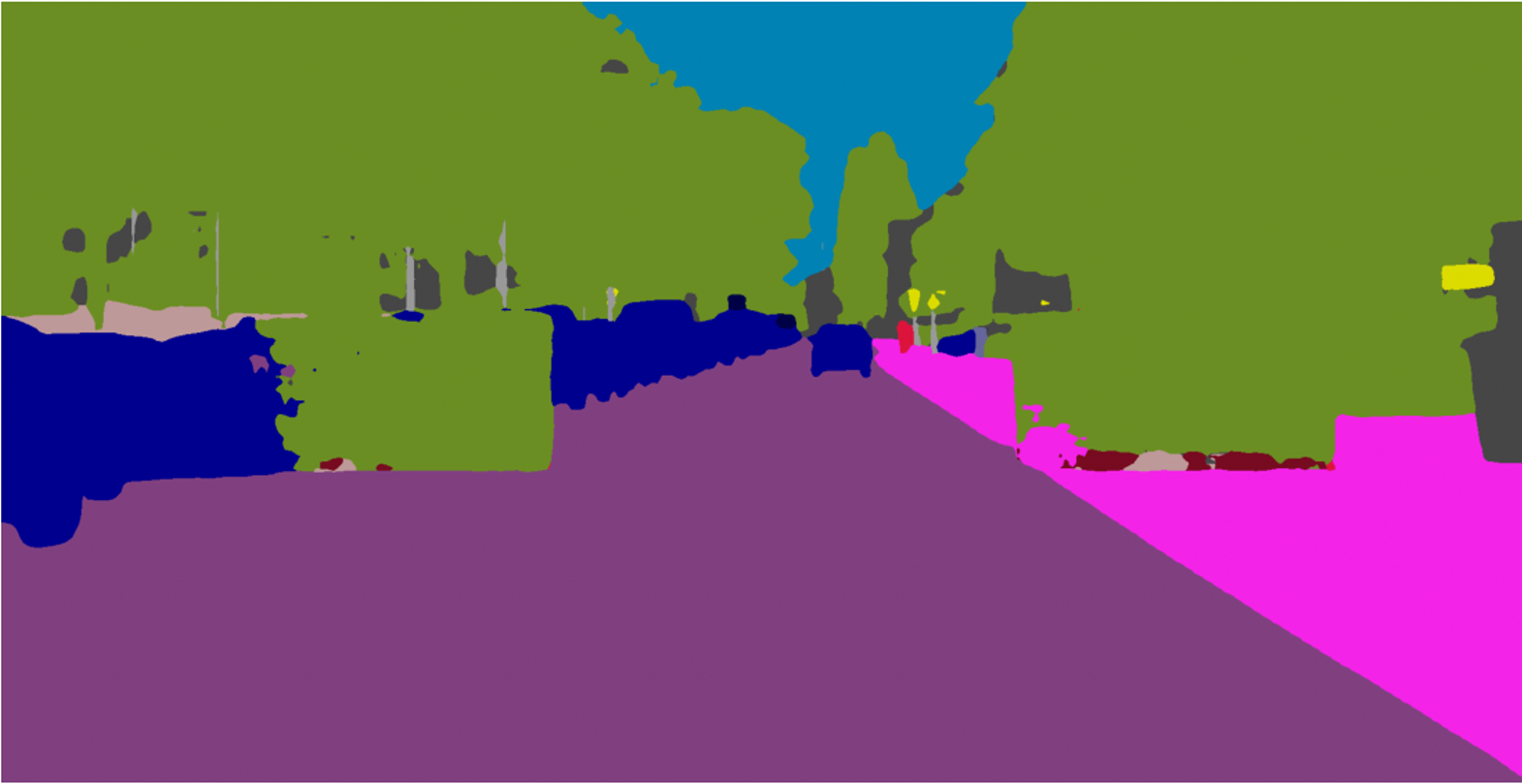}
        \vspace{-1.6em}
     \end{subfigure}
     \begin{subfigure}{0.32\textwidth}
         \centering
         \includegraphics[width=\textwidth]{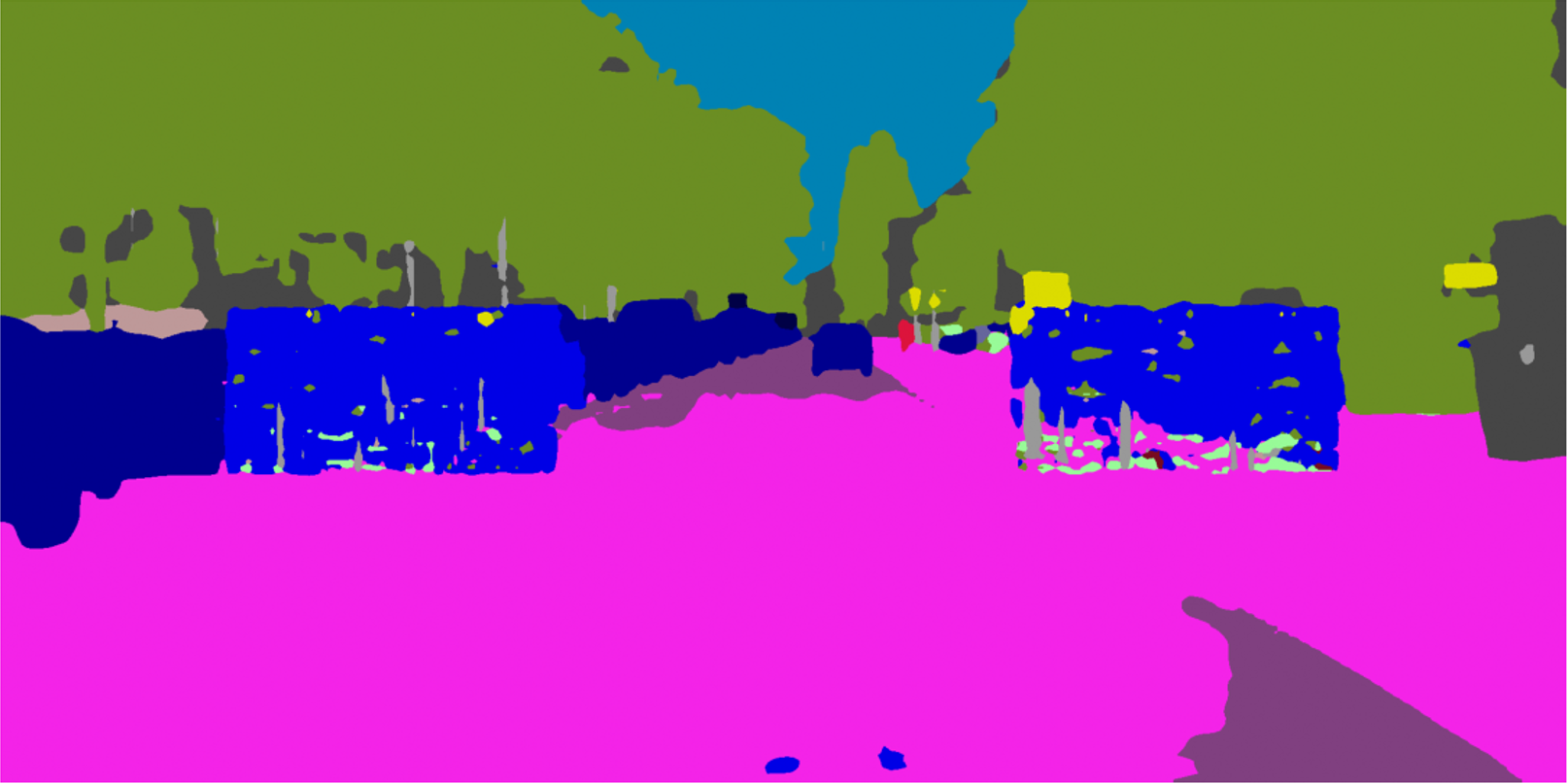}
        \vspace{-1.6em}
     \end{subfigure}
     \caption{}
     \end{subfigure}
    
    \caption{\small{Effect of targeted adversarial patches: (a) an image-specific \textbf{road}$\rightarrow$\textbf{sidewalk} attack on DDRNet, (b) an image-specific \textbf{sidewalk}$\rightarrow$\textbf{NN} attack on ICNet, (c) a universal \textbf{road}$\rightarrow$\textbf{sidewalk} attack against DDRNet. First column is the original image, middle column is the prediction with random patches, right column is the prediction with adversarial patches.}}
    \vspace{-1em}
    \label{fig:targeted_exp}
\end{figure}

These experiments lead to the conclusion that each network shows robustness for different classes. The only universal targeted attack that shows good performance is \textbf{road} $\rightarrow$ \textbf{sidewalk} on DDRNet. The others tested attacks resulted less effective.
Some of the attacks performed during the evaluation are illustrated in Figure \ref{fig:targeted_exp}. 

\begin{table} 
\centering
\resizebox{0.5\textwidth}{!}{
\begin{tabular}{|l|l|l|l|l|} \hline
Attack &  ICNet & BiSeNet & DDRNet  \\ \hline
\textbf{road} $\rightarrow$ \textbf{sidewalk} & 0.98 / 0.94 & 0.98 / 0.37 & 0.98 / 0.12 \\
\textbf{sidewalk} $\rightarrow$ \textbf{NN}   & 0.84 / 0.71 & 0.85 / 0.83 & 0.90 / 0.52 \\
\textbf{pedestrian} $\rightarrow$ \textbf{NN} & 0.76 / 0.73 & 0.85 / 0.70 & 0.89 / 0.88 \\ \hline
\end{tabular}
}
\caption{\small{Summary of the effectiveness of the selected universal targeted attacks. Each cell reports the IoU of the attacked class at the start and at the end of the optimization process.}}\label{t:universal_targeted}
\end{table}

%%%%%%%%%%%%%%%%%%%%%%%%%%%%%%%%%%%%%%%%%%%%%%%%%%%

\subsection{Effects of Untargeted Attacks on CARLA}
%{\color{red} 
In this subsection, the scene-specific attack is evaluated against the EOT-based attack using single- and double-patch attacks in CARLA.

As explained in Section \ref{ss:specific}, additional information on the relative pose of the camera and the billboards are extracted from CARLA together with the corresponding images and then used to apply accurate patch warping transformations during the optimization process, to account for different points of view of the same urban scene.

Based on preliminary experiments on the effect of the number of pixels and on the real-world dimension of the patch (described in the supplementary material), we decided to use patches of 150$\times$300 pixels (corresponding to a real-world dimension of 3.75m$\times$7.5m). For all the following experiments, $\Gamma_a$ includes contrast and brightness changes (both randomized within $\pm$10\% of the image range) and Gaussian noise (with standard deviation equal to 10\% of the image range).

Since this paper investigates the effect of real-world adversarial patches, the mIoU and mAcc scores are evaluated on an additional dataset for each tested network: this time, the patch is not digitally projected, but rather is imported in CARLA and applied to a virtual billboard as a \emph{decal} object. Hence, the resulting dataset includes images of the billboards with an adversarial patch already applied and rendered with the same level of graphic detail. This allows simulating real-world adversarial examples in CARLA.
%\TODO{Introduce the 4 reference scenes (val, scene1, scene2, scene3).} \textcolor{blue}{As described earlier, each virtual scene includes 1 or 2 billboards (according to the number of patches considered). LE SCENE SONO GIA' INTRODOTTE IN 4.1. SERVE DIRE ALTRO?}

The third column of Table~\ref{table:models_perf} reports the performance of the tested networks on the considered scenes (with no patch), while Table~\ref{table:results_carla} reports the corresponding adversarial effect in terms of mIoU and mAcc scores for both single- and double-patch attacks. 
%\TODO{why 'val' is not reported in this table?} \textcolor{blue}{'val' is only the validation set used during fine-tuning. The attacks are performed only on the scene-x datasets.}

Note that, for each setting reported in Table~\ref{table:results_carla}, the scene-specific attack outperforms the standard EOT formulation. For single-patch attacks, the performance of the two approaches is comparable and their adversarial effect is marginal. While for the Cityscapes dataset we considered double-patch and single-patch attacks with the same total patch area, in this case, double-patch attacks use total areas two times larger than those for single-patch attacks (i.e., we used two patches with the same size 3.75m$\times$7.5m). This choice was due to the poor performance of single-patch attacks.
%\TODO{Unclear sentence, please rephrase: Since for the single-patch case the performance is often comparable and the attacks are not very effective, we decided to perform double-patch attacks without keeping the same total patch area (i.e., we are using two patches with the same 3.75m$\times$7.5m area each)}. \textcolor{blue}{riformulata sopra.}
In the double-patch case, the performance of the attack largely improves, as well as the difference between the performance of the scene-specific and the EOT-based attacks.

It is worth observing that the performance of the tested networks on CARLA scenes is much worse than the one related to Cityscapes: this is partly due to the differences in evaluating the performance of the networks for the Cityscapes dataset and for the CARLA-generated datasets. For Cityscapes, the area corresponding to the patch is not considered during the evaluation: this helps ignore parts of the image that are wrongly predicted as occluded by the patch. Conversely, for CARLA-generated images, we decided to take into account also the patch area, since the patch is already present in the 3D virtual scene from CARLA. %Hence, retrieving the exact position of its pixels increases the complexity of the evaluation.

\begin{figure*}
     \centering
     \begin{subfigure}{0.19\textwidth}
         \centering
         \includegraphics[width=\textwidth]{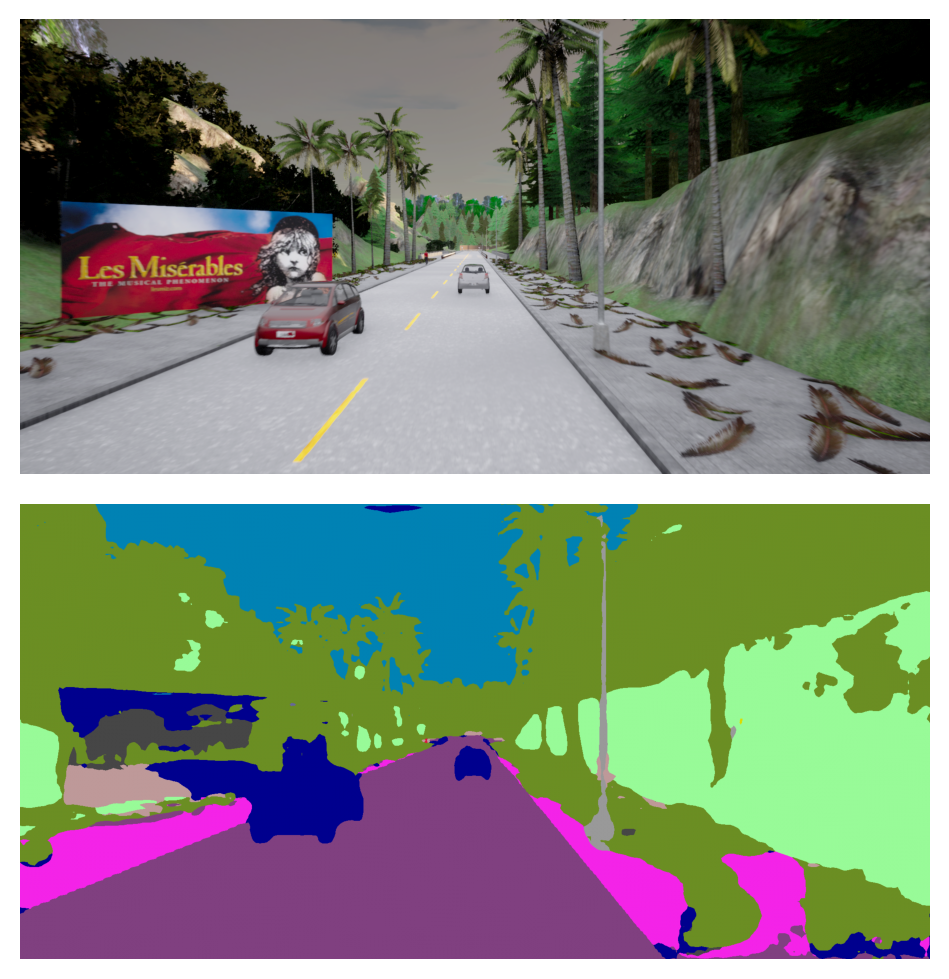}
         \vspace{-1.5em}
         \caption{}
     \end{subfigure}
     \begin{subfigure}{0.19\textwidth}
         \centering
         \includegraphics[width=\textwidth]{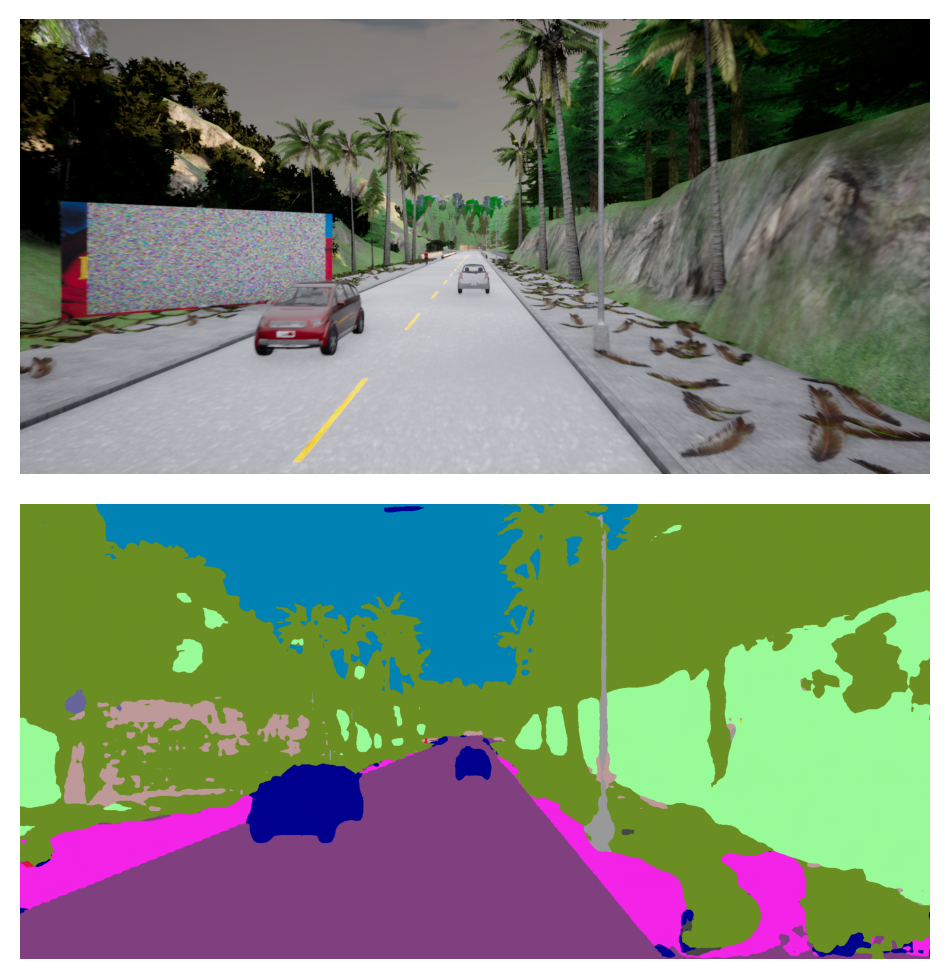}
         \vspace{-1.5em}
         \caption{}
     \end{subfigure}
     \begin{subfigure}{0.19\textwidth}
         \centering
         \includegraphics[width=\textwidth]{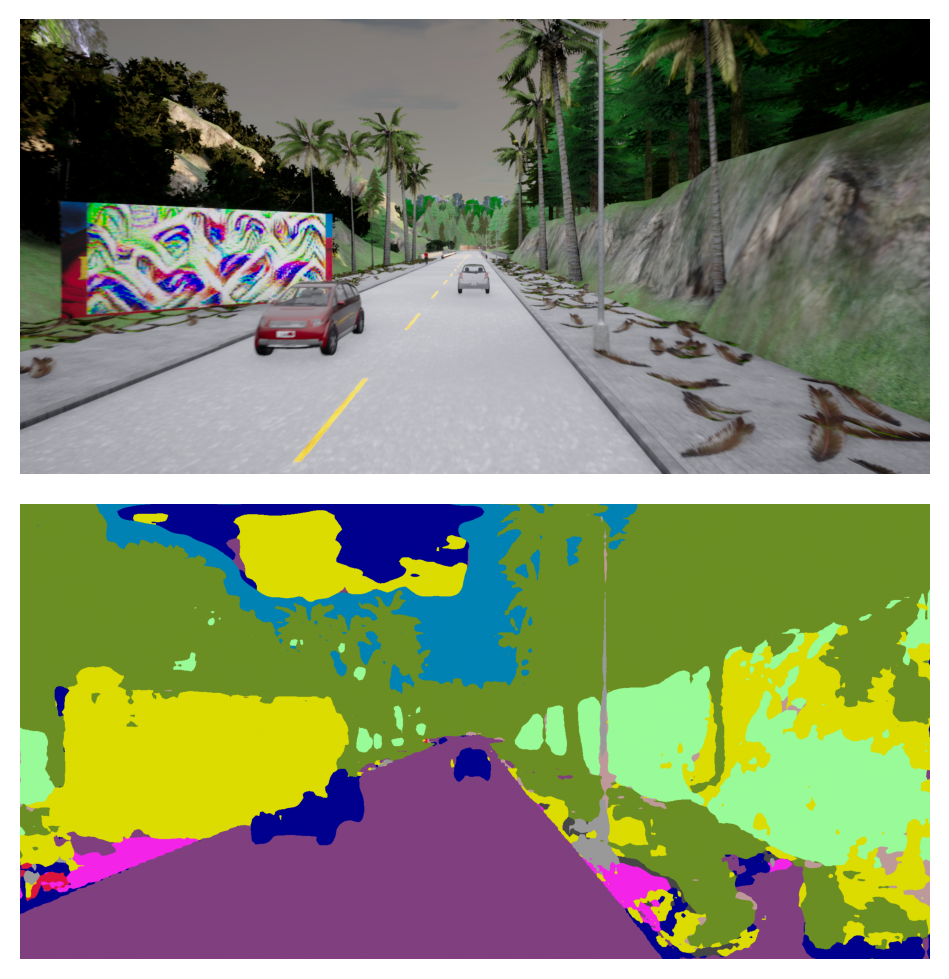}
         \vspace{-1.5em}
         \caption{}
     \end{subfigure}
     \begin{subfigure}{0.19\textwidth}
         \centering
         \includegraphics[width=\textwidth]{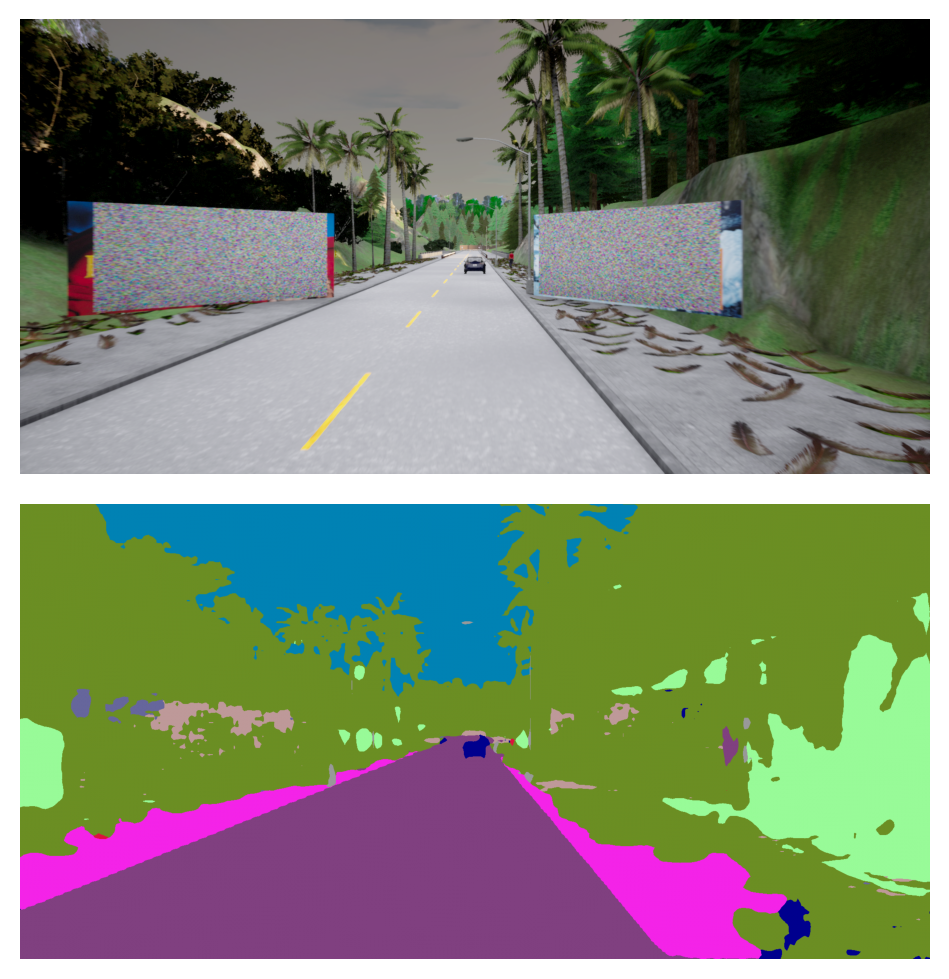}
         \vspace{-1.5em}
         \caption{}
     \end{subfigure}
     \begin{subfigure}{0.19\textwidth}
         \centering
         \includegraphics[width=\textwidth]{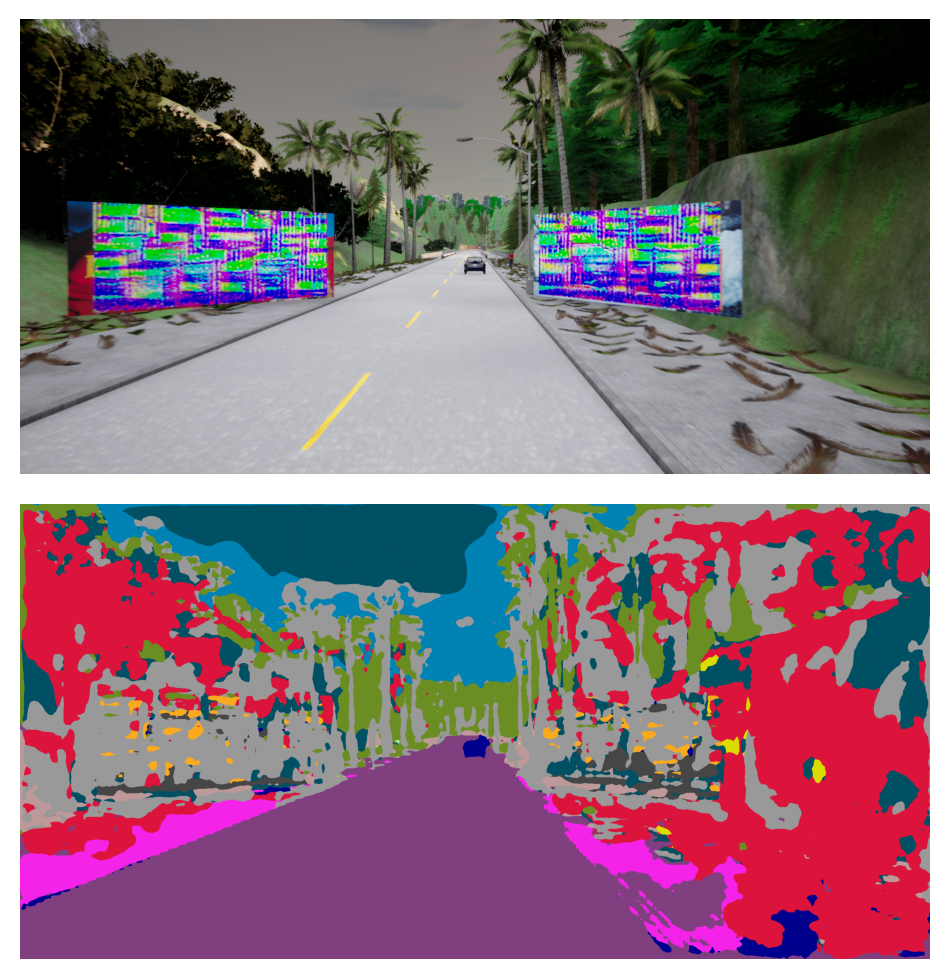}
         \vspace{-1.5em}
         \caption{}
     \end{subfigure}
        \vspace{-0.5em}
        \caption{\small{Semantic segmentations obtained with BiSeNet on CARLA scene-1 with no patch (a), single random patch (b), single scene-specific patch (c) , double random patches (d), and double scene-specific patches (e).}}
        \vspace{-1em}
        \label{fig:carla_exps}
\end{figure*}

Figure~\ref{fig:carla_exps} shows the effect of some representative attacks on CARLA images.
%}

\begin{table*}
\centering
\resizebox{\textwidth}{!}{
\begin{tabular}{|c|c|l|c|l|c|l|} 
\hline
Model                    & \multicolumn{6}{c|}{mIoU  \textbar{}  mAcc (rand / EOT / scene-specific)}                                                                                                                                                                                                                                                                            \\ 
\hline
                         & \multicolumn{2}{c|}{Scene1}                                                                                     & \multicolumn{2}{c|}{Scene2}                                                                                     & \multicolumn{2}{c|}{Scene3}                                                                                      \\ 
\hline
\multirow{2}{*}{ICNet}   & 0.51 / 0.49 / 0.48                                     & 0.60 /  0.56 / 0.54                                    & 0.64 / 0.61 / 0.61                                     & 0.74 / 0.73 / 0.73                                     & 0.63 / 0.59 / 0.59                                     & 0.76 / 0.73 / 0.74                                      \\ 
\hhline{|~------|}
                         & {\cellcolor[rgb]{0.937,0.937,0.937}}0.43 / 0.43 / 0.39 & {\cellcolor[rgb]{0.937,0.937,0.937}}0.56 / 0.55 / 0.50 & {\cellcolor[rgb]{0.937,0.937,0.937}}0.58 / 0.57 / 0.55 & {\cellcolor[rgb]{0.937,0.937,0.937}}0.66 / 0.67 / 0.67 & {\cellcolor[rgb]{0.937,0.937,0.937}}0.64 / 0.61 / 0.54 & {\cellcolor[rgb]{0.937,0.937,0.937}}0.76 / 0.73 / 0.68  \\ 
\hline
\multirow{2}{*}{BiSeNet} & 0.44 / 0.36 / 0.31                                     & 0.63 / 0.55 / 0.49                                     & 0.60 / 0.58 / 0.58                                     & 0.76 / 0.74 / 0.74                                     & 0.47 / 0.46 / 0.45                                     & 0.74 / 0.73 / 0.73                                      \\ 
\hhline{|~------|}
                         & {\cellcolor[rgb]{0.937,0.937,0.937}}0.39 / 0.37 / 0.23 & {\cellcolor[rgb]{0.937,0.937,0.937}}0.88 / 0.54 / 0.53 & {\cellcolor[rgb]{0.937,0.937,0.937}}0.55 / 0.54 / 0.53 & {\cellcolor[rgb]{0.937,0.937,0.937}}0.75 / 0.72 / 0.70 & {\cellcolor[rgb]{0.937,0.937,0.937}}0.44 / 0.43 / 0.42 & {\cellcolor[rgb]{0.937,0.937,0.937}}0.74 / 0.67 / 0.62  \\ 
\hline
\multirow{2}{*}{DDRNet}  & 0.51 / 0.46 / 0.46                                     & 0.70 / 0.69 / 0.69                                     & 0.62 / 0.52 / 0.49                                     & 0.76 / 0.71 / 0.66                                     & 0.65 / 0.58 / 0.59                                     & 0.78 / 0.76 / 0.76                                      \\ 
\hhline{|~------|}
                         & {\cellcolor[rgb]{0.937,0.937,0.937}}0.48 / 0.48 / 0.39 & {\cellcolor[rgb]{0.937,0.937,0.937}}0.67 / 0.66 / 0.66 & {\cellcolor[rgb]{0.937,0.937,0.937}}0.58 / 0.57 / 0.47 & {\cellcolor[rgb]{0.937,0.937,0.937}}0.75 / 0.74 / 0.69 & {\cellcolor[rgb]{0.937,0.937,0.937}}0.66 / 0.66 / 0.58 & {\cellcolor[rgb]{0.937,0.937,0.937}}0.79 / 0.78 / 0.76  \\
\hline
\end{tabular}
}
%\vspace{-0.5em}
\caption{\small{Adversarial patch results on the three scene CARLA datasets. The Table reports the mIoU and mAcc obtained with random, EOT-based and scene-specific patches. For each network, the first row indicates results for single-patch attacks, whereas the second row reports results for double-patch attacks.}}
\vspace{-0.5em}
\label{table:results_carla}
\end{table*}

% \paragraph{Double patch case}
% TODO INSERT TABLE

%%%%%%%%%%%%%%%%%%%%%%%%%%%%%%%%%%%%%%%%%%%%%%%%%%%

\subsection{Evaluating the proposed loss function formulation and parameters}

%{\color{red} 
The proposed loss function formulation presented in Section~\ref{ss:loss} was evaluated against the standard cross-entropy loss for several values of $\gamma$ and for the different attacks considered in the paper.

\begin{figure*}
\centering
\makebox[\columnwidth]{\includegraphics[scale=0.38]{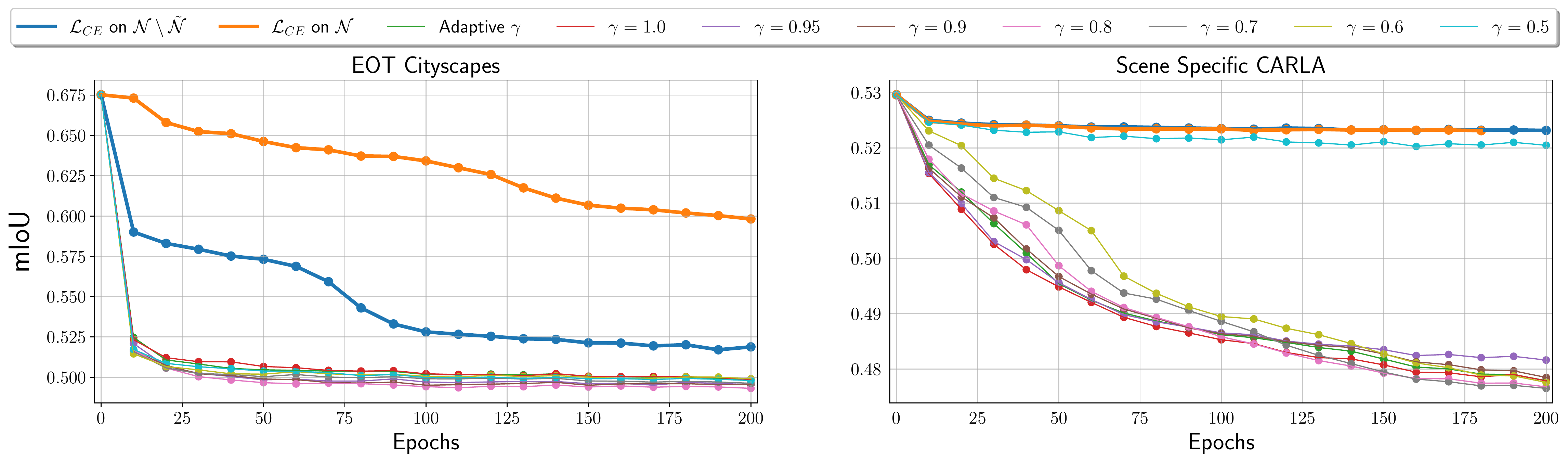}}
\vspace{-1em}
\caption{\small{Comparison of adversarial patch optimizations ($200 \times 400$) on ICNet and Cityscapes using different loss functions: two versions of the standard pixel-wise cross-entropy and our formulation with multiple values of $\gamma$. 
%The loss function used by %\cite{bar_vulnerability_2021, metzen_universal_2017, nakka_indirect_2019} 
%\cite{nakka_indirect_2019} 
%is the orange one ($\mathcal{L}_{CE}$ on $\mathcal{N}$), while the blue one is an our improved version that still %consider the common $\mathcal{L}_{CE}$.  
$\mathcal{L}_{CE}$ on $\mathcal{N}$ is the original version used by \cite{nakka_indirect_2019}, while $\mathcal{L}_{CE}$ on $\mathcal{N} \setminus \mathcal{\tilde{N}}$ is an improved version based on the rationale presented in Section \ref{ss:loss}.
%\TODO{Edit legend: $\gamma=1.0$, $\gamma=0.95$, ... .
%What's the third item of the legend referring to? Adaptive $\gamma$?} {\color{blue} [SOLVED]}
}}  
%}
\label{fig:digital_gamma_ablation}
\end{figure*}

Figure~\ref{fig:digital_gamma_ablation} shows the evolution of the mIoU score during the optimization process for the untargeted EOT-based attack on Cityscapes and the scene-specific attack on CARLA. In both cases, for each tested value of $\gamma$, our loss formulation outperforms the standard cross-entropy, both in terms of attack performance and convergence rate.

\begin{figure}
\centering
\makebox[\columnwidth]{\includegraphics[scale=0.34]{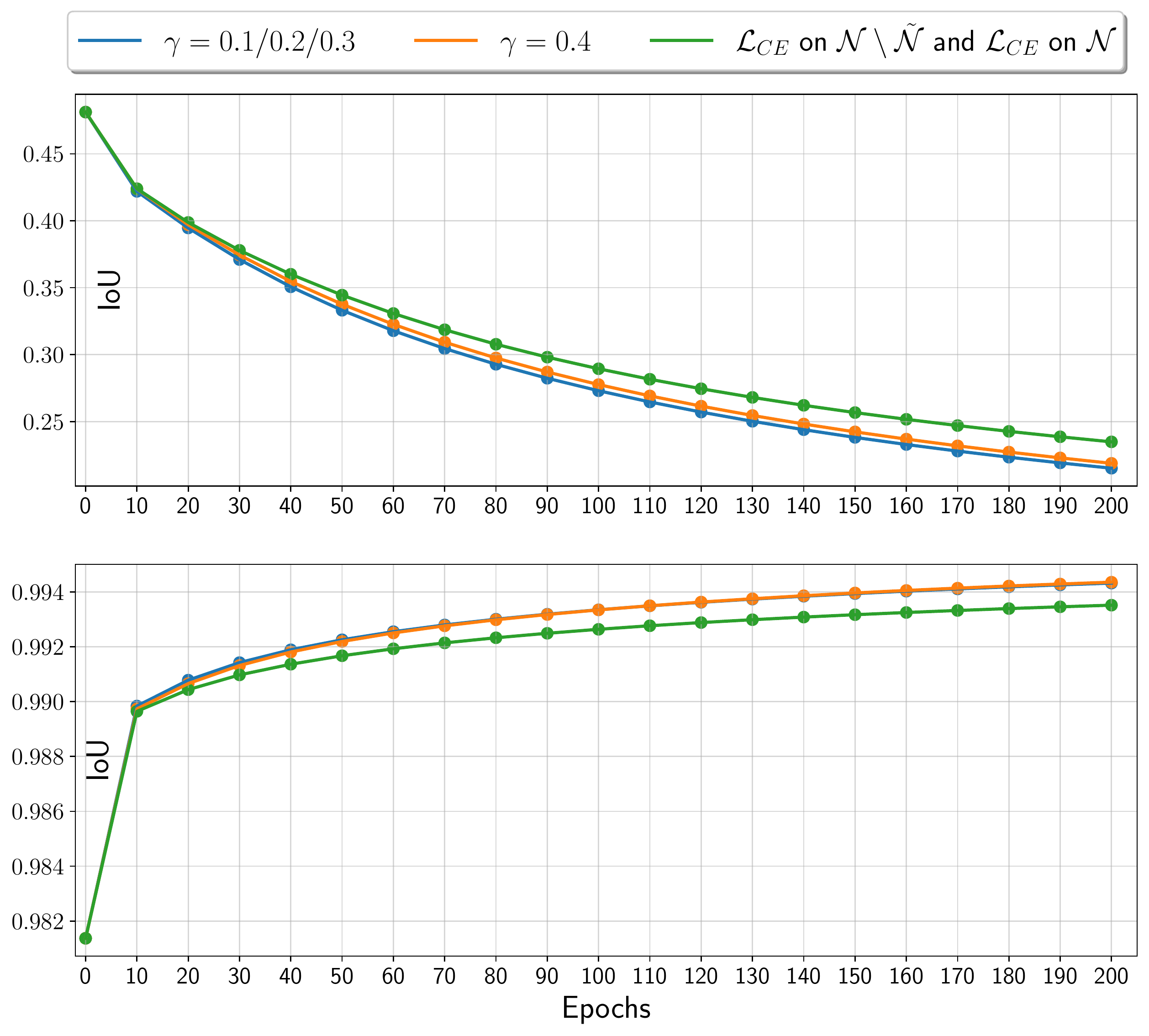}}
\centering
% \begin{subfigure}{0.495\textwidth}
% \includegraphics[width=\textwidth]{img/targeted_0.pdf}
% \vspace{-1em}
% \caption{}
% \end{subfigure}
% \begin{subfigure}{0.495\textwidth}
% \includegraphics[width=1.0\textwidth]{img/targeted_1.pdf}
% \vspace{-1em}
% \caption{}
% \end{subfigure}
\caption{\small{Summary of the optimization process for an image-specific \textbf{pedestrian}$\rightarrow$\textbf{NN} attack. The IoU values at each epoch are averaged over 100 different attacks. Top image shows the IoU of the class \textbf{pedestrian}, whereas bottom image shows the IoU of the class \textbf{road}. For each value of $\gamma$ tested, our loss formulation outperforms the standard cross-entropy loss.}}
\vspace{-1em}
\label{fig:loss_targeted}
\end{figure}

Figure \ref{fig:loss_targeted} shows the optimization process of the image-specific targeted attack \textbf{pedestrian}$\rightarrow$\textbf{NN} (the values are averaged on 100 different attacks). The plot at the top shows the IoU of the attacked class against the original labels, while the one at the bottom shows the IoU of the class \textbf{road} against the target labels.

As it is shown in the latter plot, the IoU of the class that is not under attack grows because the target label includes pixels of the class road (or other classes) that replaced the attacked class, and the prediction is forced to mimic the target label. Please note that, in the targeted case, the tested $\gamma$ values are the ones that perform best, i.e., those $< 0.5$. This is opposed to the untargeted case, since the attack is formulated to minimize the loss, and not to maximize it.

Also in the targeted case, our loss formulation outperforms the standard cross-entropy loss.
%}

% \begin{figure*}[!h]
% \centering
% \makebox[\columnwidth]{\includegraphics[scale=0.49]{img/carla_gamma_ablation.png}}
% \caption{\small{Comparison of adversarial patch optimizations on CARLA-scene 1 using different loss functions with the scene specific attack (two versions of the standard pixel-cross entropy and our formulation with multiple values of $\gamma$). The reference network here is ICNet \cite{icnet_paper}.}}
% \label{fig:app_specific_gamma_ablation}
% \end{figure*}

%%%%%%%%%%%%%%%%%%%%%%%%%%%%%%%%%%%%%%%%%%%%%%%%%%%

% \subsection{Testing patches in the real world}
% GIULIO - da spostare alla fine e mettere insieme detection
% Put and describe real world images here. 

%%%%%%%%%%%%%%%%%%%%%%%%%%%%%%%%%%%%%%%%%%%%%%%%%%%

\subsection{Detecting robust adversarial patches}
This section provides a set of experiments aimed at investigating the detection and timing performance of the FPDA, described in Section \ref{s:proposed_det}. 

\subsubsection{Tuning the thresholds}
The Cityscapes dataset is used to set the algorithm's parameters $\mu_\ell, \sigma_\ell, \textit{$\nu$-percentile}$. In particular, a set of features $\mathcal{C}_\ell$ extracted from $1000$ images of the original training set are used to compute $\mu_\ell$ and $\sigma_\ell$. 
Then, after the normalization, the corresponding features $\hat{\mathcal{C}}_\ell$ are used to compute the $\textit{$\nu$-percentile}$ (where $\nu$ is set to $0.999$ for both HN and FPDA) and so $\theta_\nu$.
A second dataset (composed of other $1000$ clean images from the training set, plus the corresponding adversarial images) is used to compute the decision threshold $\varrho$.

%{\color{red}
% To evaluate the detection capabilities of FPDA, another set of $1000$ images, extracted from the same training set, was used. The original clean images (i.e., non-patched) were used to generate 1000 adversarial patched images.
%}
%This part is discussed more in details in the following for those experiments that involved $\delta$.
% In particular, the original $2000$ images extracted by the training set of dataset
% We performed our evaluation on the Cityscapes dataset and then on a personal set of images (see Section X). In both cases, we trained the algorithm parameters using $2000$ images obtained by the training set of Cityscapes. 
% In particular, we split the above set of $2000$ images in two subsets, the first one was used for extracting $\mu_\ell$, $\sigma_\ell$ and $\zeta\text{-quantile}$, where $\zeta$ is set to $0.999$ for both out approach and [CITE], while the others $1000$ images was used for tuning the $\delta$ threshold. For simplicity, we denoted the latter subset as the calibration set. 

The detection algorithm requires specifying a given layer $\ell$. The choice of the layer has been the subject of many preliminary experiments. The real-time SS models considered in this paper fuse information extracted from different parallel branches. The selected layer $\ell$ is chosen as the first layer that joins all the model's branches.  
In fact, it is reasonable to detect anomalies in the first fusion point: it might be possible to create adversarial patches capable of bypassing the detection mechanism placed within some branches of the model by exploiting the vulnerabilities of others. Moreover, we empirically found it to be the point with the best detection performance. 
\\
% We selected as layers $\ell$ the first layer of each model that join multiple branches. These layers are the output of fusion-like layers available on both ICNet, BiSeNet and DDRNet.
% The rationale of this choice relies on the fact that it is possible to craft adversarial patches capable to skip detection mechanism posed inside certain branches of the model while exploiting vulnerabilities of others. Posing a statistical analysis on the join point of model help avoiding the previous issue. 

\subsubsection{On the validity of the method}

\begin{figure}
\centering
\makebox[\columnwidth]{\includegraphics[scale=0.265]{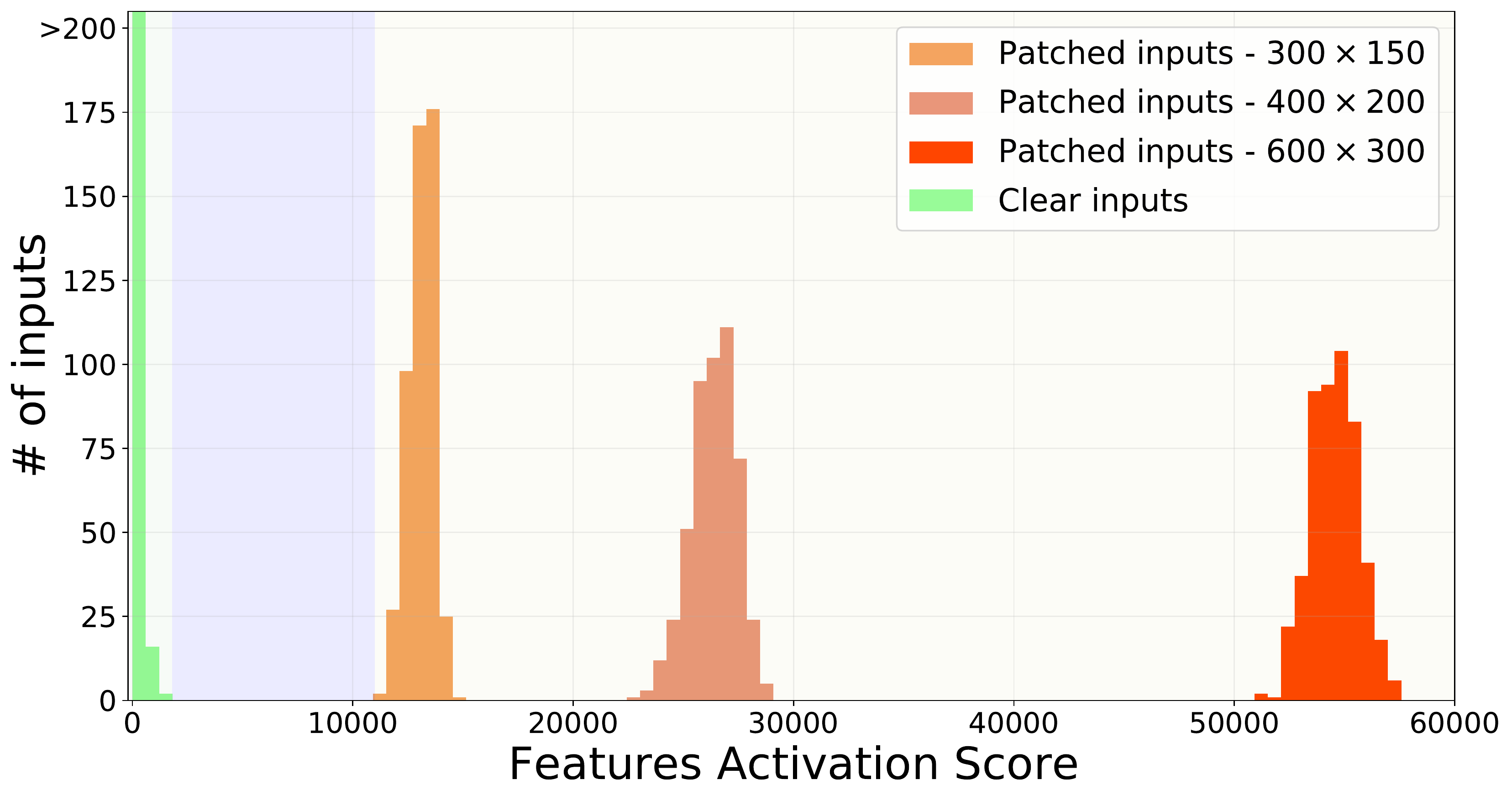}}
\caption{\small{Distribution of the clean and patched images as a function of the features over-activation \textit{score} computed by the Algorithm \ref{alg:agg-SRC} with BiSeNet. The grey area highlights the set of $\varrho$ thresholds that achieve a perfect detection accuracy with the tested images.}}
\vspace{-1em}
\label{fig:dist_detection}
\end{figure}

Figure \ref{fig:dist_detection} reports the distribution of clean and patched images extracted from the validation set of Cityscapes as a function of the \textit{score} computed by the proposed method with the BiSeNet model. 
The same kind of patches analyzed in Section \ref{s:unt_cityscapes} are used ($150\times300, 200\times400, 300\times600$).
%with width 300, 400, and 600 pixels. 

\begin{figure}
\centering
\makebox[\columnwidth]{\includegraphics[scale=0.24]{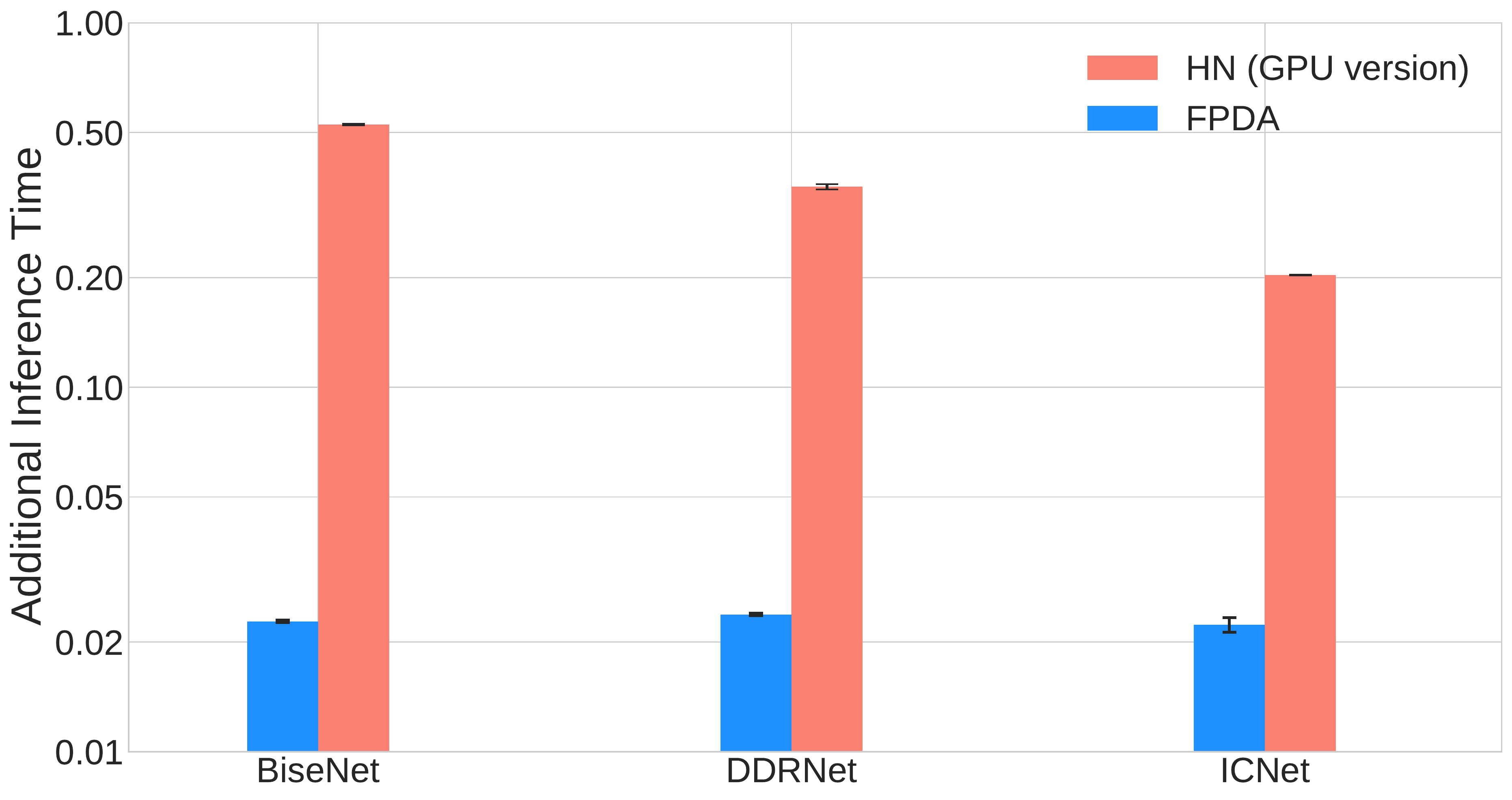}}
\caption{\small{Additional inference time computed as the ratio between the execution time of the detection algorithm and the original model inference time.}
%{\color{blue} TBD: Ho riportato il confronto solo tra HN GPU e FPDA GPU. Non ha molto senso secondo me testare la versione CPU. }
}
\vspace{-1.5em}
\label{fig:time_detection}
\end{figure}

As shown in the Figure \ref{fig:dist_detection}, the distribution of the clean images is much closer to zero than all the other adversarial distributions. 
In particular, the larger the adversarial patch (and consequently more effective), the higher the score computed by Algorithm \ref{alg:agg-SRC}.
The area of score values colored in grey highlights the set inside which any selected threshold $\varrho$ is able to detect all the adversarial images without introducing false negatives (i.e., no clean image is detected as unsafe).
\\

\subsubsection{Timing analysis}
As already mentioned in Section \ref{s:proposed_det}, the proposed method allows processing large-sized layer activations without affecting the timing performance of the model. 
Figure \ref{fig:time_detection} plots the additional inference time averaged over $8000$ iterations: the y-axis reports the relative execution time fraction introduced by Algorithm \ref{alg:agg-SRC} with respect to the nominal inference time required by the model.
%, i.e., $t_\textit{Alg} / t_\textit{model}$. 
As it can be noted from the figure, the proposed method adds a small latency, whereas HN presents larger additional execution times: from $20\%$ with the ICNet model to $50\%$ with BiSeNet.
% provided both in the original CPU formulation and on an extended GPU version, presents larger execution times by $20-50\%$ in the GPU case, and $200-800\%$ {\color{red}(CONTROLLARE)} with the original CPU formulation.

%(to provide a fari comparison with respect our GPU-based solution). 
While it is fair to remark that HN is not conceived for semantic segmentation models (where the number of features to take into account is usually larger than common image classification models), the proposed approach solves this issue by processing a large number of features in negligible additional time. 
\\
\subsubsection{Defense-aware attack}
\label{s:def_aware}
Although the adversarial patches crafted as explained in the previous sections are easily detectable, it is important to assess the goodness of the FPDA against ad-hoc attacks that exploit the knowledge of the applied defense.
To this purpose, an extension of the attack formulation proposed in this paper was concevied to craft untargeted adversarial patches with the intention of also deceiving the FPDA.

Since the FPDA is based on the detection of over-activated features, we crafted patches that are adversarial (i.e., capable of misclassifying a large number of pixels predicted in SS outcomes), while, at the same time, are able to keep the activated features within the range of the original distribution (i.e., minimizing the over-activations).
This is accomplished by solving the iterative optimization method discussed in Section \ref{s:proposed} but using a loss function $\mathcal{L}(f(\tilde{x}), y)$ that includes both the adversarial loss function $\mathcal{L}_{adv}(f(\tilde{x}), y)$ (in short $\mathcal{L}_{adv}^{\tilde{x}}$) and an additional \emph{activation loss} $\mathcal{L}_{\hat{\mathcal{C}_\ell}}^{\tilde{x}}$. The former is the loss function introduced in Section \ref{s:proposed}, while the latter is a new loss function that returns a cost proportional to the squared over-activation of the normalized features $\hat{\mathcal{C}_\ell}$ (computed by performing the first operations of Algorithm \ref{alg:agg-SRC} on $\tilde{x}$), i.e, $\mathcal{L}_{\hat{\mathcal{C}_\ell}}^{\tilde{x}} = ||\hat{\mathcal{C}_\ell}||_2^2$.

Therefore, the gradient of $\mathcal{L}(f(\tilde{x}), y)$ used during the optimization method was redefined as:

\begin{equation}
\nabla_{\delta_k} \mathcal{L}(f(\tilde{x}), y) = \beta \cdot \frac{\nabla_{\delta_k} \mathcal{L}_{adv}^{\tilde{x}}} {||\nabla_{\delta_k} \mathcal{L}_{adv}^{\tilde{x}}
||_{2}} - (1-\beta) \cdot \frac{ \nabla_{\delta_k} \mathcal{L}_{\hat{\mathcal{C}_\ell}}^{\tilde{x}}}{||\nabla_{\delta_k} \mathcal{L}_{\hat{\mathcal{C}_\ell}}^{\tilde{x}})||_{2}}  ~,
\end{equation}

\noindent where $\beta$ is a  parameter  introduced  to  balance  the  importance  of $\mathcal{L}_{adv}$ and $\mathcal{L}_{\hat{\mathcal{C}_\ell}}$, and $k=\{1, ..., N_p\}$.
In particular, when $\beta=1.0$, the resulting patch will be optimized to minimize the activated features. Conversely, moving $\beta$ to lower values reduces the previous effect and increases the adversarial strength.

The following experiments investigate the relation between detectability and adversarial effect. To this end, several adversarial patches (300x600 and 200x400 pixels) are crafted using several values of $\beta \in \{0.0, 0.1, 0.2, 0.3, 0.4, 0.5, 0.6, 0.7, 0.8, 0.9, 1.0\}$.
%In particular, given a certain $\beta$, the detectability is evaluated by performing a ROC analysis from the $1000$ clean and 1000 adversarial images introduced earlier \TODO{``earlier'' is ambiguous: refer the specific section}.
In particular, given an adversarial patch crafted with a certain $\beta$, the detectability is evaluated by performing an ROC analysis with $1000$ images of the Cityscapes train set and the corresponding attacked ones. Thus, such a subset is used to study the True Positive Rate and False Positive Rate as a function of the decision threshold $\varrho$.
%As anticipated in \ref{ss:setup}, other 1000 inputs of the train set are used to compute $\mu_\ell$, $\sigma_\ell$ and $\theta_\nu$, which are parameters of FPDA.}
On the other side, the adversarial effect is evaluated using the mIoU computed on the Cityscapes validation set. Lower values indicate a more effective adversarial attack.
%while the adversarial effect is obtained by computing the mIoU on the same set of clear images.
%In particular, we labeled the clear images as negative (safe) samples and the same images perturbed with the corresponding adversarial patch as positive (unsafe) samples. 

\begin{figure*}
\centering
\begin{subfigure}{0.57\textwidth}
\includegraphics[width=\textwidth]{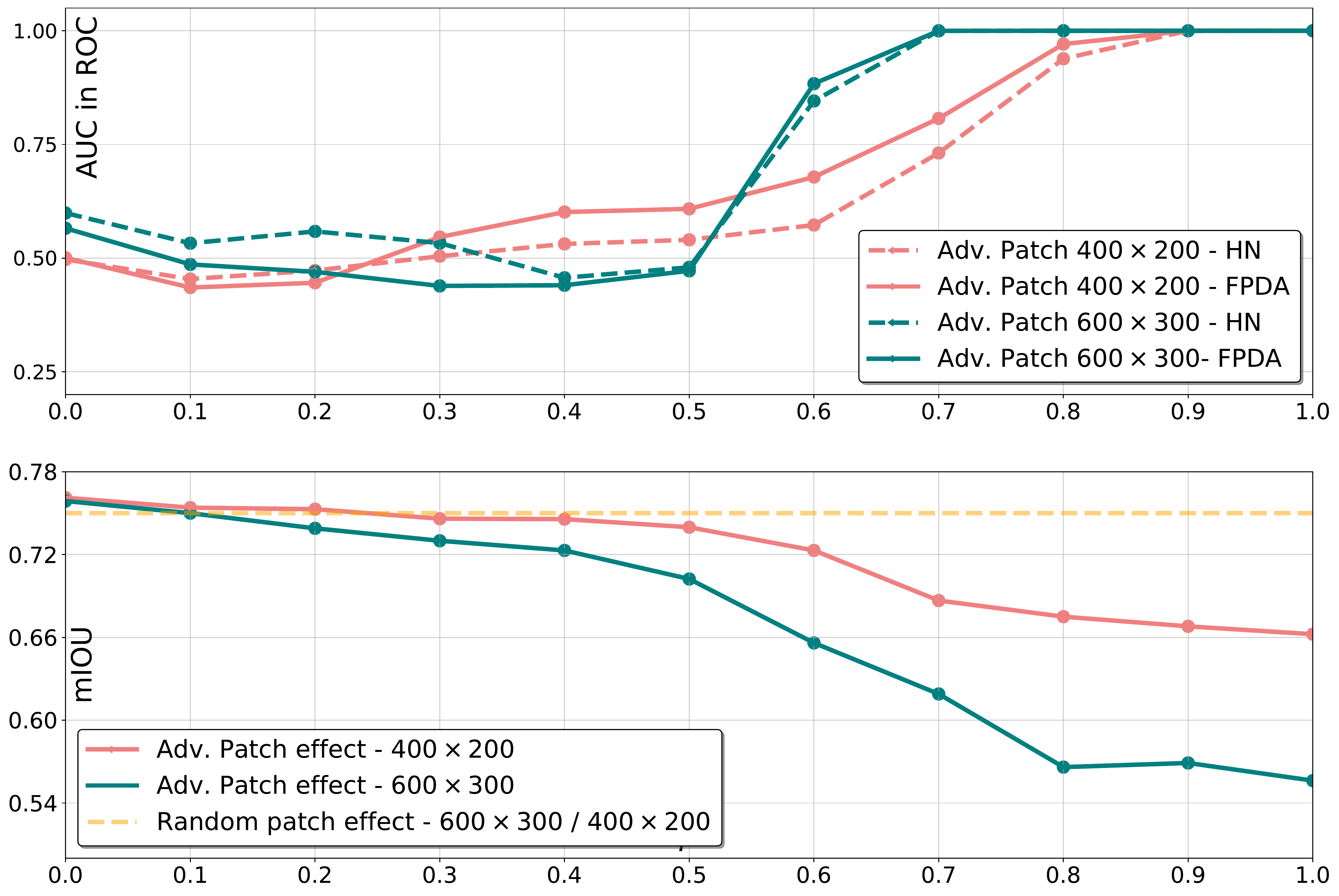}
\caption{}
\label{fig:addaptive_AUC_icnet}
\end{subfigure}
\begin{subfigure}{0.405\textwidth}
\includegraphics[width=1.0\textwidth]{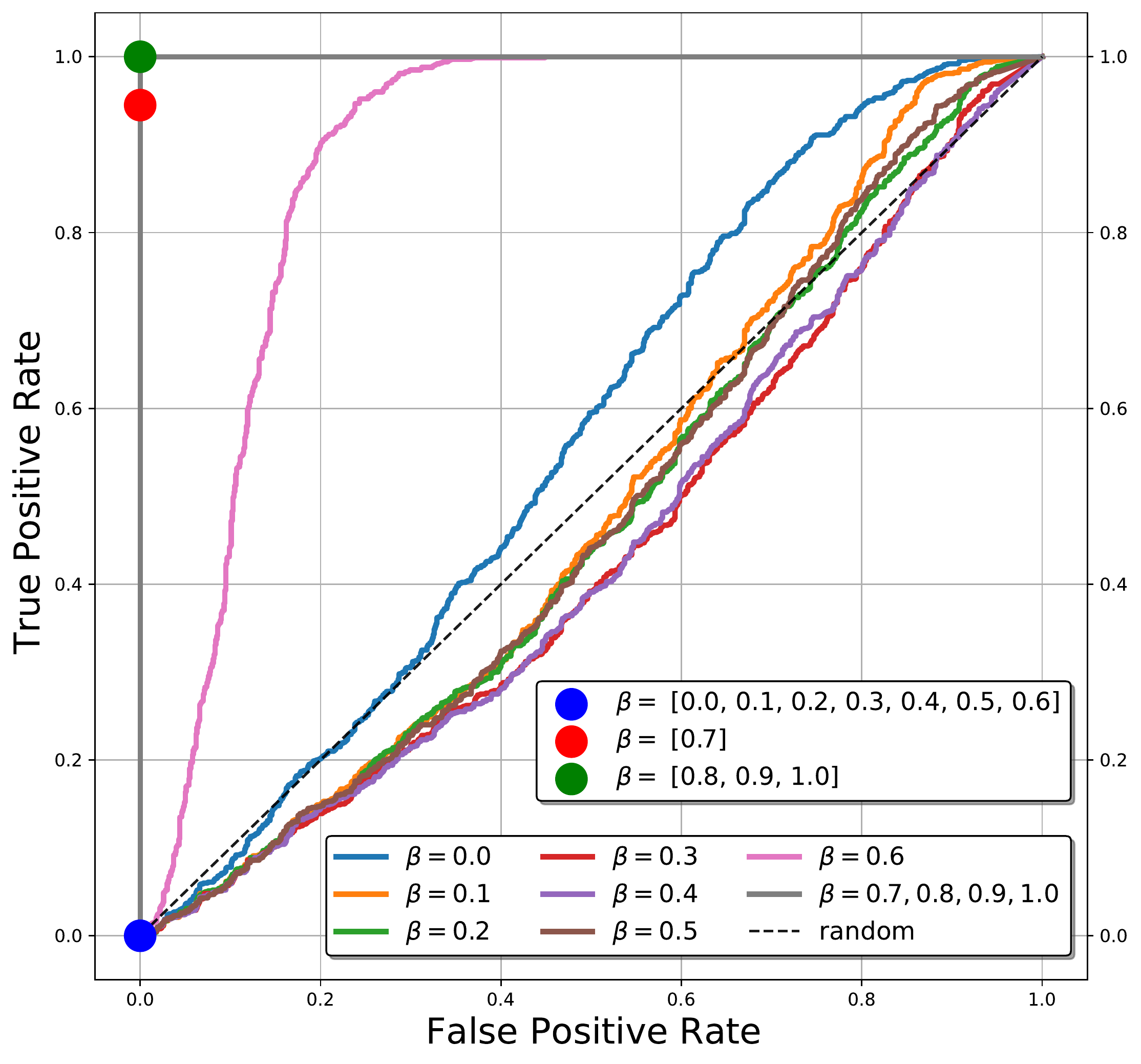}
\caption{}
\label{fig:addaptive_ROC}
\end{subfigure}
\caption{\small{Figure (a) reports a comparison between the detectability and adversarial effect of mulitple patches crafted changing the $\beta$ parameter; Figure (b) plots the ROC curves corresponding to all the previous tested adversarial patches and highlights their TFPs/FPRs corresponding to an unique threshold selected with $\beta=0.8$. These results were obtained with the ICNet model.}}
\end{figure*}

Figure \ref{fig:addaptive_AUC_icnet} shows a comparison between the detection accuracy, specified through the AUC values (extracted from each ROC curve) and the adversarial effect obtained with each tested version of $\beta$.
% The detection accuracy is specified through the AUC values that is extracted by each ROCs curve. The adversarial effect instead is given by the mIoU computed on the same set of clear images, the lower means a more effective adversarial effect.
Clearly, for low values of $\beta$ (when the optimization is mainly focused on keeping the activated features small), the detectability of all the tested methods is very poor. 
However, these patches have a low adversarial effect since the obtained mIoUs are close to the ones computed with random patches. 
%random patch case 
%\TODO{refer to the mIoU to evaluate effectiveness}.
Increasing the value of $\beta$, the adversarial effect of the patches increases but, at the same time, also their detectability grows. 
These results remark that it exists an intrinsic relation between the adversarial effect and the over-activation of features, such that highly effective patches are more prone to be detected by the proposed strategy, as also observed by previous works \cite{co_real}. %\TODO{(CITARE)}{\color{blue}[SOLVED]}

% \begin{figure}
% \centering
% \includegraphics[width=0.49\textwidth]{img/placeholder/ROC_adaptive_icnet.pdf}
% \caption{\small{PERCHE' IL RISO POI SPEGNE TUTTO} ddrnet mettere legen piu' al centro. BLABLA BLABLA BLABLA BLABLA BLABLA BLABLA BLABLA BLABLA BLABLA BLABLA A}
% \label{fig:roc_icnet}
% \end{figure}

Figure \ref{fig:addaptive_ROC} illustrates another important result, showing all the ROC curves computed for the $300 \times 600$ patches (their AUC score is the one shown in Figure \ref{fig:addaptive_AUC_icnet}).
%\TODO{How can one read the figure without a legend for each curve?} {\color{blue} [SOLVED]}
By looking at these curves, it is to possible to figure out the True Positive Rate (TPR) and False Positive Rate (FPR) with respect to a fixed threshold $\varrho$. 
%is the ROC analysis performed with clear images and the same perturbed with the adversarial patch crafted using a certain verison of $\beta$.
The threshold $\varrho$ is set as the optimal value (cut-off point) of the ROC analysis performed with $\beta=0.8$, where the AUC achieves $1.0$, meaning that safe and unsafe samples can be perfectly classified for that $\beta$ value.  

As highlighted by the plot's legend, only the most dangerous patches are detected correctly (those with $\beta \geq 0.8$). 
Also, and most importantly, all the ROC points have an FPR equal to 0.0, meaning that clean images are always classified as safe. 

The above experiments were also performed on the BiSeNet and DDRNet models, showing similar results (see the supplementary material). 

\subsection{Adversarial patches in the real world}\label{ss:rw_patch}
This section evaluates the effectiveness of the proposed attack pipeline and the detection algorithm in the real world.

\subsubsection{Effect of a real-world patch} 
\begin{figure*}
     \centering
     \begin{subfigure}{1.0\textwidth}
         \begin{subfigure}{0.195\textwidth}
             \centering
             \includegraphics[width=\textwidth]{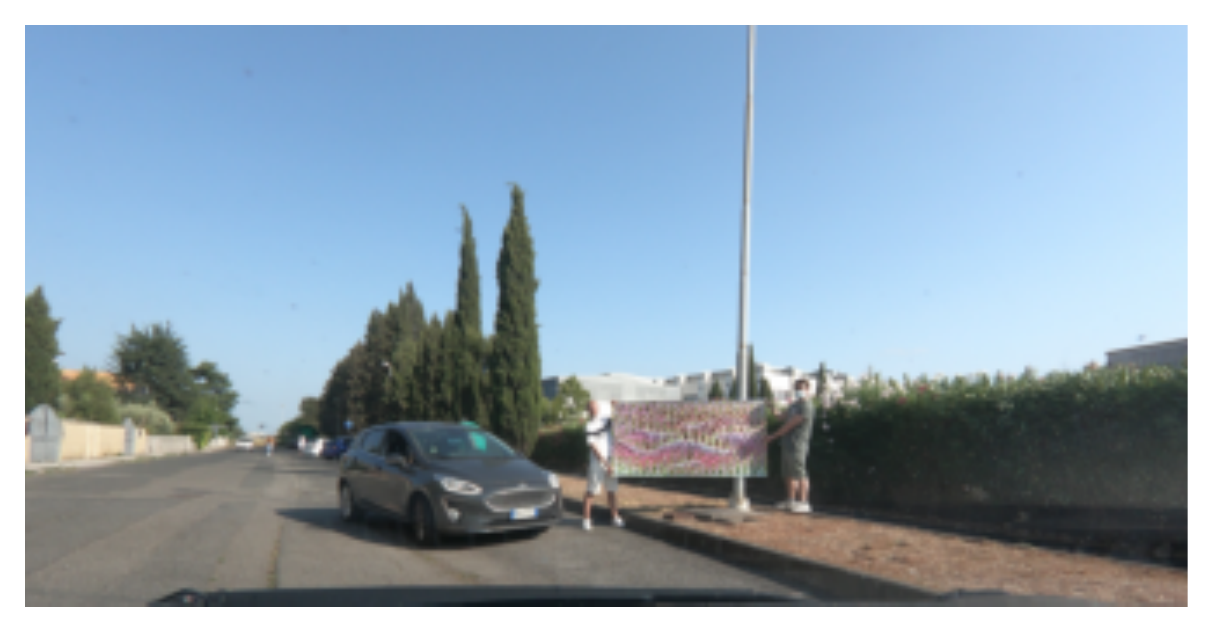}

         \end{subfigure}
         \begin{subfigure}{0.195\textwidth}
             \centering
             \includegraphics[width=\textwidth]{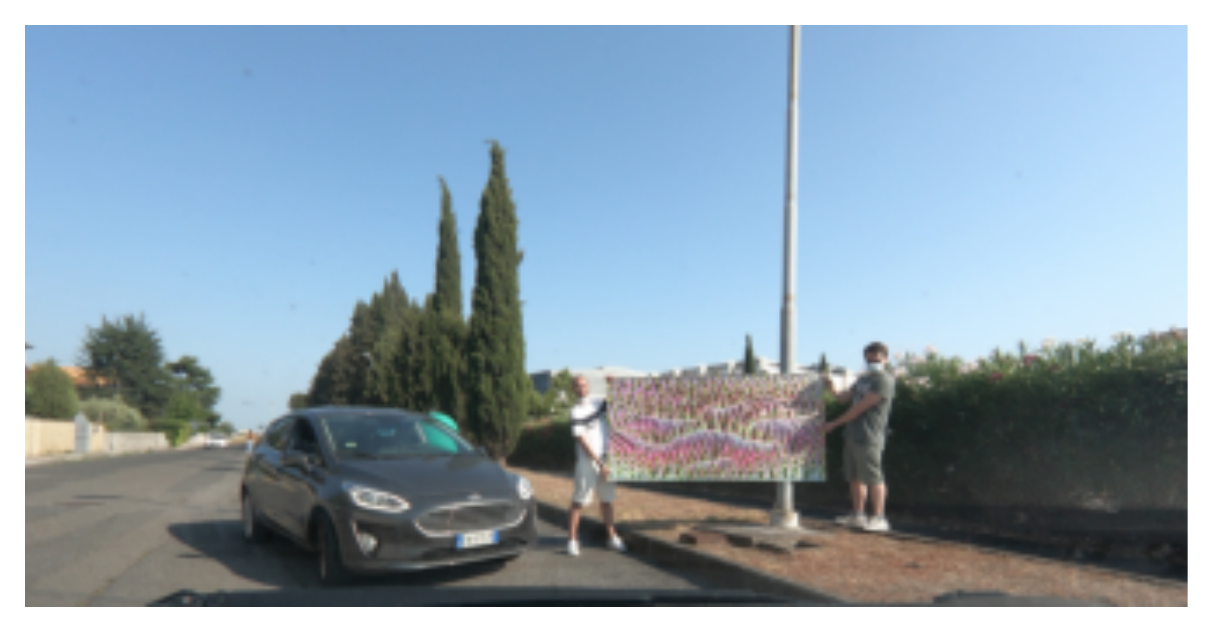}

         \end{subfigure}
         \begin{subfigure}{0.195\textwidth}
             \centering
             \includegraphics[width=\textwidth]{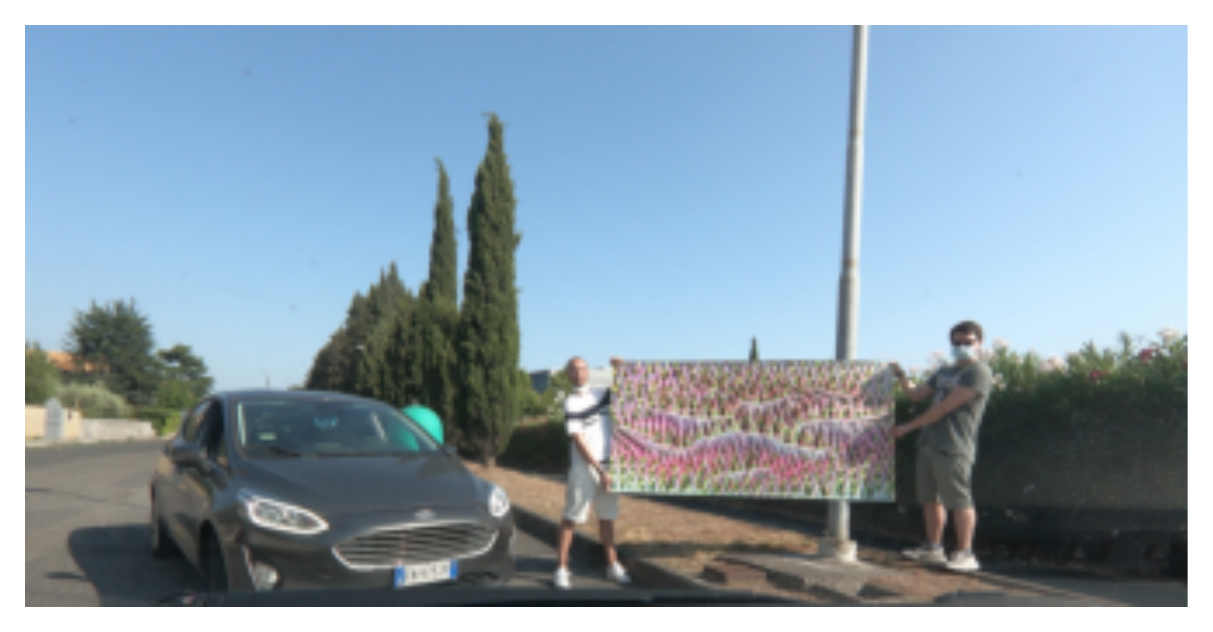}

         \end{subfigure}
         \begin{subfigure}{0.195\textwidth}
             \centering
             \includegraphics[width=\textwidth]{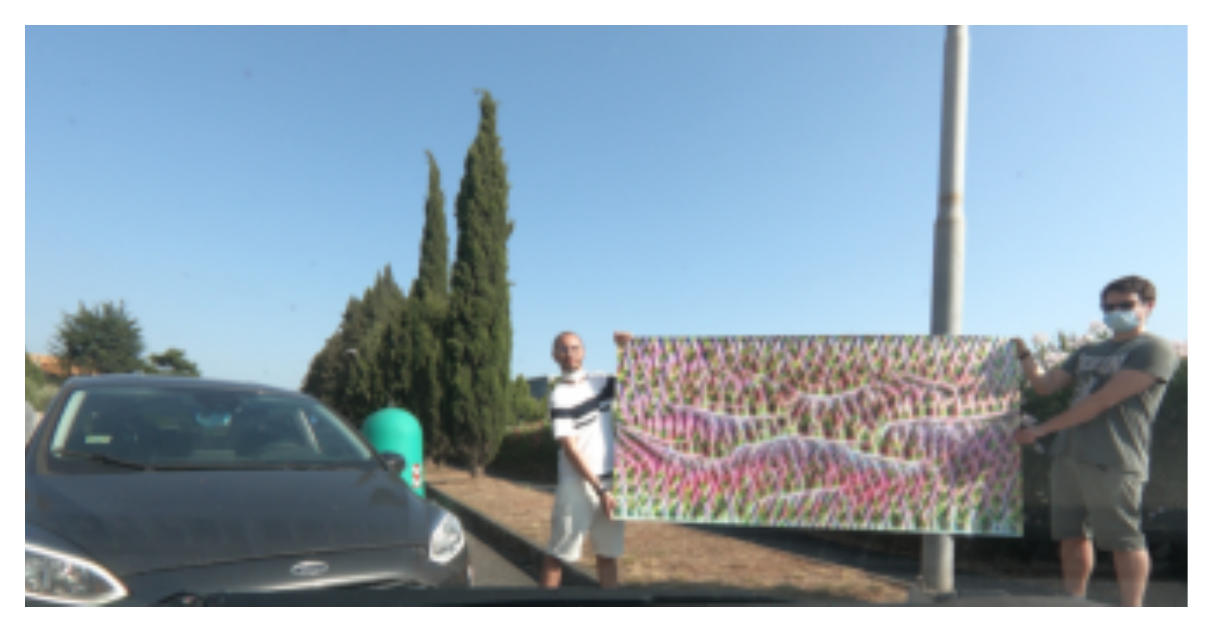}

         \end{subfigure}
         \begin{subfigure}{0.195\textwidth}
             \centering
             \includegraphics[width=\textwidth]{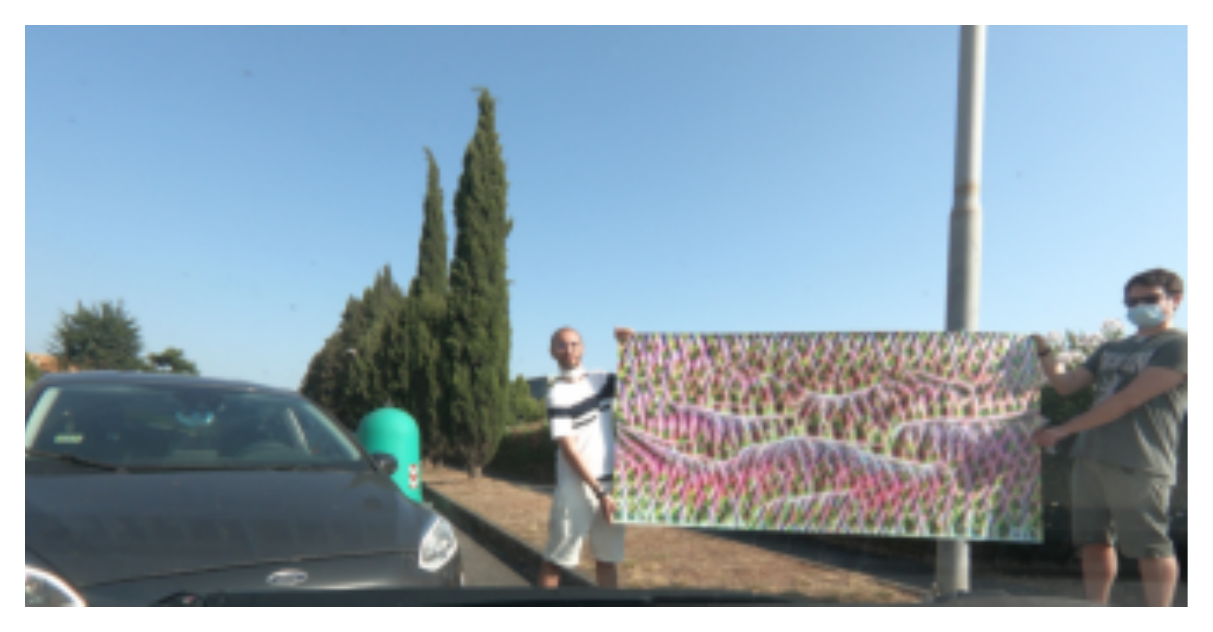}

         \end{subfigure}
    \end{subfigure}
    
    \begin{subfigure}{1.0\textwidth}
         \begin{subfigure}{0.195\textwidth}
             \centering
             \includegraphics[width=\textwidth]{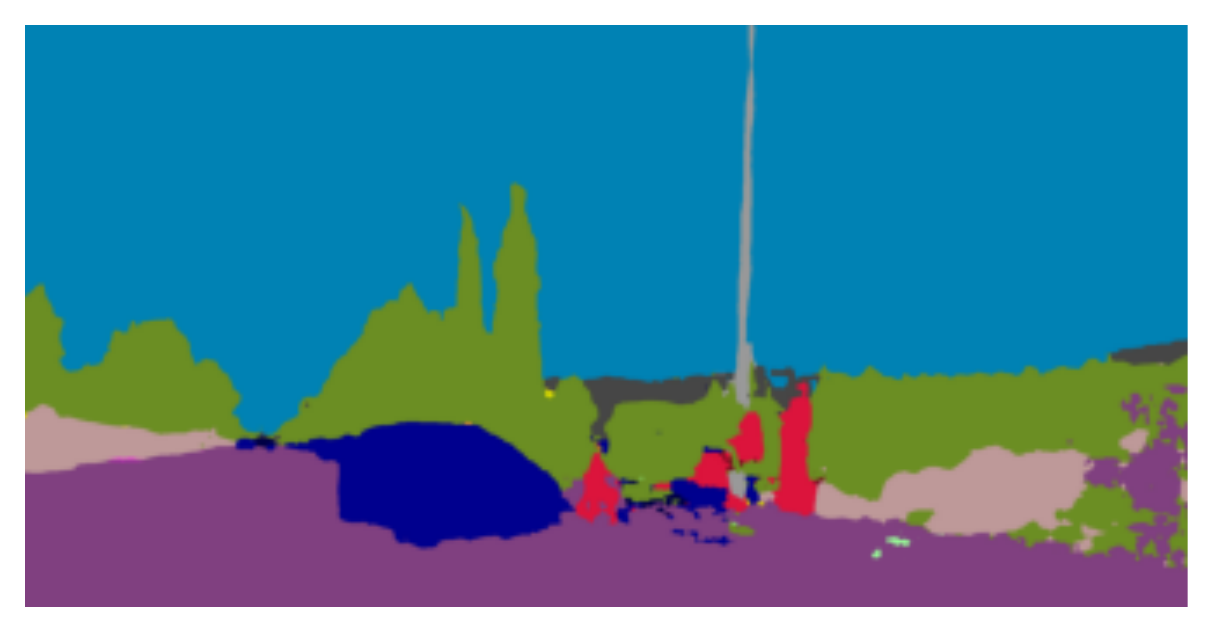}

         \end{subfigure}
         \begin{subfigure}{0.195\textwidth}
             \centering
             \includegraphics[width=\textwidth]{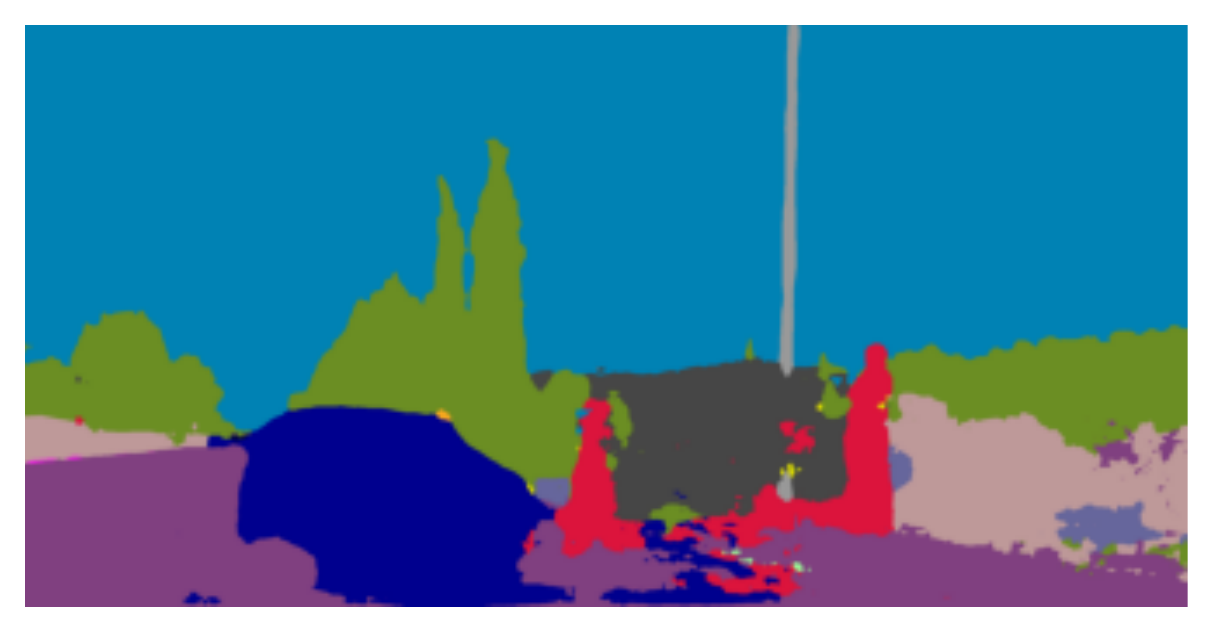}

         \end{subfigure}
         \begin{subfigure}{0.195\textwidth}
             \centering
             \includegraphics[width=\textwidth]{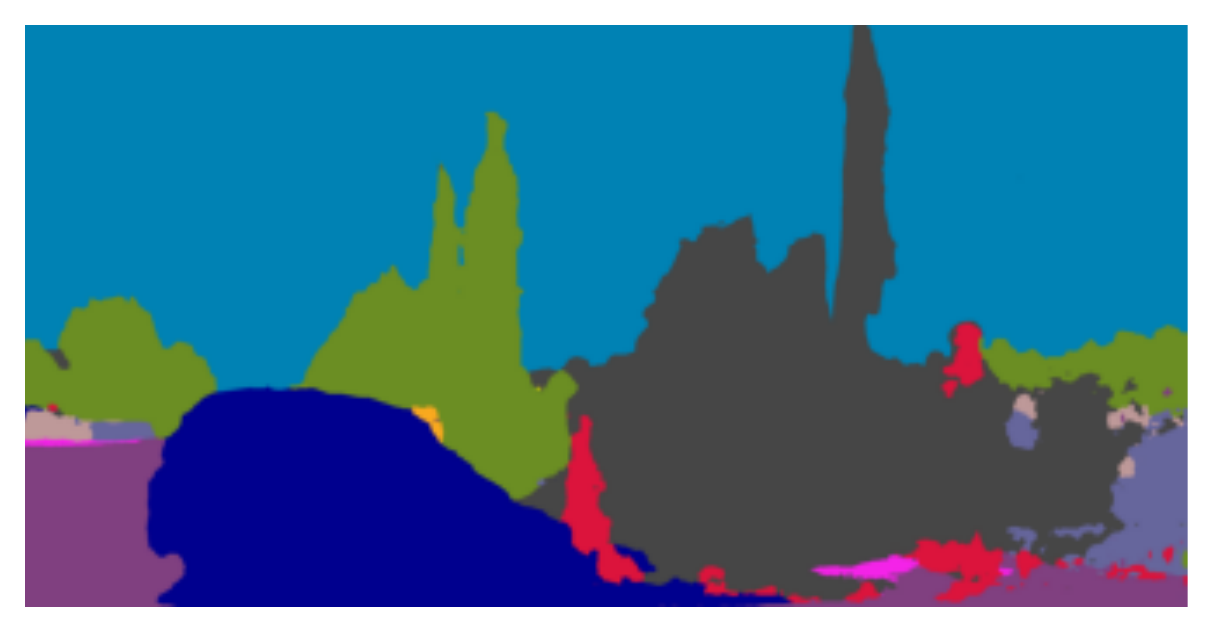}

         \end{subfigure}
         \begin{subfigure}{0.195\textwidth}
             \centering
             \includegraphics[width=\textwidth]{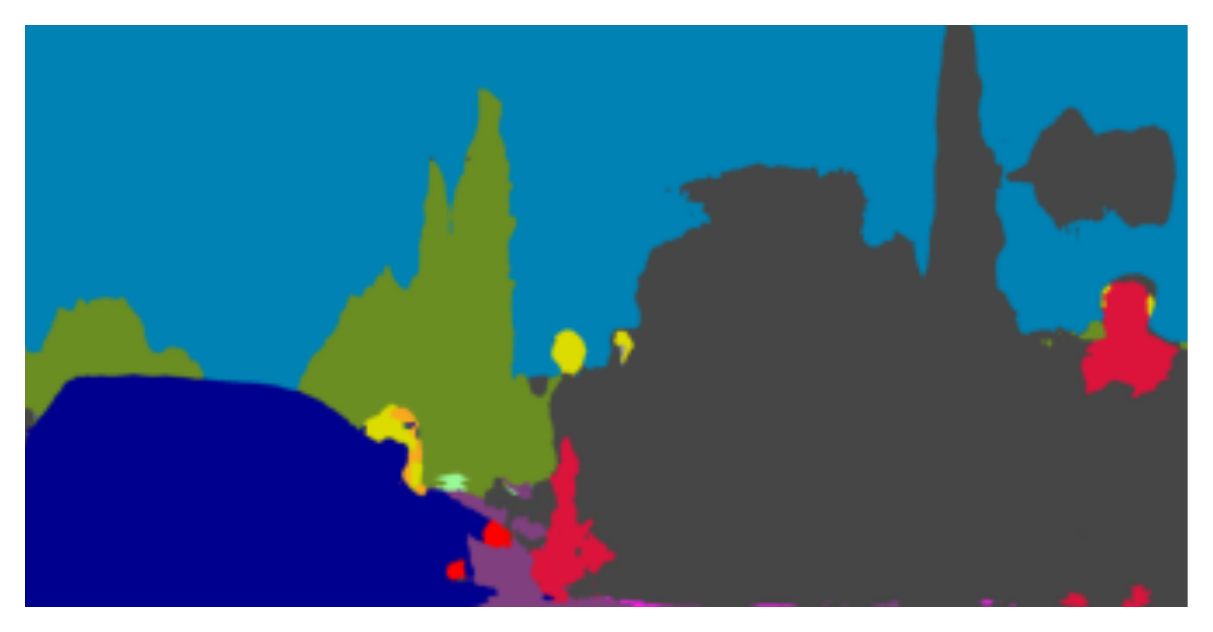}
         \end{subfigure}
         \begin{subfigure}{0.195\textwidth}
             \centering
             \includegraphics[width=\textwidth]{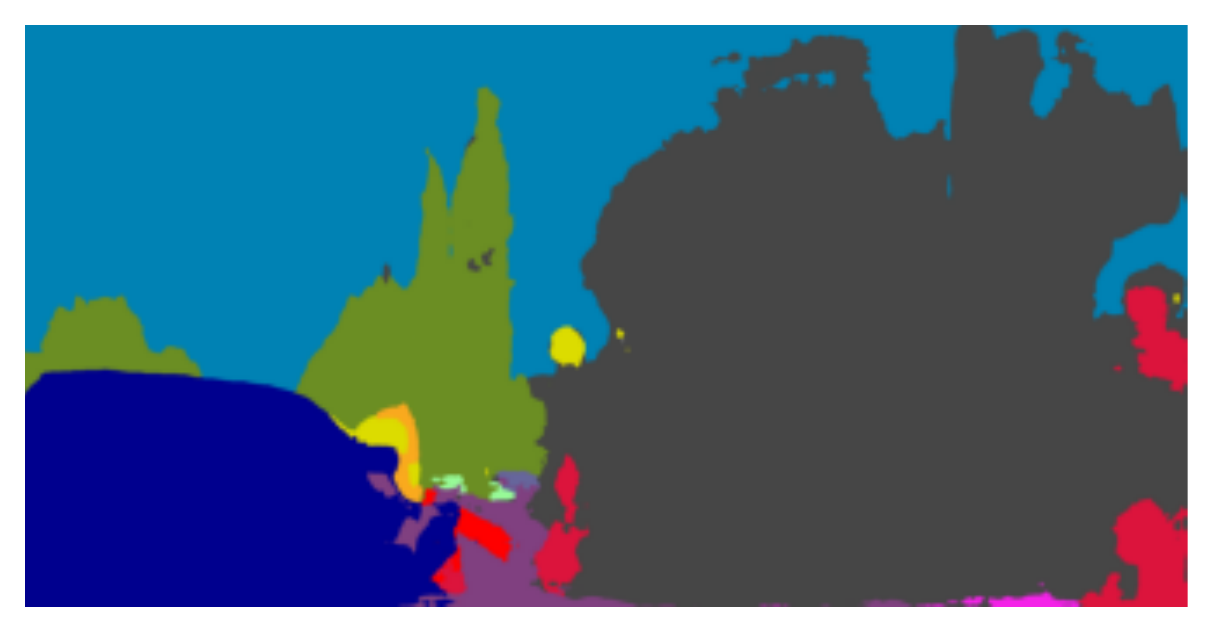}
         \end{subfigure}
    \end{subfigure}

    \begin{subfigure}{1.0\textwidth}
         \begin{subfigure}{0.195\textwidth}
             \centering
             \includegraphics[width=\textwidth]{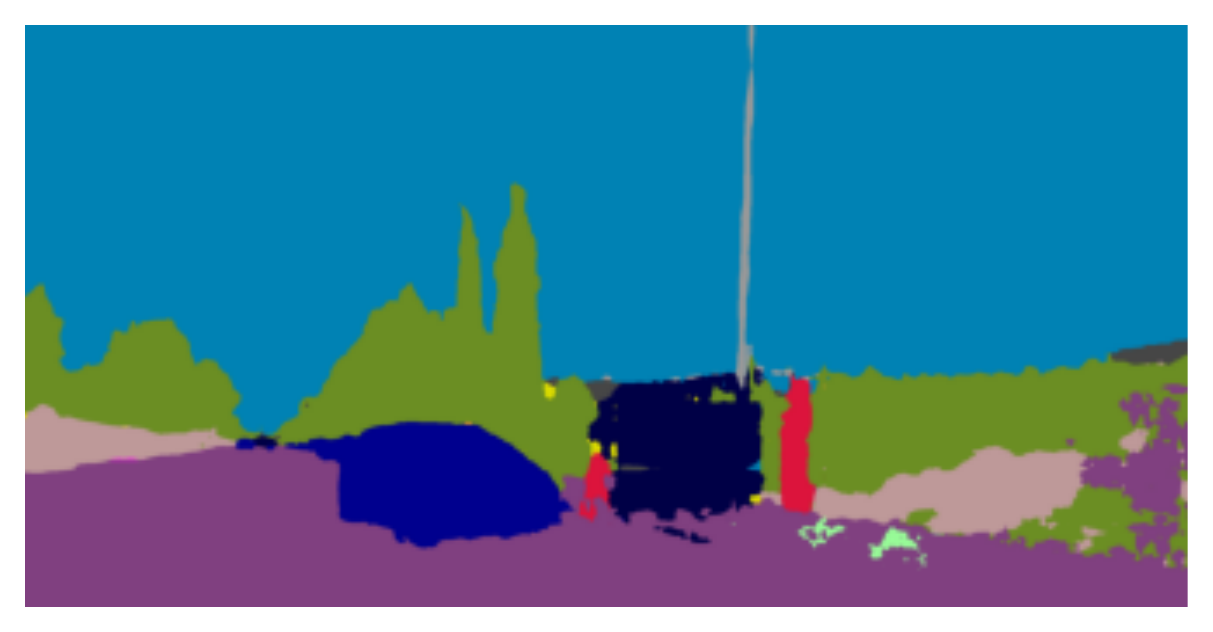}

         \end{subfigure}
         \begin{subfigure}{0.195\textwidth}
             \centering
             \includegraphics[width=\textwidth]{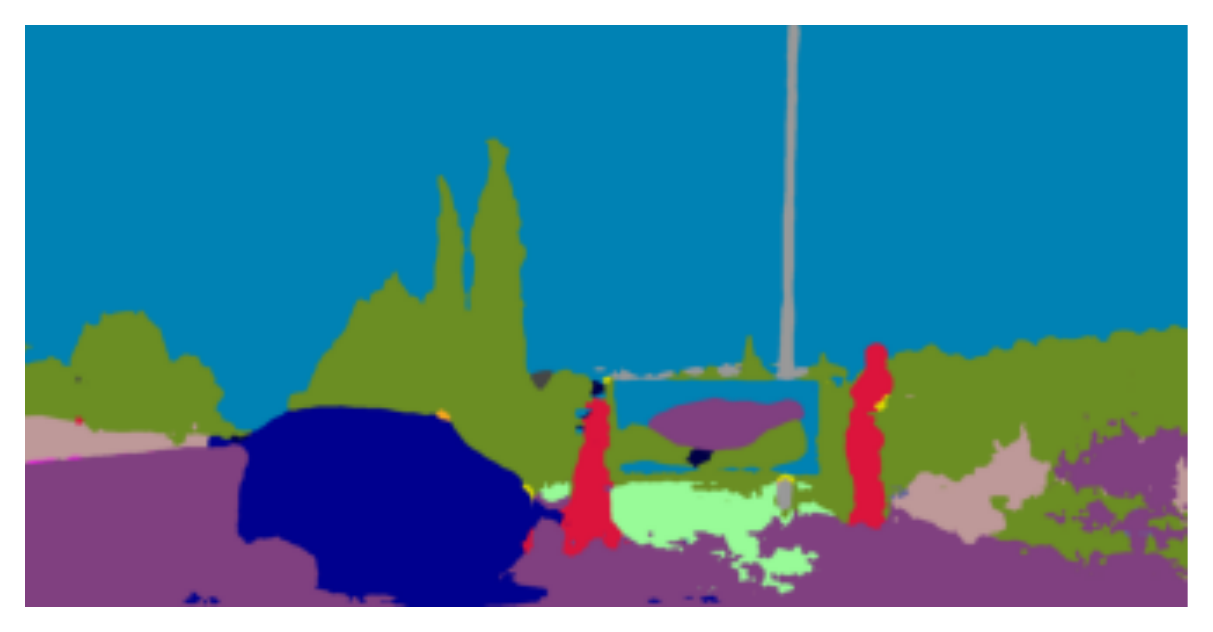}

         \end{subfigure}
         \begin{subfigure}{0.195\textwidth}
             \centering
             \includegraphics[width=\textwidth]{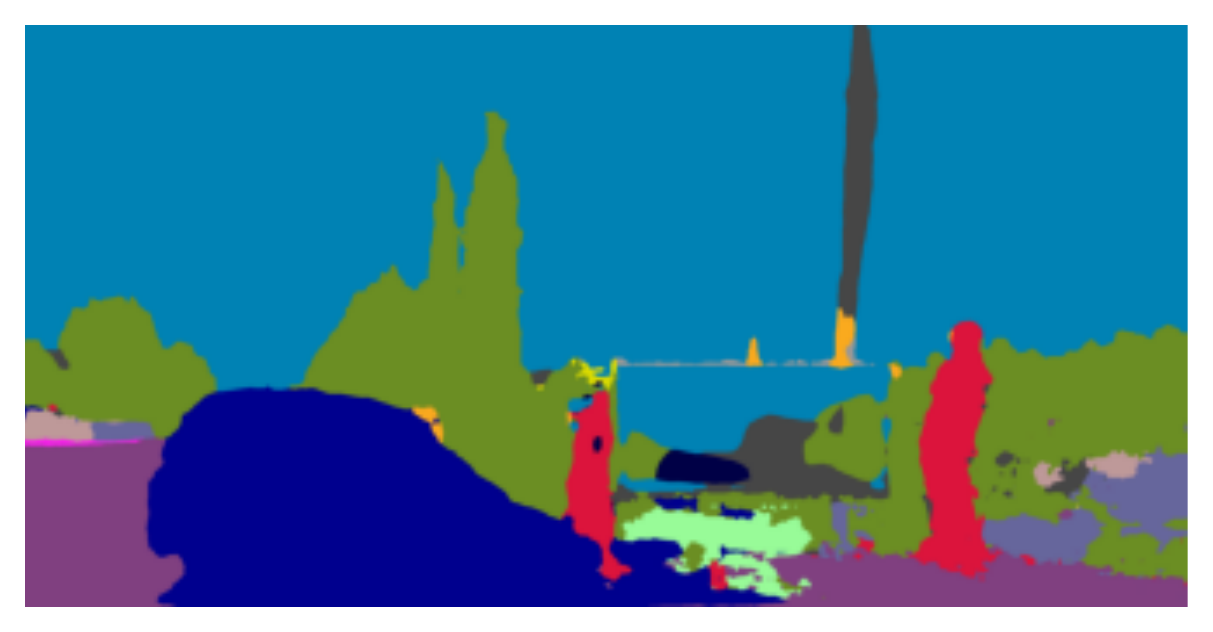}

         \end{subfigure}
         \begin{subfigure}{0.195\textwidth}
             \centering
             \includegraphics[width=\textwidth]{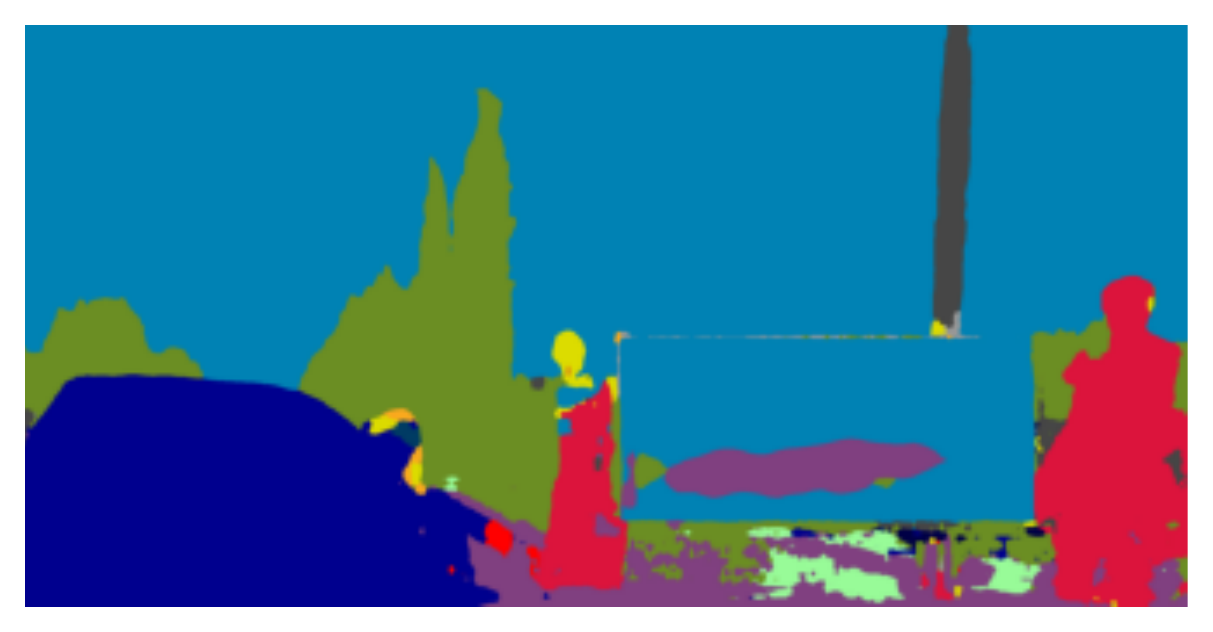}
         \end{subfigure}
         \begin{subfigure}{0.195\textwidth}
             \centering
             \includegraphics[width=\textwidth]{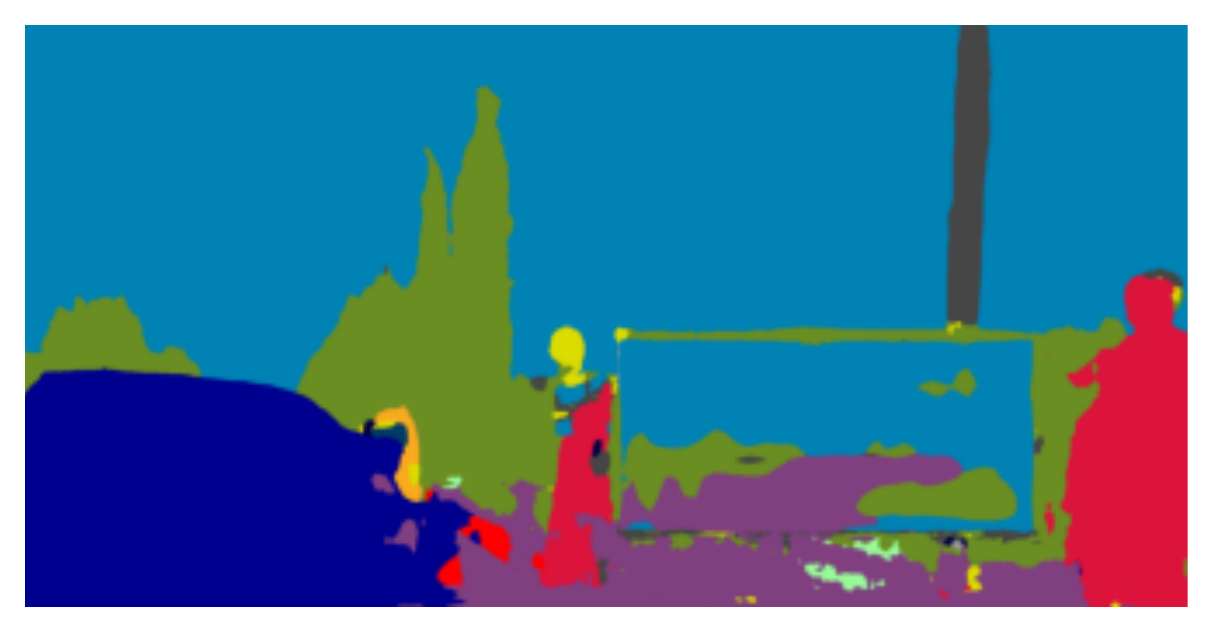}
         \end{subfigure}
         \caption{}
         \label{fig:realworld}
    \end{subfigure}
    
    \begin{subfigure}{1.0\textwidth}
         \centering
         \includegraphics[width=\textwidth]{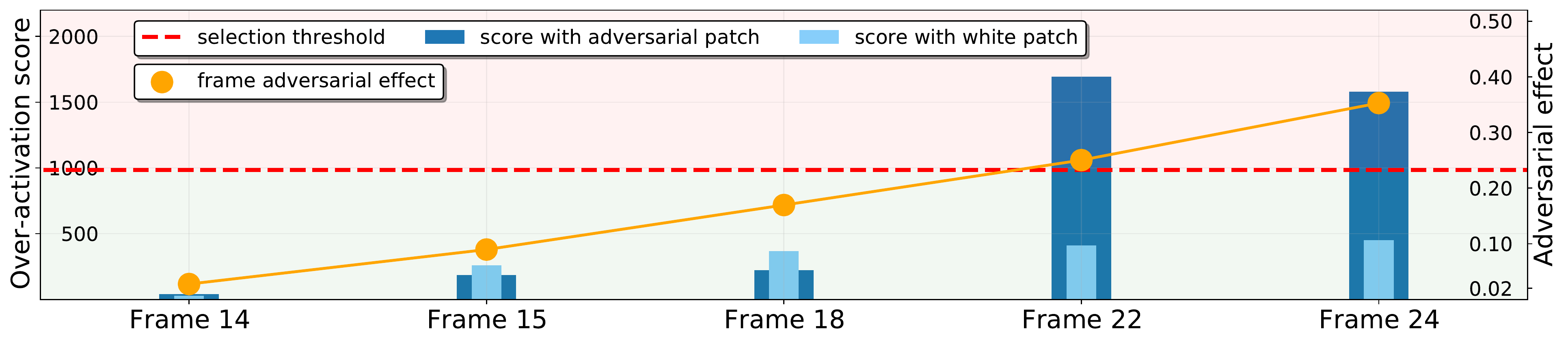}
         \caption{}
         \label{fig:def_realworld}
     \end{subfigure}
    \caption{\small{Real-world evaluation on ICNet of the attack method and the defense algorithm. In inset (a), the first row contains the original images with the printed adversarial patch, while the second row contains their obtained semantic segmentations. The third row instead shows the outcomes obtained by masking the patch areas of the previous inputs with white pixels. Inset (b) shows the over-activation score of the corresponding frames for the SS predictions in rows 2 and 3, and which of them are detected by using a proper selection threshold. Also, the adversarial effect is shown for each frame, which is extracted by comparing the original attacked images with the corresponding white-masked version.}}

\end{figure*}

To prove the effectiveness of the attack pipeline proposed in Section~\ref{s:proposed}, we used a custom dataset to craft an adversarial real-world patch using the EOT-based formulation. The dataset is composed of 1000 images that were collected by mounting an action camera on the dashboard of a real car, using a setup similar to the one of the Cityscapes dataset, and then driving the car through the streets of our city. %\TODO{which one? Pisa images? be specific} 
The patch was optimized for 200 epochs on the original pre-trained version of ICNet (since it showed good performance also on our personal real-world dataset).
Figure~\ref{fig:realworld} shows a sequence of frames recorded while moving in the direction of the adversarial patch, printed as a $1m \times 2m$ poster.

Although the presence of the optimized patch alters a significant area of the predicted SS, while the random patch does not (see the supplementary material), portions of the image far from its position are not affected. Furthermore, the attack performance increases as we move close to the patch. Additional analyses are provided in the supplementary material, while the complete video is available on the published repository.

It is worth remarking that testing adversarial patches for autonomous driving  in the real world poses a series of difficulties that heavily limited the tests. First, it is not easy to find a urban corner that shows good performance and is not crowded with moving vehicles (which  might  be  dangerous). Second, the patch must be printed in the highest resolution possible on a large rigid surface, which  might get  expensive.
\\

\subsubsection{Detectability of a real-world patch}
\label{s:def_rw}
Figure \ref{fig:def_realworld} reports the \textit{over-activation score} corresponding to each frame shown in Figure~\ref{fig:realworld}.
The red dashed line represents the threshold $\varrho$, used to distinguish between safe and unsafe predictions. 
In order to provide a clear visualization of the spatial effect caused by the printed adversarial patch (second row in Figure~\ref{fig:realworld}), we also report the SS obtained from the same images where the adversarial patch was masked with white pixels (third row in Figure~\ref{fig:realworld}) and also their corresponding scores (light blue) in Figure~\ref{fig:def_realworld}.

Since the mIoU does not properly encode the adversarial effectiveness on individual images, we show a different metric that we call \emph{adversarial effect}, computed as $ 1 - \frac{1}{|\mathcal{N} \setminus \mathcal{\tilde{N}}|} \sum_{i \in \mathcal{N} \setminus \mathcal{\tilde{N}}} (y_i == SS_i(x))$. The adversarial effect measures the average pixel dissimilarity between the predicted SS and a ground truth $y$. 
%such an effect use the overall accuracy to quantify the effectiveness of that adversarial patch, which is computed for each image as $( 1 - \frac{1}{|\mathcal{N}|} \sum_{i \in \mathcal{N}} (y_i == SS_i(x)) )$. 
Given the absence of proper ground truth labels $y$, we use the SS predicted from the white-masked images.
%,where the adversarial patch was masked with white pixels (third row in Figure~\ref{fig:realworld}) . 
As done in Section \ref{s:unt_cityscapes}, when evaluating the effect of patches on Cityscapes, we did not consider those pixels corresponding to the masked areas (which approximate the patch placement in the real world).
%\TODO{Can we add a measure of the attack effectiveness (mIoU?) below each bar?} {\color{blue} GUARDA MODIFICHE. DOMANDA: pensi sia necessario anche aggiungere immagini input masked? oppure possiamo lasciare così?}

In this real-world experiment, the tuning of $\varrho$ was performed on the same $1000$ images of the Cityscapes dataset with an adversarial patch having $\beta = 0.8$ and $300 \times 600$ (the same patch used to compute the threshold in Figure \ref{fig:addaptive_ROC}). 
However, it is worth noting that transferring patches in a real-world environment reduces the activation level of the perturbed features (as well as their adversarial effect). 
%To this purpose, before evaluating the detectability of the method among realistic images, an additional assumption was introduced for properly tuning the threshold $\varrho$ among the ROC curve. 
To this purpose, the threshold $\varrho$ should be decreased. This can be obtained by considering an FPR equal to 0.01 as a reasonable compromise (i.e., 1\% of clean images in the calibration set are wrongly classified as unsafe). 
This helps shift the threshold to a lower value, more likely to be closer to the over-activation score implied by patches transferred in the real world.

As it can be noted from Figure~\ref{fig:def_realworld}, frames 22 and 24 are above the threshold $\varrho$ (red line) meaning that the detection mechanism works also in a real world scenario. 
Moreover, their masked counterparts achieve a lower score, clearly below threshold $\varrho$, which stems to reason that high over-activations are mainly caused by the presence of the printed adversarial patch.
% Conversely, earlier frames are classified as safe, meaning that the number of over-activated features is close to the one of a 
%\TODO{``clean distribution'' is vague, be specific}. 
Conversely, earlier frames are classified as safe, meaning that the internal features do not have a considerable number of over-activated values. In fact, in these latter frames the patch has a lower adversarial effect, which is under 0.15, meaning that less than $15\%$ of out of patch pixels are classified differently from the masked versions. Therefore, the corresponding safe classifications could be acceptable 
However, also some notable adversarial effects could overcome the defense mechanism, as, for instance, for frame 18, which has several areas affected by the patch effect.
Improving the performance of the detection mechanism, while also providing deeper analysis on its transferability on real-world critical situations, is left as future work.

%METTERE NEI THREATS$
% Furthermore, since  weather  conditions  are  not  controllable  and  change through out the day, results  can  diverge  from  what  is  expected We  were  not  able  to  perform  the  scene-specific  attack since  it  requires  additional  geometric  information.    Thiswork  will  be  extended  in  the  near  future  to  perform  thescene-specific attack in the real world

% \subsection{Other experiments}
% TODO: fare un recap di tutti gli altri esperimenti inseriti negli additional materials.
\section{Conclusions} \label{s:conclusions}

This paper presented an extensive study of the real-world adversarial robustness of real-time semantic segmentation models for autonomous driving. 
%This was accomplished by evaluating the effect introduced by adversarial patches robust for a realistic environment. 
Two attack objectives (targeted and untargeted) were presented and tested using single- and double-patch configurations.
The investigations was carried out by testing benchmarks with increasing realisticness, using the Cityscapes data set, the CARLA simulator, and finally real-world scenarios.

All the proposed formulations leveraged a novel loss function introduced in the paper to improve the state-of-the-art methods for optimizing adversarial patches. The proposed loss function proved to be a more general and efficient alternative to the classic cross-entropy function for this kind of problems.

Furthermore, a new method called \textit{scene-specific attack} was introduced and tested on the CARLA simulator, showing that it improves the EOT formulation for a more realistic and effective patch placement.

%setting first studied the effect of targeted and untargeted patches on Cityscapes. Then, the untargated analysis is extended to a virtual 3D scenario and in a real-world setting.
Finally, a novel real-time detection algorithm (FPDA) was presented and evaluated with a large set of experiments, proving its capability of identifying those SS models that can be affected by dangerous adversarial patches on the Cityscapes data set and in a realistic driving scenario. 
%\TODO{Comment on take-away messages from the experimentations}
The extensive experimentation performed in this paper showed how real-time SS models present a certain robustness to real-world adversarial attacks, especially when considering the spatial distribution of the error induced by the adversarial patch. Moreover, it showed that the proposed FPDA detection algorithm is able to drastically reduce the threat associated with adversarial patches, as the adversarial effect is strictly related to neurons over-activations.
%real-world attack formulations are presented in order to evaluate , to investigate the limits of real-world attacks for segmentation neural networks in an autonomous driving scenario.
%Carrying out the investigations with increasingly ``real-world" benchmarks, we studied the effect of non-robust and EOT-based patches on the Cityscapes dataset, on a virtual 3D scenario, and in a real-world setting. 

% HO TOLTO QUESTA, RIMETTILA SE PENSI CHE ABBIA SENSO
% The achieved results opens to a new point of view for studying SS models in autonomous driving. Although the proposed attacks were able to reduce the baseline model accuracy, the SS models proved to be somehow robust to real-world patch-based attacks. This was especially noticeable when the tests were performed in more realistic settings using CARLA and the real world, where, in most cases, the patch only affected the proximity of the attacked surface.
% Nevertheless, this is a promising result, since it shows how the prediction provided by these models is not easily corruptible, especially in real-world scenarios. 
%This is in contrast with previous work on patch-based adversarial attacks against classification and object detection models.

As a future work, we aim at deepening several analyses in this field.
Since many variables of a realistic driving environment are not controllable (e.g., weather, geometric information, external objects), assessing the robustness of a model directly in the real world appears to be practically infeasible. 
As such, future work aims at studying the transferability of robust assessments derived from virtual scenarios (e.g., CARLA simulator) to real-world environments. 
Furthermore, the experiments conducted on the detection mechanisms highlight a strong relationship between the area attacked from an adversarial patch and the over-activation of internal features. This suggests the need for a deeper investigation on the spatial robustness of SS models. 
% \subsection{Threats to validity}

% % \begin{itemize}
% %     \item Transferbaility from Carla to Real-World
% %     \item Intra-class robustness analysis
% %     \item On the real-time defense
% % \end{itemize}

% \subsection{Final remarks}

%\vfill

\bibliographystyle{IEEEtran}
\bibliography{main}

\end{document}